\documentclass[pdflatex,sn-mathphys-num]{sn-jnl}

\usepackage{amssymb}
\usepackage{amsmath}
\usepackage[font=normalsize]{subcaption}
\usepackage{csquotes} 
\usepackage[inline, shortlabels]{enumitem}

\usepackage[dvipsnames]{xcolor}
\usepackage{soul}
\DeclareRobustCommand{\hlblue}[1]{{\sethlcolor{SkyBlue}\hl{#1}}}
\DeclareRobustCommand{\hlyellow}[1]{{\sethlcolor{Yellow}\hl{#1}}}
\DeclareRobustCommand{\hlgreen}[1]{{\sethlcolor{SpringGreen}\hl{#1}}}
\DeclareRobustCommand{\hlpink}[1]{{\sethlcolor{pink}\hl{#1}}}
\DeclareRobustCommand{\hlorange}[1]{{\sethlcolor{Dandelion}\hl{#1}}}

\usepackage{algorithm}
\usepackage{algpseudocode}
\algdef{SE}[DOWHILE]{Do}{doWhile}{\algorithmicdo}[1]{\algorithmicwhile\ #1}%

\newtheorem{definition}{Definition}
\newtheorem{property}{Property}
\newtheorem{proposition}{Proposition}

\newcommand{\eqdef}{\overset{def}{=}}  %NAR

\begin{document}

%\begin{frontmatter}

%% Title, authors and addresses

%% use the tnoteref command within \title for footnotes;
%% use the tnotetext command for theassociated footnote;
%% use the fnref command within \author or \affiliation for footnotes;
%% use the fntext command for theassociated footnote;
%% use the corref command within \author for corresponding author footnotes;
%% use the cortext command for theassociated footnote;
%% use the ead command for the email address,
%% and the form \ead[url] for the home page:
%% \title{Title\tnoteref{label1}}
%% \tnotetext[label1]{}
%% \author{Name\corref{cor1}\fnref{label2}}
%% \ead{email address}
%% \ead[url]{home page}
%% \fntext[label2]{}
%% \cortext[cor1]{}
%% \affiliation{organization={},
%%             addressline={},
%%             city={},
%%             postcode={},
%%             state={},
%%             country={}}
%% \fntext[label3]{}

%\title{A Formal Model of Human Values for Value-Aware Multi-Agent Systems}
\title{Modelling Human Values for Value-Aware Multi-Agent Systems}

%% use optional labels to link authors explicitly to addresses:
%% \author[label1,label2]{}
%% \affiliation[label1]{organization={},
%%             addressline={},
%%             city={},
%%             postcode={},
%%             state={},
%%             country={}}
%%
%% \affiliation[label2]{organization={},
%%             addressline={},
%%             city={},
%%             postcode={},
%%             state={},
%%             country={}}

\author*[1]{\fnm{Nardine} \sur{Osman}}\email{nardine@iiia.csic.es}

\author[1,2]{\fnm{Mark} \sur{d'Inverno}}\email{dinverno@hkbu.edu.hk}
%\equalcont{These authors contributed equally to this work.}

\affil*[1]{\orgname{Artificial Intelligence Research Institute (IIIA-CSIC)}, \orgaddress{\city{Barcelona}, \country{Spain}}}

\affil[2]{\orgdiv{Academy of Music}, \orgname{Hong Kong Baptist University}, \orgaddress{\city{Hong Kong}, \country{China}}}

%% Abstract
\abstract{
%\begin{abstract}
One of today's most pressing societal challenges is building AI systems whose behaviour, or the behaviour it enables within communities of interacting human and artificial agents, aligns with \emph{relevant} human values.  To address this challenge, 
%2026 we propose a formal model of human values that enables their explicit computational representation. 
we propose a formal computational framework for representing human values that provides the foundational structures required for value-aware reasoning in multi-agent systems. To our knowledge, this has not been attempted as yet, which is surprising given the growing volume of research integrating human values into AI systems. Taking as our starting point the wealth of research in human values from the field of social psychology, 
%2026 we set out to provide a formal model which captures value relations, value importance, and computational semantics in order to support both foundational reasoning and practical implementation. 
we set out to provide a formal model which captures value relations, value importance, and computational semantics in order to support the evaluation of behaviour with respect to values and the development of value-aware decision-making mechanisms in agent-based systems. 
%2026 We demonstrate how the model supports value-driven AI behaviour across real-world scenarios, establishing a bridge between abstract human values and concrete AI decisions. 
We demonstrate how the model supports the evaluation of behaviour in terms of value alignment across a real-world scenario, establishing a bridge between abstract human values and concrete agent behaviour. 
We illustrate how our model captures key concepts from social psychology research and outline a roadmap for 
%2026 future interdisciplinary research. 
incorporating values as first-class constructs in multi-agent systems.

%The ability to automatically reason over values not only helps address the value alignment problem but also facilitates the design of AI systems that can support individuals and communities in making more informed, value-aligned decisions. Increasingly, individuals and organisations are motivated to understand their values more explicitly and explore whether their behaviours and attitudes accurately reflect them. The work presented in this paper will enable the design and deployment of AI systems to meet this growing need. 
%\end{abstract}
}

%%Graphical abstract
%\begin{graphicalabstract}
%\includegraphics[width=\textwidth]{values-graphicalAbstract.png}
%\end{graphicalabstract}

%%Research highlights
%\begin{highlights}
%\item Introduces a formal model for the computational representation of human values. 

%\item Captures value relations, value importance, and computational semantics for AI reasoning. 

%\item Supports value-driven reasoning, linking abstract values to computational semantics. 
%AI reasoning

%Defines values via taxonomies, enabling computational reasoning and alignment. 

%\item Aligns the model with research from social psychology into human values.

%\item Identifies key research challenges and a roadmap for value-driven AI research. 
%Highlights should consist of 3 to 5 bullet points, each a maximum of 85 characters, including spaces
%\end{highlights}

%% Keywords
%\begin{keyword}
%% keywords here, in the form: keyword \sep keyword

%% PACS codes here, in the form: \PACS code \sep code

%% MSC codes here, in the form: \MSC code \sep code
%% or \MSC[2008] code \sep code (2000 is the default)
%\end{keyword}

%\end{frontmatter}
\maketitle
\vspace{-2em}
\bmhead{Statements and Declarations}\phantom{}\\
\textbf{Funding} This work was supported by the EU-funded VALAWAI (\#101070930) project and the Spanish-funded EVASAI (\#PID2024-158227NB-C31) project.\\
\noindent \textbf{Competing Interests} The authors have no competing interests to declare.

%\textbf{Author Contributions} These authors contributed equally to this work.

%% Add \usepackage{lineno} before \begin{document} and uncomment 
%% following line to enable line numbers
%% \linenumbers

%% main text
%%

\section{Introduction}\label{sec:intro}

%\subsection{Background}\label{sec:intro}
 
Across governments, industry, and the general public, there is an increasing recognition of the urgency for ethical approaches to the design and deployment of AI, as evidenced by the numerous publications of ethics guidelines (e.g., \cite{ec2019ethics,euArtificialIntelligenceAct,ieee,jobin2019global,UNESCO}). 
In academia, a growing body of research investigates the role of human values in designing ethical AI~\cite{GFAIH21,10.1007/s11023-020-09539-2,russell2019human,Weide2010}. 
Indeed, one of our leading AI research luminaries, Stuart Russell, believes the overarching goal of AI should change from ``intelligence" to ``intelligence provably aligned with human values"~\cite{russell2019human}.
This call to arms gave birth to the \emph{value alignment problem}. 

This challenge of engineering values into AI in response to the value alignment problem has resulted in a range of research areas: how human values can be learnt~\cite{valawai23,Lin2018AcquiringBK,10.5555/3463952.3464048,Liu2019PersonalityOV}; how individual values can be aggregated to the level of collectives~\cite{Lera-LeriBSLR22}; how arguments that explicitly reference values can be made~\cite{WallnerWZ24,Bench-Capon2009}; how decision making can be value-driven~\cite{RodriguezSoto2024AWAI,ChhogyalNGD19,ijcai2017p26,TostoD12}; how online institutions can ensure value-aligned behaviours in hybrid communities~\cite{noriega22,noriega23}; and how norms are selected or synthesised to maximise value-alignment~\cite{Rodriguez-SotoS25,MontesS21,SerramiaLR20,abs-2110-09240}. 

The ultimate goal is to develop AI that identifies and understands human values, reasons about them, and is capable of explaining behaviour in terms of those values~\cite{VALE23,VALE24,Osman24}. 
Yet despite the research on these various research fronts, no formal model of human values exists today that provides a concrete foundation for the emerging theoretical and computational approaches in values and AI. %al platform from which data structures and algorithms can be designed to build AI architectures that address the value-alignment problem. 
In response, we propose such %2026 a formal model of human values 
a formal computational framework for representing human values that serves as a foundational component for value-aware reasoning and decision-making in multi-agent systems, 
built on the following guiding principles: 
\begin{enumerate}[1) ]

\item we employ a formal language to be precise about modelling values and related concepts~\cite{dinverno97,luck95};  

\item we construct the formal components of this model to provide the foundations for designing %2026 the data structures and algorithms design needed for computational reasoning on values;  
data structures and algorithmic mechanisms that support value-aware decision-making and behaviour evaluation in multi-agent systems;

\item we design the model to be agnostic on any specific implementation or theory of human values, whilst providing examples of how the model can be used in practical scenarios; 

\item in designing the model we aim to subsume and relate to established concepts in AI research as much as possible; 

\item we ensure the model draws upon the wealth of work from within social psychology and we explicitly demonstrate the grounding of our model within this research; demonstrating how the proposed model incorporates the agreed conceptual underpinnings of this field; and  

\item we provide illustrative examples of applying the model in real-world use cases; demonstrating the practical value of the model and its general applicability.  
\end{enumerate} 

We do not propose a complete agent reasoning model; rather, we focus on providing the representational and computational foundations necessary for integrating values into agent reasoning and decision-making processes.

\paragraph{Impact} We argue that computational modelling of human values is essential for enabling AI systems to reason about values in a principled and interpretable way. 
In one of our ongoing work with Hospital del Mar, Barcelona~\cite{RodriguezSoto2024AWAI}, we are developing a system that helps medical professionals make value-aware decisions, considering the alignment of different decisions with the four bio-ethical principles~\cite{Beauchamp1979-BEAPOB}: beneficence, non-maleficence, autonomy, and justice. This is especially useful in cases of value conflicts. The developed tool also helps board members of a hospital ensure their medical protocols are aligned with their most relevant values.   
Building a computational model of values becomes fundamental for the tools and mechanisms reasoning with values~\cite{RodriguezSoto2024VALE}. Furthermore, not only the understanding of the computational semantics of `beneficence' in a hospital setting is critical, but also how it varies from one culture to another. These collaborations help us showcase the impact of our proposed formal model, as well as ensure it practicality, where we have the theory and practice tightly coupled. We do, however, choose another %a third 
use case as an illustrative example in this paper, due to its simplicity and ease of covering all aspects of our proposed model (more on this in Section~\ref{sec:valueTax}). 

\paragraph{Story and Scope} 
This paper does not focus on developing mechanisms for ensuring value aligned behaviour, 
%Whilst the research detailed in this paper provides a number of contributions,  it does provide the necessary mechanisms for ensuring value aligned behaviour, 
linking values with norms, or addressing conflicting norms and values (all of which are critical issues that we have been actively working on~\cite{RodriguezSoto2024VALE,RodriguezSoto2024AWAI,alba24,valawai23b,CurtoMSOC22,MontesS21,abs-2110-09240}). 
Instead, it takes a step back to provide a formal computational representation of human values that has not been previously attempted. 

%In making that attempt, we set out the proposal in eight as follows. 
This proposal is structured as follows. 
Section~\ref{sec:whyValues} motivates using values as critical modelling and architectural concepts for AI systems. 
It outlines why values are considered vital to understanding human behaviour and why explicit modelling and computational reasoning of human values is needed in AI. 
Section~\ref{sec:whatValues} explores the nature of human values from the perspective of social psychology, providing the background and framework for us to engineer a novel formal model. 
We analyse definitions of human values from existing works and consider their commonalities and distinctions. 
Armed with this knowledge, we then set out the details of our formal model in Section~\ref{sec:valueTax}. 
We formally define values by introducing value taxonomies, which encompass concepts such as value relations, value importance, and value semantics. We discuss how individuals or collectives may hold values, how values change with context and time, and how our proposed model can support reasoning about the value alignment of behaviours. 
With the formal model in place, we undertake, in Section~\ref{sec:SSHalignment}, a detailed mapping back to social psychology to assess the relationship of our modelling choices with this research. 
This is a critical step in ensuring and demonstrating exactly how our model is consistent with the research in this field. 
In Section~\ref{sec:roadmap}, we present a roadmap of AI research that needs to be undertaken to achieve value-driven behaviour.  
We close with a reflection on the contributions and further work in Section~\ref{sec:strengthsLimit} along with some concluding remarks in Section~\ref{sec:conc}.

We note that the work presented in this paper is the extension of a previously published conference paper~\cite{OsmanDInverno2024} which outlined the basic formal model. 
In this paper, we extend the model and provide more algorithmic details. We also motivate the need for promoting values as a key concept in symbolic modelling of agent-based systems (Section~\ref{sec:whyValues}), review the extensive work in social psychology (Sections \ref{sec:whatValues} and~\ref{sec:SSHalignment}) to demonstrate how our work is grounded in that research, and outline a detailed roadmap for future research toward our ultimate goal of building value-driven AI systems (Section~\ref{sec:roadmap}).

%%%%%%%%%%%%%%%%%%%%%%%%%%%%%%%%%%%%%%%%%%%%%%%%%%%%%%%%%%%%%%%%%%%%%%%%

\section{Promoting Values as a fundamental construct for AI}  
\label{sec:whyValues}

Human values have been heavily investigated within the humanities and social sciences to understand their role in behaviour choice and our attitudes towards various states of affairs. 
Schwartz, one of the prominent social psychologists working in the field of human values and well-recognised for his theory of basic human values, stated that ``theorists have long considered values central to understanding social behaviour... because they view values as deeply rooted, abstract motivations that guide, justify, and explain attitudes, norms, opinions, and actions''~\cite{zis-Schwartz2007Value}. 

%{\bf Check on ''Schwartz'' Nardine as I just changed one! Also is the word ''basic'' OK? Would ''fundamental'' be better in the paragraph above?}
%Nardine's answer: "basic" is the right word as it is the one used by Schwartz. His theory is officially called "Theory of basic human values".
 
Like Schwartz and many AI researchers, we recognise the criticality of factoring in values for automated reasoning. This is necessary both to guide the operation of AI itself and to build AI systems that support individual and collective human decision making. 
Traditionally, AI has used different human-associated qualities to guide behaviour, from calculating reward or utility (such as game theory~\cite{chalkiadakis2011computational,osborne1994course}), or following (or not) the norms of a community or society (normative systems~\cite{aagotnes2007logic,shoham1995social}), to the internal beliefs, goals and intentions inspired by cognitive sciences (agent BDI models~\cite{de2020bdi,dinverno98,rao1997modeling}), to name a few. 
However, as AI becomes increasingly prevalent in our everyday lives and issues arise around the trust we have in AI systems, it is becoming increasingly critical to consider how we design AI systems that can explicitly reason over  
human values. 
Significant research has made a compelling case for using human values in the design of ethical AI systems~\cite{GFAIH21,10.1007/s11023-020-09539-2,russell2019human,Weide2010}. 
There is a need to design AI systems that can explicitly reason about values to determine behaviour choices and, subsequently, communicate to various stakeholders how that value-based reasoning occurred. 
But how can we best incorporate values into the range of other primary constructs in AI that guide behaviour, such as norms or beliefs, for example? 

In multi-agent systems, behaviour is typically guided by constructs such as beliefs, goals, norms, or utility functions. However, these constructs do not explicitly capture the human values that motivate or justify such mechanisms. This limitation becomes particularly salient in hybrid human–AI collectives, where agents must act in ways that align with shared or individual values. Our work addresses this gap by introducing a formal framework that enables the explicit representation of values and the evaluation of agent behaviour with respect to those values.

\begin{figure}[!t]
\centering
\includegraphics[width=\textwidth]{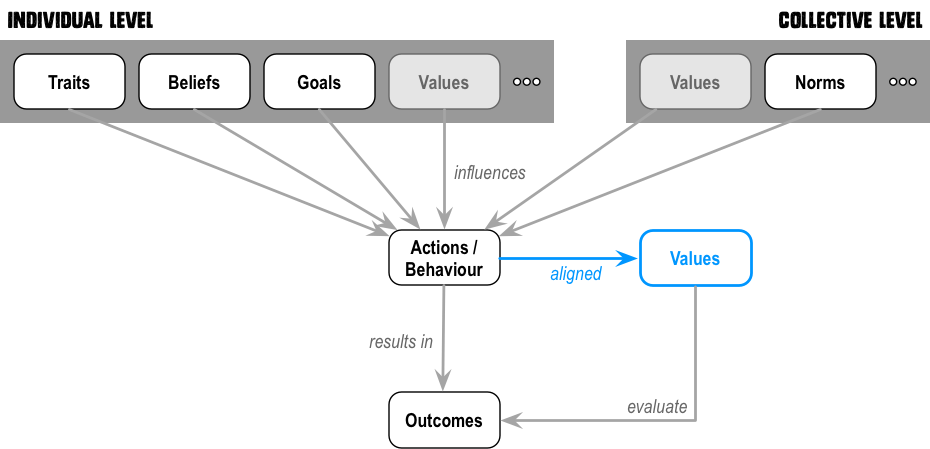}
\caption{Influencers of behaviour, and evaluating behaviour with respect to its alignment with values}
\label{fig:behaviour}
\end{figure}

In Schwartz's paper on the Theory of Basic Values~\cite{Schwartz2012AnOO}, he asks, ``How do values relate to attitudes, beliefs, traits and norms''? He states, ``[w]hen trying to explain why individuals behave as they do, people often refer to attitudes, beliefs, traits, or norms.'' 
We agree that the constructs guiding human behaviour are numerous, resulting in complex decision-making processes that select certain behaviours over others. 
As we illustrate in Figure~\ref{fig:behaviour}, there are many factors, both at the individual and at the collective level, which influence action choice and the resulting behaviour. 

%%% Nardine - 
%
Examples include an individual's traits, beliefs, goals, and values, as well as the values and norms of collectives. This list is not exhaustive, but it aims to illustrate the complexity of the decision-making process, individually and collectively. 

Investigating the dynamics between the constructs that influence behaviour is an issue that requires further scrutiny, as we discuss in our roadmap in Section~\ref{sec:roadmap}.  
Here, we argue that values should be elevated to a primary construct alongside beliefs, intentions, and norms %2026 (some of the constructs traditionally considered in AI-driven decision making). 
in agent-based decision-making. Explicitly representing values enables agents to evaluate and justify behaviour in terms of value alignment, complementing existing mechanisms such as norm compliance or utility maximisation.
To introduce values as a novel construct, we draw on the range and depth of investigations into values from social psychology. 

As such, values become a critical motivator of action choice and resulting behaviour in Figure~\ref{fig:behaviour}. 
In the past, research into AI considered mechanisms for behaviours and action choices with respect to beliefs, goals and intentions, as well as norms and maximising utility. 
One question was how behaviours can be understood in terms of (or aligned with) these key constructs.
Now, research efforts aim to understand how values influence decisions about behaviour and how this behaviour aligns with human values.
%

%Up to now, we have spoken of ``action choice'' and ``resulting behaviours''. This is a slightly arbitrary disambiguation, and to avoid getting swept into any issues regarding the semantics of these words, we will, from here, decide to use the words behaviour(s) and action(s) interchangeably.  

But how can alignment of behaviour with values be analysed? 
Figure~\ref{fig:behaviour} illustrates how the actions of human or artificial agents change the world. They result in observable outcomes that can be used to evaluate chosen behaviour.\footnote{Up to now, we have spoken of ``action choice'' and ``resulting behaviours''. This is a slightly arbitrary disambiguation, and to avoid getting swept into any issues regarding the semantics of these words, we will, from here, decide to use the words behaviour(s) and action(s) interchangeably.} 
If the outcome of a chosen behaviour results in a state of the world that respects, upholds or promotes a given value, we say that the behaviour that brought about that outcome is \emph{aligned} with that value. 
Furthermore, it means that the processes leading to actions/behaviour choices ---i.e. whatever constructs were used to make that behaviour choice, be it a norm, a personal belief, some combination, and so on--- are also aligned with that value. In other words, by evaluating the outcome of behaviour (bottom layer in Figure~\ref{fig:behaviour}), we can extrapolate to assess the behaviour itself (middle layer) and the constructs bringing about that behaviour (top layer). 
Of course, we understand that the process of selecting actions is often a complex process, especially in humans, as it might not always be clear whether the action is the result of following a norm, adhering to a personality trait, or being aligned with one's values. 

Our point of view is that if an outcome is aligned with a given value, then the constructs that brought about that outcome must also be aligned with that value, regardless of the intentions to be value-aligned. For example, if a regimented norm results in being aligned with a particular value, then, regardless of people's intentions, we say that the norm is aligned with that value.  

Evaluating a given behaviour and the decision-making process that produces it opens the door to AI (as well as AI-supported human) deliberation about why decisions were made and how they should be made in the future by considering the role of values. 

We conclude this section by noting that the motivation for introducing values into AI reasoning and decision making, as presented in the discussion above, %is based on the assumption of evaluating outcomes (or states of the world) with respect to values. 
rests on the assumption that it is possible to evaluate outcomes (or states of the world) with respect to their alignment with values. 
However, an explicit computational model for values is needed to achieve this. 
This is precisely the aim of this paper, which is presented in Section~\ref{sec:valueTax}: to provide a formal foundation for representing and evaluating values as part of agent behaviour in multi-agent systems. 
But before we proceed, we first consider research in social psychology on values to help us understand what values are, which will provide the foundation for designing our computational model. 

\section{Understanding Investigations in Human Values from Social Psychology}\label{sec:whatValues}

%{\bf Nardine - Just checking we are consistent in using European English except where it is in quotes!} YES

A significant body of work has set out to define or characterise values from the humanities and social sciences. Rohan~\cite{Rohan2000} acknowledges that ``[d]efinitional inconsistency has been epidemic in values theory and research'', and that ``[t]he status of values theory and research suffers because the word values is open to abuse and overuse by non-psychologists and psychologists alike.'' 
Kluckhohn~\cite{kluckhohn1951values} expands further on this:
\begin{displayquote}
 ``Reading the voluminous, and often vague and diffuse, literature on the subject in the various fields of learning, one finds values considered as attitudes, motivations, objects, measurable quantities, substantive areas of behavior, affect-laden customs or traditions, and relationships such as those between individuals, groups, objects, events.''
\end{displayquote}

As such, before proposing our formal model for values, we first investigate the range of different value definitions from the values literature. 
%2026, added:
The purpose of this section is not to provide a comprehensive survey of social psychology, but to extract those characteristics of human values that are necessary to inform their computational representation in agent-based systems. 
Figure~\ref{fig:valueDefs} provides a selection of those definitions from two review papers from social psychology~\cite{ChengFleischmann2010,Rohan2000}. 
We have set out to colour code this diagram to identify the various qualities characterising values and to support the reader in understanding where consensus lies or does not. 

\begin{figure}[t]
\footnotesize
\rule{\textwidth}{1pt}
    {\bf Kluckhohn~\cite[p.395]{kluckhohn1951values}.} 
        A value is ``a \hlyellow{conception}, explicit or implicit, distinctive of an \hlpink{individual}, or characteristic of a \hlpink{group}, of the \hlblue{desirable} which \hlgreen{influences the selection} from available \hlorange{modes, means, and ends of action}''.
    \rule{\textwidth}{0.1pt}
    {\bf Lewin~\cite[p.41]{lewin1952}.} 
        ``Values \hlgreen{influence behavior} but have not the character of a goal (i.e., of a force field)... the individual does not try to `reach' the value of fairness, but fairness is \hlgreen{`guiding' his behavior}... values are not force fields but they ``induce'' force fields.''
    \rule{\textwidth}{0.1pt}
    {\bf Guth and Tagiuri~\cite[p.124--125]{guth1965personal}}. 
    ``A value can be viewed as a \hlyellow{conception}, explicit or implicit, of what an \hlpink{individual} or a \hlpink{group} regards as \hlblue{desirable}, and in terms of which he or they \hlgreen{select}, from among alternative available \hlorange{modes, the means and ends of action}''.
    \rule{\textwidth}{0.1pt}
    {\bf Hutcheon~\cite[p.184]{Hutcheon72}}. 
    ``... values are not the same as ideals, norms, desired objects, or espoused beliefs about the `good', but are, instead, \hlyellow{operating criteria} for \hlorange{action}...''.
    \rule{\textwidth}{0.1pt}
    {\bf Rokeach~\cite[p.5]{rokeach1973nature}}. 
    ``A value is an enduring \hlyellow{belief} that a specific \hlorange{mode of conduct} or \hlorange{end-state of existence} is \hlpink{personally} or \hlpink{socially} \hlblue{preferable} to an opposite or converse mode of conduct or end-state of existence''.
    \rule{\textwidth}{0.1pt}
    %{\bf \cite[p.3--4]{Schwartz2012AnOO}}. 
    %(1) Values are \hlyellow{beliefs} linked inextricably to affect... (2) Values refer to \hlblue{desirable} \hlorange{goals} that \hlgreen{motivate action}... (3) Values transcend specific actions and situations... (4) Values serve as \hlgreen{standards or criteria}... (5) Values are ordered by \hlblue{importance} relative to one another... (6) The relative importance of multiple values \hlgreen{guides action}.
    {\bf Schwartz~\cite[p.20]{schwartz1994there}}. 
    A value is ``a \hlyellow{belief} pertaining to \hlblue{desirable} \hlorange{end states} or \hlorange{modes of conduct} that transcends specific situations; \hlgreen{guides selection or evaluation of behavior, people, and events}; and is ordered by the \hlblue{importance} relative to other values to form a system of value priorities''.
    \rule{\textwidth}{0.1pt}
    {\bf Feather~\cite[p.222]{feather1996values}}.  
    ``I regard values as \hlyellow{beliefs} about \hlblue{desirable or undesirable} \hlorange{ways of behaving} or about the desirability or otherwise of general \hlorange{goals}.''
    \rule{\textwidth}{0.1pt}
    {\bf Braithwaite and Blamey~\cite[p.364]{BraithwaiteRussell98}}. 
    ``Values... are \hlyellow{principles} for \hlorange{action} encompassing abstract \hlorange{goals in life} and \hlorange{modes of conduct} that an \hlpink{individual} or a \hlpink{collective} considers \hlblue{preferable} across contexts and situations''.
    \rule{\textwidth}{0.1pt}
    {\bf Friedman et al.~\cite[p.349]{friedman2013value}}. 
    ``A value refers to what a \hlpink{person} or \hlpink{group} of people consider \hlblue{important} in life''.
    \rule{\textwidth}{0.1pt}
    {\bf van de Poel and Royakkers~\cite[p.72]{van2023ethics}}. 
    Values are ``lasting \hlyellow{convictions} or \hlorange{matters that people feel should be strived for} in general and \hlpink{not just for themselves} to be able to \hlgreen{lead a good life} or \hlgreen{realize a good society}''.
\rule{\textwidth}{1pt}
\caption{A selection of value definitions, adapted from~\cite{ChengFleischmann2010,Rohan2000}. Text highlighted in yellow describes the nature of values, orange describes what values refer to or the issues they address, green describes the purpose of values or what they are for, blue highlights the notion of what is important or desirable, and pink describes who holds values or to whom they apply.}
\label{fig:valueDefs}
\end{figure}

\begin{itemize}
 \item The text highlighted in yellow identifies the different characterisations of values as either conceptions~\cite{guth1965personal,kluckhohn1951values}, beliefs~\cite{feather1996values,rokeach1973nature,schwartz1994there},  principles~\cite{BraithwaiteRussell98}, or convictions~\cite{van2023ethics}. 
 \item The text highlighted in orange illustrates how theorists agree on values referring to either outcomes or actions. 
 Within outcomes, the different terminology used includes end states~\cite{rokeach1973nature,schwartz1994there} or goals~\cite{BraithwaiteRussell98,feather1996values} (also referred to as ends of action~\cite{guth1965personal,kluckhohn1951values}). 
 Within actions, the different terminology used includes referring directly to actions~\cite{Hutcheon72,BraithwaiteRussell98} or ways of behaving~\cite{feather1996values} (also referred to as modes of conduct~\cite{BraithwaiteRussell98,rokeach1973nature,schwartz1994there}, or modes and means of action~\cite{guth1965personal,kluckhohn1951values}). 

\item The text in green highlights the various proposals on the purpose of values. 
It includes guiding or influencing behaviour~\cite{lewin1952}.  
This is also referred to as influencing the selection of modes, means and ends of actions~\cite{kluckhohn1951values,guth1965personal}, or guiding selection or evaluation of behaviour, people, and events~\cite{schwartz1994there}. 
Other values-related purposes include leading a good life or realising a good society~\cite{van2023ethics}. 

\item The blue text identifies how researchers describe what is desirable, preferable or important~\cite{BraithwaiteRussell98,feather1996values,friedman2013value,guth1965personal,kluckhohn1951values,rokeach1973nature,schwartz1994there}.

\item The text highlighted in pink illustrates how values apply to the level of the individual and to that of groups, communities or collectives~\cite{BraithwaiteRussell98,friedman2013value,guth1965personal,kluckhohn1951values,rokeach1973nature,van2023ethics}.

\end{itemize}

In our proposal for defining values, we remain neutral on whether values are considered beliefs, principles, convictions, or any other concept identified above or in other research. We understand that there might be no consensus on the terminology used by different authors, but due to the lack of clarity on the nature of values, we choose to remain neutral on this topic. 
However, we adopt the other aspects of values outlined above, where there seems to be a significant agreement on those concepts. %as many theorists' outputs show significant agreement. 
Namely, \emph{values guide behaviour by considering what actions or outcomes are desirable or essential for either individuals or collectives}. 
In other words, we propose our formal model for value representation by distilling the consensus of social psychology and building from there. 

Furthermore, we do not identify with any specific theory of values, which is one of our guiding principles, as described in the introduction (Principle 3).
We do not, for example, take Schwartz's defined set of ten ``universal'' human values~\cite{Schwartz2012AnOO}, or any other value theory~\cite{ChengFleischmann2010}, as the definitive theory. 
We follow in the footsteps of some value-sensitive design approaches~\cite{davis2015value}, which argue that values are open-ended and should be elicited from stakeholders bottom-up. 
Whilst there is evidence for humans sharing fundamental human values (as illustrated by Schwartz~\cite{Schwartz2012AnOO}), there is also evidence that new values are continuously emerging with new application domains and technologies (as shown by van de Poel~\cite{vandePoel2018}).  Our ongoing fieldwork and collaboration with stakeholders in real-life use cases from various domains further validate this. For example, Subsection~\ref{sec:howEG} illustrates how a social networking app can prioritise different values for different communities, and our ongoing work with firefighters illustrates how values may also change from one geographical location to another (such as the values of the firefighters in Ecuador being categorically different from those in Catalonia). 
In summary, the view that values may change is central to our model; hence, we must remain agnostic toward the selected value theory and accommodate value change. 

We now present our formal model in Section~\ref{sec:valueTax}.
This will then be followed by a thorough analysis of the alignment of our proposal with selected works from social psychology in Section~\ref{sec:SSHalignment}. 
\section{A Formal Computational Model for Human Values}\label{sec:valueTax}

Building on the body of research from social psychology described in the previous section as our starting point, we aim to develop a formal computational model for representing human values. 
%2026, added:
In this section, we present the formal components of our model and show how they support the representation and provide the basis for evaluation of values with respect to agent behaviour. The model is designed to capture key aspects of human values, including their structure, their grounding in observable properties, and their role in evaluating behaviour. %Together, these components provide the basis for assessing the value alignment of behaviours and for comparing alternative courses of action in multi-agent systems.
%
%This computational approach for representing values provides the foundations for future developments of computational mechanisms that reason over values, enabling AI systems to make value-aware decisions. 

%%%%%

The proposed computational model of human values is presented in four subsections.
The first subsection provides a formal definition of values through the notion of ``value taxonomies'', which encompasses concepts such as value relations, value importance, and value semantics.
The second subsection extends our model to consider the widely held belief that values change over time, providing an initial account of the dynamic nature of values. 
To achieve this, we introduce the notion of contexts, enabling agents to evaluate their current values within their current situation to determine future behaviour.
The third discusses modelling values for individuals and groups of individuals, which we call collectives. 
In our proposal, a collective is a hybrid multi-agent system containing artificial and human individuals, such as online communities or other organisational setups.  
The final subsection considers the problem of ensuring the extent (or degree) to which the behaviours of individuals or collectives are aligned with an agreed set of values; the \emph{value alignment problem}. 
A formal description of value alignment is then presented, %demonstrating some of the utility of our model, which is the capability to capture key research questions in our field of research formally. 
demonstrating how our model enables the structured formulation of core research challenges in this field. 

Each of those four subsections is divided into three parts: our proposal, a discussion of implementation choices arising from the model, and a running descriptive example of a system we have developed that demonstrates a concrete implementation choice. 
The running example is introduced to support the reader in understanding our proposed foundational model and its significance. 

\subsection{What are human values? A computational response to value representation}\label{sec:what}
As per Section~\ref{sec:whatValues}, we take values to be human abstract concepts that
guide behaviour, whose exact meaning and interpretation vary both with the current context and over time. 
However, a concrete computational representation of values requires a machine to reason with values. 
For example, while one may talk about the value `fairness' in general, for a specific application supporting a mutual aid community, fairness might be understood as ``one does not ask for help more than what they volunteer''. That is, the abstract concept of fairness acquires a concrete definition (grounding semantics) through a {\it property} whose satisfaction (or degree of satisfaction) can be automatically verified in a given system (more on this example in Subsection~\ref{sec:ex-vtax}). In this case, the property specifies that a user does not ask for more help than that which they volunteer. 

\begin{figure}
\centering
\includegraphics[width=0.55\textwidth]{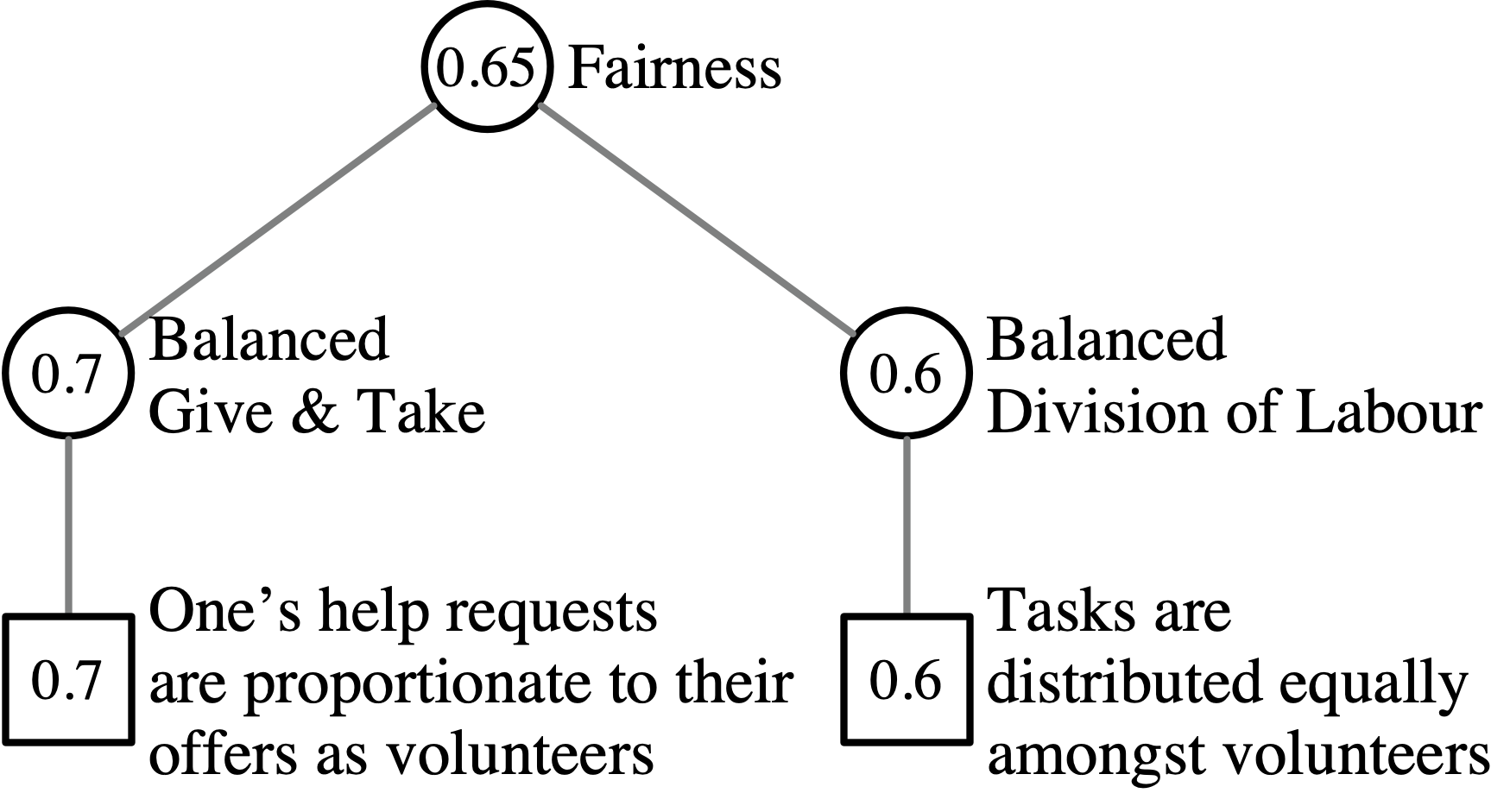}
\caption{Different levels of abstraction for the value fairness, with various grounding semantics: numbers indicate node importance}
\label{fig:egGeneral}
\end{figure}

Of course, there may be different levels of abstraction and grounding semantics for a value. For example, suppose the application were to be adopted by another community where volunteers specifically support older people. The new community might find the old view of fairness ---that one only asks for help precisely what they have volunteered--- unsuitable because this community expects to have volunteers who support older people. In contrast, older people usually only ask for help with their day-to-day tasks without volunteering to help others. As such, this new community would emphasise that fairness is not about a balanced give and take, but about a balanced division of volunteer workload. Here, we encounter different levels of abstraction and grounding properties, illustrated in Figure~\ref{fig:egGeneral}. 

The top node in Figure~\ref{fig:egGeneral} presents the abstract concept of fairness; the middle nodes show different concepts of fairness that are more specific than the top node, 
(such as reciprocity, which is understood as balanced give and take, or fair treatment that may be defined in terms of balanced division of workload). This highlights different abstraction levels and interpretations; the bottom nodes present properties that ground the semantics of abstract concepts, allowing for a computational evaluation of values. 
For example, the bottom left node represents the grounding semantics of fairness for the mutual aid community. 
Conversely, the bottom right node represents the grounding semantics of fairness for the community supporting older people. 
We use square nodes for nodes that ground the semantics of abstract concepts.

This interplay between the abstract and the grounding semantics of values is reflected through the relations between nodes. %, as shown in Figure~\ref{fig:egGeneral}. 
Relations also capture the different abstraction levels that may exist for value concepts. % and their grounding semantics. %, enabling computational approaches to reasoning over values. 

Another key concept in social psychology is that of value priority, which determines how \emph{important} a value is for individuals or collectives. In other words, it is not just the semantics of fairness that influence the behaviour of an individual or a collective, but how important fairness is as a value for that individual or collective.  As Schwartz articulates in~\cite{Schwartz2012AnOO}: ``Values influence action when they are relevant in the context... and important to the actor''. 
We model \emph{value importance} by attaching a measure of \emph{importance} to each node of Figure~\ref{fig:egGeneral}. As for Schwartz's reference to the relevance of a value in  the context, this is addressed in Subsection~\ref{sec:context}, where relevance in context is also captured through value importance.

Given our view of values having different abstraction levels and different grounding semantics, we propose values to be defined through taxonomies where the general concept becomes more specific as one travels down the taxonomy, and it becomes concrete and computational at (some) leaf nodes. 
This is consistent with research in value-sensitive design on value change taxonomies~\cite{vandePoel2018}, as well as more traditional value theories that introduce value hierarchies, such as Schwartz's Theory of Basic Values~\cite{Schwartz2012AnOO}. Subsection~\ref{sec:SSrohan2} elaborates in detail on the alignment of our proposed taxonomy with some of the works in social psychology, and Figure~\ref{fig:valueTypes} illustrates how Schwartz's value types can be represented through our proposed taxonomies. %, minus the opposing value relations of Schwartz. 

Furthermore, using taxonomies facilitates navigation between abstract and concrete grounding semantics of values, which is especially useful when defining values, tracking their evolution, and supporting deliberation and negotiation around them (more on this in Section~\ref{sec:roadmap}). While one might argue that machines only need property nodes to align behaviour with values (as discussed in Subsection~\ref{sec:why}), our example of the community of volunteers supporting older people illustrates how different groups can develop distinct interpretations of the same value, like fairness.
A taxonomy representation makes it possible to deliberate over these varying semantics and support their evolution across contexts and over time.
As demonstrated in Subsection~\ref{sec:context}, we expect such semantics to continually evolve with new interaction requirements. %In this way, taxonomies also aid in deliberating not only over value meanings but also their relative importance.
%We expect semantics to continuously evolve with new interaction requirements, as seen in Subsection~\ref{sec:context}. Moreover, a taxonomy can support the deliberation process over values and their importance. 

Definition~\ref{def:valueTaxonomy} formalises a value taxonomy $\mathcal{V}$ as a triple of nodes $N$ (where nodes can be of two types, nodes representing abstract concepts $N_{l}$ and property nodes that ground abstract concepts $N_{\phi}$), edges between those nodes $E$ that describe the relations between value concepts (which is more general/specific than which), and an importance function $I$ which is a total or a partial mapping from the nodes to a number.

\begin{definition}[Value taxonomy]\label{def:valueTaxonomy}
A value taxonomy $\mathcal{V}=(N,E,I)$ is defined as a directed acyclic graph, where: 
\begin{enumerate}
    \item The set of nodes $N=N_{l}\cup N_{\phi}$ represents value concepts, and it is composed of two types of nodes: i) those that are specified through labels, with $N_{l} \subset \mathrm{L}$ representing the set of label nodes and $\mathrm{L}$ is the set of all value labels representing abstract value concepts like `fairness' or `reciprocity'; and ii) those that are specified through concrete properties, with $N_{\phi} \subset \Phi$ representing the set of property nodes and $\Phi$ is the set of all value properties whose satisfaction (or degree of satisfaction) can be automatically verified at different world states, such as having the number of times one offers help in a mutual aid community to be greater than the number of times one asks for help. 
    \item The set of edges $E : N \times N $ is a set of directed edges $(n_{p},n_{c}) \in E$ that represent the relation between value concepts $n_{p}$ and $n_{c}$ (the parent and child nodes, respectively) illustrating that the value concept $n_{p}$ is a more general concept than $n_{c}$. 
    \item The importance function $I: N \to C\!D_{\bot}$ either assigns an importance measure from the codomain $C\!D$ to value concepts in $N$, or assigns $\bot$ to value concepts when their importance measure is undefined.
\end{enumerate}
\end{definition}

First, we note that property satisfaction can exhibit varying degrees, rather than being limited to a binary true/false evaluation. This is illustrated in Section~\ref{sec:why} (through the satisfaction degree function $sd$ of Equation~\ref{eq:alignment}, and the concrete examples of Equations \ref{eq:p1Sat} and~\ref{eq:p3Sat}), which explores how the satisfaction of property nodes influences the computation of value alignment.  

\begin{figure}[t]
\centering
\includegraphics[width=0.4\textwidth]{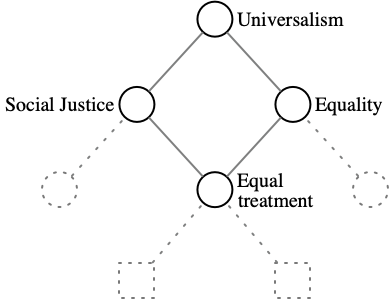}
\caption{This fragment of a taxonomy shows how some value concepts may have more than one parent node}
\label{fig:multiple_parent_nodes}
\end{figure}

We note here that we specify the value taxonomy as a directed acyclic graph, as opposed to the more traditional taxonomy tree, because certain value concepts (nodes) may be used to narrow the understanding of more than one general concept: that is, one value concept (node) may have more than one parent node. For example, the value `equal treatment' in the taxonomy of Figure~\ref{fig:multiple_parent_nodes} can act as a more specific concept for both the `social justice' and `equality' values (where both `social justice' and `equality', along with their inclusion under the broader `universalism'  value, follow Schwartz's value hierarchies~\cite{Schwartz2012AnOO}). 
Of course, Figure~\ref{fig:multiple_parent_nodes} presents only a fragment of a sample taxonomy to illustrate the need for a directed acyclic graph. For this taxonomy to be helpful in reasoning, first, `equal treatment' should have property nodes as its descendants to provide the abstract nodes with computational semantics. Furthermore, `social justice' and `equality' would be expected to have distinct child nodes beyond their shared child, `equal treatment'. While the shared node reflects a point of semantic overlap between the two concepts, their unique child nodes would capture how their semantics diverge. Without such distinct children (that is, if all their children were shared), the two nodes would be semantically identical.  

We also require property nodes to be restricted to leaf nodes. In other words, the concrete grounding semantics of a value concept (specified as a property node) cannot be more general than another value concept (specified as another property or label node). For example, in Figure~\ref{fig:egGeneral}, the property of having one's help requests proportionate to their volunteering offers cannot be more general than abstract concepts such as balanced give and take. This requirement is defined formally as follows: there is no pair of parent/child nodes in any set of edges where the parent node is a property node. 
    $$ \nexists \; (n_{p},n_{c}) \in E \cdot n_{p} \in N_{\phi}$$
As we illustrate next, we introduce this condition to simplify the construction and interpretation of value taxonomies. Recall that the children of any parent are more specific than that parent. Conceptually speaking, it is arguably paradoxical for concrete property nodes to be parents of abstract value concepts specified through labels. However, we may allow a property node to be a parent node of another property node, provided that the parent node's property subsumes the child node's property. This, however, would require mechanisms that check the subsumption of properties. 
Furthermore, we illustrate in Subsection~\ref{sec:why} how property nodes are used to assess value alignment. Value alignment is computed based on the satisfaction of the different property nodes. Suppose property nodes may appear anywhere in a taxonomy. In that case, a question that arises is whether all property nodes should be treated equally or whether the depth of a property node might affect its weight and, hence, the value alignment computation. 
For all these reasons, we choose to simplify the structure of value taxonomies for the time being by restricting property nodes to leaf nodes. Future work may relax this requirement if property subsumption mechanisms are introduced and value alignment mechanisms are designed to consider the location of property nodes in a taxonomy. 

Coherence is an essential property regarding value importance within a well-defined taxonomy. 
We articulate this coherence by requiring that a parent node's importance be aligned with the importance of its children. 
For example, if the importance of all children nodes is low, then the importance of the parent node cannot be high, and vice versa. We formally define the coherence of value importance to hold in a taxonomy if and only if all parent nodes' importance measures are an aggregation of their children nodes' measures:  

\begin{definition}[Coherence of value importance]\label{def:coherence} 
Importance within a value taxonomy $\mathcal{V}=(N,E,I)$ is said to be coherent if and only if, for all nodes $n\in N$ with children nodes (there exists $(n,n_{c})\in E$), the importance of the parent node is an aggregation of the importance of its children nodes:
    \begin{equation}\label{eq:coherence}
        \forall \, n\in \{n_{p}\,|\, (n_{p},n_{c}) \in E\} \;\cdot\; I(n) = \mathop{\mathbf{A}}\limits_{n'\in X_{n}} I(n')
    \end{equation}
where $X_{n} = \{n_{c} \,|\, (n,n_{c}) \in E \}$ is the set of all children nodes of $n$, and $\mathbf{A} : C\!D^{m} \to C\!D$ is an aggregation function that takes a set of size $m\in \mathbb{N}^{*}$ of importance measures in $C\!D$ (specified as $C\!D^{m}$) and returns an aggregation of those measures, where the aggregation also falls in the same range $C\!D$. 
\end{definition}

Of course, humans are not always coherent when expressing value importance, nor are they always coherent in their behaviour (they might not always be aligned with the values they believe are essential). %when expressing value importance and how they use it to inform attitudes or decision making. 
Nevertheless, we aim to design AI that is coherent in its reasoning over values, and hence the need for coherent taxonomies that support coherent value reasoning. %that enables the design of and reasoning over coherent taxonomies (i.e. coherent value-reasoning). 
We are interested in designing AI that can help analyse discrepancies between stated values and observed behaviour.  
%
%In our case study work, we see how AI can help identify non-coherence in the behaviour of individuals or communities by highlighting what needs to be considered in order to achieve coherence as we have defined it. 

The importance measures of some of the taxonomy nodes may be provided manually by humans, but can also be learned from other sources, such as past interactions or experiences. 
A coherence mechanism is then needed to ensure that the obtained importance measures are coherent within a taxonomy. Propagation mechanisms could be constructed to calculate the importance of nodes that have not been provided, building on existing propagation mechanisms~\cite{OsmanSS10,LiquidPub}. Any propagation mechanism must respect the coherence of value importance within the taxonomy. 

Now let us now consider the aggregation function $\mathbf{A}$. We argue that symmetry, idempotence and monotonicity, which we define below, are some of the desirable properties to be held by any such function. 
Symmetry states that the ordering of the importance measures being aggregated does not matter; it is precisely what we require from our model. 
When calculating a parent node's importance, the ordering of the importance measures of its child nodes should not affect the aggregation. 
Idempotence states that the aggregation of several instances of the same measure yields the same measure, which is also what we require from our model. 
If we assume that all children of a parent node have the same importance $i$, then the parent node should also share that importance. It should neither be more important nor less important than $i$. 
Monotonicity states that any increase in the importance of aggregated measures implies a non-decrease with respect to the aggregated value, which is, again, what we require from our model. Increasing any importance measure of the children nodes should not decrease the parent node's importance. 

Formal definitions of these properties are presented next.
\begin{property}[Symmetry of aggregation]\label{prop:symmetry} 
The aggregation function $\mathbf{A}$ is symmetric if, for all sets of importance measures $\lambda \in C\!D^{m}$ and all permutations $\pi \in \Pi_{\lambda}$ of those sets (where $\Pi_{\lambda}$ is the set of all permutations of $\lambda$), we have:
$$\mathbf{A}(\lambda) = \mathbf{A}(\pi)$$
\end{property}
That is, the ordering of aggregated values does not matter.

\begin{property}[Idempotence of aggregation]\label{prop:idempotence}
An aggregation function $\mathbf{A}$ is idempotent if, for all %$m\in \mathbb{N}^{*}$ and all $\lambda \in C\!D^{m}$ where $\lambda = \{i, \ldots, i\}$
importance measures $i\in C\!D$, we have: 
$$\mathbf{A}(i, \ldots, i) = i $$
\end{property}
That is, aggregating several instances of the same measure will return that same measure.

\begin{property}[Monotonicity of aggregation]\label{prop:monotonicity} 
An aggregation function $\mathbf{A}$ is monotonous if, for all sets of importance measures $\lambda, \lambda' \in C\!D^{m}$, we have: 
$$ (\; \forall \; 0<p\leq m \;\; \cdot \;\; \lambda_{p} \leq \lambda'_{p}\;) \quad \Rightarrow \quad \mathbf{A}(\lambda) \leq \mathbf{A}(\lambda')$$
where $\lambda_{p}$ represents the element in position $p$ of the set $\lambda$.
\end{property}
That is, if the measures in set $\lambda$ are smaller than or equal to the measures of $\lambda'$ (per position), then the aggregation of the measures of $\lambda$ should be smaller than or equal to the aggregation of the measures of $\lambda'$. 

Our requirements for Properties \ref{prop:symmetry}--\ref{prop:monotonicity} help define the type of aggregation operator, as we show next.
First, from~\cite[p.~14]{12-Ma98}, we know that the idempotence and monotonicity properties imply compensativeness.
\begin{proposition}\label{Compensative operator}
If an aggregation function $\mathbf{A}$ satisfies the idempotence and monotonicity properties (Properties \ref{prop:idempotence}~and~\ref{prop:monotonicity}), then $\mathbf{A}$ is a compensative aggregation.
\end{proposition}
Compensative operators are a class of aggregation operators that fall outside the classes of conjunctive and disjunctive operators. These operators are limited between the $\min$ and $\max$, which are the bounds of the t-norm and t-conorm families (the operations that generalise the logical conjunction and logical disjunction). This implies that for any set of importance measures (for all $\lambda \in C\!D^{m}$), the aggregation will fall between the minimum and maximum measures in that set:
$$\min \lambda \leq \mathbf{A}(\lambda) \leq \max \lambda $$
We believe falling between the minimum and maximum is appropriate for our context: a parent node should not be more important than its most important child node, nor less important than its least important child node.

Furthermore, from~\cite[p.~13]{12-Ma98}, we also know that compensative aggregation operators that satisfy the symmetry and monotonicity properties are averaging operators.
\begin{proposition}\label{Averaging}
If a compensative aggregation function $\mathbf{A}$ satisfies the symmetry and monotonicity properties (Properties \ref{prop:symmetry}~and~\ref{prop:monotonicity}), then $\mathbf{A}$ is an averaging operator.
\end{proposition}

As such, we propose $\mathbf{A}$ to be an averaging operator, though the exact choice of this operator is left for implementation.

\subsubsection{Implementation Choices}\label{sec:whatImp} 
The following implementation choices illustrate how the formal model can be instantiated. They are not intended as prescriptive or exhaustive.

\paragraph{Specifying property nodes} 
We use properties to specify the grounding semantics of values, leveraging their traditional role in describing world states and the fact that their satisfaction can be computationally evaluated~\cite{Stirling2001}. %as properties have traditionally been used to describe the world state, and their satisfaction can be computed~\cite{Stirling2001}. 
The exact language used for specifying properties is an implementation decision. 
Our proposal is agnostic regarding the choice of language; we intentionally do not tie our proposal to any specific implementation language or decision. 
Our model can encompass any theory written in propositional, first-order, deontic, modal and other logics. 

We note that the use of properties might initially appear to embody a consequentialist view, where only the outcome of behaviour is what matters. 
However, in our model, properties can be designed to evaluate not only action outcomes but also those attached to actions themselves (such as whether an action is permitted; for example, lying is never acceptable). 
The expressiveness of the chosen language for implementing properties plays a significant role here. 

However, in this paper, and as illustrated in our example in Subsection~\ref{sec:ex-vtax}, we expressly limit the examples to propositional logic and, more specifically, simple propositions to improve readability. Complex languages with higher expressivity can be chosen. 
%
%Alternative and more simplified approaches may also be investigated. For example, one may choose to describe the world state through percepts instead of properties, where the detection of percepts can also be verified or checked. Again, this is a matter of implementation choice. The bottom line is that the satisfaction of property nodes must be computationally verifiable.

\paragraph{Choosing the codomain of value importance} 
Concerning the importance measure of a value concept, choosing the codomain $C\! D $ that evaluates this importance $ I $ is also an implementation decision. 
Schwartz, for instance, uses the range $\{-1,0,1,2,3,4,5,6,7\}$ for people to specify the importance of a value, such as `equality' (equal opportunity for all) or `pleasure' (gratification of desires), as a guiding principle in their lives, where $-1$ represented opposing a value, $0$ represented considering the value to be non-important, and positive numbers meant different degrees of supporting a value (`supreme importance', `very important', `important', and so on)~\cite{Schwartz2012AnOO}. Other ranges may be considered. For example, importance could be specified as a number in the range $[0,1]$, where $0$ represents complete non-importance of a value, and $1$ represents its utmost importance, or a number in the range $[-1,1]$, where $-1$ represents opposing a value, $1$ represents utmost importance, and $0$ represents indifference; or even a number from the set of integers $\mathbb{Z}$ that does not limit the degree of importance/opposition. 

Of course, alternative approaches, such as defining importance measures as ranges instead of specific numbers or even a normal distribution, may also be considered. Another possible approach is to define importance as a partial or total order, rather than having a codomain. 
In fact, Schwartz argues that it is this order (the relevant importance) that impacts reasoning about values, where the order is deduced from the importance assigned by people~\cite{Schwartz2012AnOO}. 

A partial order might be more intuitive for humans to provide, as opposed to giving numerical importance measures. %Human stakeholders may use partial orders to specify importance measures when this is the case. 
As partial orders can be easily translated into numeric values, this aligns with our proposal (Definition~\ref{def:valueTaxonomy}). 
An example of translating partial orders into numerical values is provided by Serramia et al.~\cite{10.5555/3237383.3237891}.

The implementation choice should depend on the specific requirements of the application. For example, in our medical domain use case~\cite{RodriguezSoto2024AWAI}, all three medical doctors involved in the discussions on bio-ethical values provided a total order, concurring that (at least at their hospital) justice was the most important value (understood as requiring appropriate distribution of benefits, risks, and costs fairly), followed by autonomy (requiring respecting the decisions of autonomous patients), then non-maleficence (requiring not causing harm to patients) and finally beneficence (requiring providing benefits to patients).  

In the example of this paper, we set the codomain to $[-1,1]$ as reasoning with numbers is computationally easier than reasoning with partial orders. Furthermore, the range $[-1,0]$ is more intuitive for describing opposition to a value. This expands Schwartz's opposition measures, as there may also be degrees in opposing a value, just as there are degrees in supporting a value. The choice to limit the range to $[-1,1]$, as opposed to (say) $\mathbb{Z}$, reflects a preference for bounded, normalised scales that facilitate comparison, aggregation, and interpretability. 
Recall that partial orders provided by stakeholders can easily be translated into the chosen range $[-1,1]$, as mentioned above.

\paragraph{Ensuring the coherence of value importance} 
As for the aggregation function $\mathbf{A}$ that ensures the coherence of value importance within a taxonomy, different averaging operators may be explored. In this paper, we propose a simple average:
\begin{equation}\label{eq:coherence_imp}
    %I(n_{p}) = \frac{\displaystyle\sum_{n_{c}\in X_{n_{p}}} I(n_{c})}{|X_{n_{p}}|}
    I(n_{p}) = \displaystyle\sum_{n_{c}\in X_{n_{p}}} I(n_{c})\;/\;|X_{n_{p}}|
\end{equation}

As argued above, we assume the importance measure of some nodes to be either provided manually by stakeholders or learned from ongoing behaviour. 
A propagation mechanism may then be implemented to calculate the importance of nodes that have not been provided nor learnt. % require manual provision or learning. 
Different propagation mechanisms can be explored to verify the coherence of existing importance measures as they propagate these measures to other nodes, while ensuring the overall coherence of value importance within the taxonomy (following Definition~\ref{def:coherence}). One sample propagation mechanism is presented in Algorithm~\ref{alg:propagation}.

%%%% ALGORITHM 1 
\begin{algorithm}[!p]
\caption{Propagation algorithm for value importance }\label{alg:propagation}
\scriptsize
\begin{algorithmic}[1]
%%%
\Require $N$ to be a set of nodes, $E$ to be set of edges between nodes ($E : N \times N$), and $I$ to describe the importance assigned to nodes in $N$ ($I : N \times [-1,1]$).  
\Require \textsc{GetRoots}($N$,$E$) returns roots in $N$, given edges $E$; \textsc{Children}($n$) returns children of $n$; 
\textsc{Val}($n$) returns importance of $n$; 
\textsc{VSum}($N$) returns sum of the importance of nodes in $N$, and 
\textsc{DVW}($N$) returns true if some descendant of $N$ has its importance set, false otherwise.
%%%
\Function{Propagate}{$N$,$E$,$I$} \label{ln:main}
    \State $Roots \leftarrow$ \textsc{GetRoots}($N$,$E$); \label{ln:roots}
    %\State $R \leftarrow Roots$;
    %\State $I^{t} \leftarrow I$;
    \State $I^{t} \leftarrow$ \textsc{Propagate2}($Roots$,$E$,$I$); \label{ln:propagateRoots}
    \State \Return $I^{t}$
\EndFunction

\Function{Propagate2}{$Nodes$,$E$,$I$}
    \Do
    %%%
        \For{$n \in Nodes$}
            \State $C \leftarrow$ \textsc{Children}($n$);
            \State $C' \leftarrow  \{c' \in C~|~$ \textsc{Val}$(c')\neq$ nil\};
            \State $C'' \leftarrow C\backslash C'$;
            \If{\textsc{Val}($n$)$\neq$ nil} \label{ln:case1Start}
                \If{$C == \emptyset$}\label{ln:noProp1}  %\Comment{do nothing}
                    \State do nothing
                \ElsIf{$|C''|==0$} \label{ln:check}
                    \If{\textsc{Val}($n$)$\neq$(\textsc{VSum}$(C)/|C|)$}
                        \State {\bf output} ``ERROR with $n$'s value" 
                        \State \Return nil
                    \EndIf
                \ElsIf{$C'' == \{c''\}$}\label{ln:propagateDC}
                    \State $imp \leftarrow $ (\textsc{Val}($n$)$\times |C|)$ $-$ \textsc{VSum}($C'$);
                    \State $I^{t} = (c'',imp) \cup I^{t}$;
                        %\Comment{c's only possible importance value is assigned to it}
                \ElsIf{$|C''| > 1~\wedge~\neg$\textsc{DWV}($C''$)}\label{ln:propagateDA}
                    \State $imp \leftarrow$ $(($\textsc{Val}($n$)$\times |C|)-$\textsc{VSum}($C'$)$)/|C''|$;
                    \For{$c''\in C''$}
                        \State $I^{t} = (c'',imp) \cup I^{t}$;
                    \EndFor
                \Else%{$|C''| > 1~\wedge~\neg$\textsc{DWV}($C''$)}
                \label{ln:nopropagate}
                    %\Comment{This is the case where $n$ has more than one child without an assigned importance and at least one of those has a descendant with an assigned importance}
                    %\Comment{do nothing}
                    \State do nothing
                \EndIf
                \State $I^{t} \leftarrow$ \textsc{Propagate2}($C$,$E$,$I^{t}$); \label{ln:recursive1}
            \Else %\Comment{$r$ is not assigned an importance value} 
            \label{ln:case2Start}
                \If{$C == \emptyset$} \label{ln:noProp2}%\Comment{do nothing} 
                    \State do nothing
                \ElsIf{$|C''|==0~\wedge~|C'|>0$}\label{ln:propagateUC}
                    \State $imp \leftarrow$ \textsc{VSum}($C'$)$ /|C'|$;
                    \State $I^{t} = (n,imp) \cup I^{t}$;
                \ElsIf{$|C'|>0~\wedge~\neg DWV(C'')$}\label{ln:propagateUA}
                    \State $imp \leftarrow $ \textsc{VSum}($C'$) $/ |C'|$;
                    \State $I^{t} = (n,imp) \cup I^{t}$;
                    \For{$c''\in C''$}
                        \State $I^{t} = (c'',imp) \cup I^{t}$;
                    \EndFor
                \Else \label{ln:nopropagate2}%\Comment{do nothing}
                    \State do nothing
                \EndIf
                \State $I^{t} \leftarrow$ \textsc{Propagate2}($C$,$E$,$I^{t}$);\label{ln:recursive2}
            \EndIf
        \EndFor
    %%% 
    \doWhile{$I^{t}\neq I$}\label{ln:conditionRepeat}
\EndFunction
%%%
\end{algorithmic}
\end{algorithm}

The algorithm begins with the \textsc{Propagate} function (line~\ref{ln:main}), where it obtains all root nodes (line~\ref{ln:roots}) and then initiates the propagation process for each of these nodes (line~\ref{ln:propagateRoots}). 
This is achieved by calling the \textsc{Propagate2} function. This function is recursive. After it propagates importance measures with respect to a given node, it moves on to continue the propagation with respect to the children nodes (lines \ref{ln:recursive1} and~\ref{ln:recursive2}). When propagating for a given node, either we attempt to propagate the importance measure down to its children nodes if the node already has an importance measure assigned to it (case of lines~\ref{ln:case1Start}--\ref{ln:recursive1}), or we attempt to propagate the importance measures of the children up if the node does not have an importance measure assigned to it (case of lines~\ref{ln:case2Start}--\ref{ln:recursive2}). 

Nothing must be done when propagating down if the node has no children nodes (line~\ref{ln:noProp1}). 
If the node has no children nodes with no assigned importance measures, then nothing needs to be propagated, but the algorithm checks whether the parent node's importance measure is coherent with those of its children nodes. 
If this is not the case, the algorithm prints an error message and halts (case starting with line~\ref{ln:check}). 
If there is exactly one child node with no assigned importance, and regardless of how many children nodes do have assigned importance measures, then the child node with no importance measure is assigned the measure that makes the parent node's importance measure the average of those of its children (case starting with line~\ref{ln:propagateDC}). 
Suppose there are several (more than one) children nodes with no assigned importance measures, and all of those nodes do not have any descendants with assigned importance measures. In that case, we calculate the sum of those nodes so that the parent node's importance measure is the average of its children's measures. 
Then, we assume that the sum is divided equally amongst all the children nodes with no assigned importance measures (case starting with line~\ref{ln:propagateDA}). 
Finally, if there are several (more than one) children nodes without assigned importance measures, and at least one of those nodes does have a descendant with an assigned importance measure, then no propagation takes place (case starting with line~\ref{ln:nopropagate}). This is because those specific nodes require careful consideration to ensure their coherence with both their ancestor and descendant nodes. So, calculating their importance measure is postponed to another round.

Nothing must be done when propagating up if the node has no children nodes (line~\ref{ln:noProp2}). 
If all the children nodes have assigned importance measures, then the importance of the parent node becomes the average of its children nodes (case starting with line~\ref{ln:propagateUC}). 
Suppose some of the children nodes have an assigned importance measure, and some do not, and all of the latter's descendants have no descendants with assigned importance measures. In that case, there is no information from the descendants, and as such, we can only use the information from the children nodes with assigned importance measures. 
So, we calculate the average of the available measures for the children nodes and assign it to the parent node. 
Then we propagate downwards again by assigning that importance measure to those children nodes with no assigned importance (case starting with line~\ref{ln:propagateUA}). In all other cases, we take no action. 

The algorithm repeats until no new importance measures are propagated (line~\ref{ln:conditionRepeat}). 
Please note that Algorithm~\ref{alg:propagation} has been implemented in Prolog, and the code is available online.\footnote{The code is available at: \url{https://gitlab.iiia.csic.es/-/snippets/32}. %https://swish.swi-prolog.org/p/value_propagation2.pl}. 
It is also available to run online at: \url{https://swish.swi-prolog.org/p/valuePropagation.pl}. 
The \texttt{nodes}, \texttt{edges}, and \texttt{importance} predicates help define the value taxonomy. Running \texttt{?- propagate(X).} will return the importance of the remaining nodes, satisfying coherence. If there is any incoherence in the input values, the user is informed and propagation halts.}

In what follows, we present a sample value taxonomy of our running example, noting that the impact of property nodes and value importances on a working system is addressed in Section~\ref{sec:why}, which illustrates how value taxonomies can guide behaviour.

\subsubsection{The Running uHelp Example}\label{sec:ex-vtax}

The running example used throughout this section is uHelp. uHelp is a fully implemented app\footnote{\url{https://uhelpapp.com}} developed at IIIA-CSIC by a team that includes the first author of this paper. %It is available on Google Play\footnote{\url{https://play.google.com/store/apps/details?id=es.csic.iiia.uhelpapp}} and Apple Store.\footnote{\url{https://itunes.apple.com/es/app/uhelp/id1089461370}}
The app is a social networking app that was initially designed to support people in finding help within their social network with their day-to-day activities, such as finding someone to substitute another at work tomorrow or someone to lend chairs for a party~\cite{KosterMOSSSFJPG13,KosterMOSSSJFPG12,uhelpARXIV}. 

To support the reader's better understanding of our formal model described in this subsection, we will specifically consider the value {\it fairness}. 
%
%This value emerged as essential in the participatory design meetings we held with potential app users. 
%
As mentioned, we choose not to confine our model to predefined value theories, such as Schwartz's renowned basic universal human values~\cite{Schwartz2012AnOO}, but instead opt for values that emerge from different stakeholders. 
In our case, fairness arose as a candidate value through the participatory design meetings we held with potential app users. 

We note here that recognising the relevant high-level values (root nodes in the taxonomy) can be achieved either through a bottom-up approach, where the relevant values emerge from discussing the context with the appropriate stakeholders (as in the uHelp use case), or through a top-down approach, where a given stakeholder provides the relevant values based on prior knowledge (as in the medical use case domain with Hospital del Mar, Barcelona, where the four bio-ethical values are universally acknowledged within the field of medical ethics~\cite{Beauchamp1979-BEAPOB,Beauchamp2007-BEATFP} and adopted by the hospitals with which we have been interacting). 

\begin{figure}[!t]
    \centering
    \includegraphics[width=\linewidth]{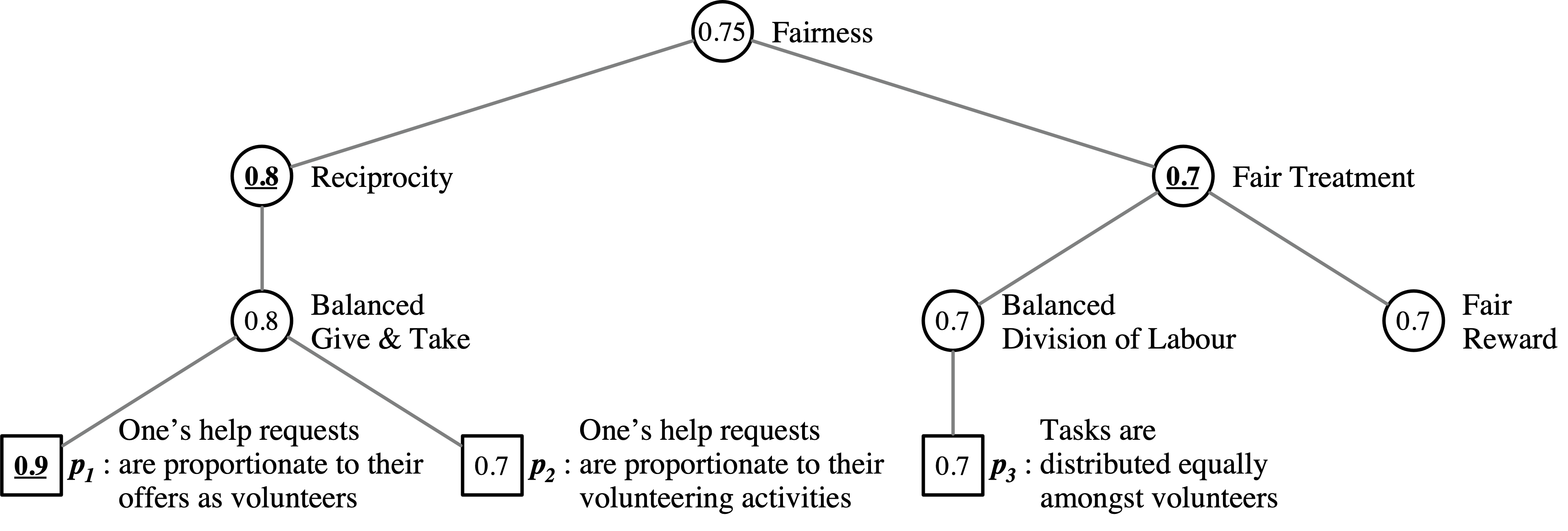}
    \caption{uHelp's value taxonomy for the value \emph{fairness}}
    \label{fig:valueTaxonomy}
\end{figure}

The value taxonomy then grows by defining how high-level values, like fairness, are understood. In uHelp, the community of single parents (Barcelona's Association of Monoparental Families) were the first to use the app. They viewed fairness as balanced give-and-take. In other words, they wanted members to refrain from acting greedily by constantly asking for help without offering it. When later incorporating a community supporting older people, fairness for them was interpreted as balanced division of labour among volunteers. This is because for them, some community members will mostly ask for help while others will mostly provide it. Balancing the workload among those providing help then becomes a key concern. 

This highlights how values are shaped by experiences and contexts, which is discussed in detail in Subsection~\ref{sec:context}. In this subsection, however, we focus on presenting the overall general taxonomy (Figure~\ref{fig:valueTaxonomy}) to illustrate how the higher-level value concepts can remain the same (such as fairness), while the specific nodes explaining them can vary, adopting different semantics (in this case, balanced give-and-take versus balanced division of labour). 

For illustrative purposes and to present a richer taxonomy, we have added the rightmost branch to Figure~\ref{fig:valueTaxonomy} that introduces fair reward as yet another aspect of fairness. This concept falls along with the balanced division of labour under the abstract concept of fair treatment. Fair rewards reflect the idea that not all volunteering tasks should be considered equal, as some are more demanding than others. However, since this value concept was never specified for uHelp, it lacks a property node that grounds it. %This illustrates how uHelp's value taxonomy for fairness is understood in terms of reciprocity and fair treatment. Reciprocity is understood as a balanced give-and-take, and fair treatment is understood in terms of a balanced division of labour as well as fitting rewards.

Next, we provide some of the grounding semantics of those abstract concepts, i.e., the property leaf nodes. Recall that property nodes are presented as squares, whereas label nodes are shown as circles.

For illustrative purposes, we provide two different computational approaches that define balanced give and take in terms of properties $p_1$ and $p_2$.
The first states that one's help requests are proportionate to the number of times the user offered help, whereas the second states that one's help requests are proportional to the number of times the user was chosen to help (because not all those offering help get selected). 
One straightforward approach to specifying $p_1$ and $p_2$ is through the ratio of offers/volunteering to requests being greater than 1, as illustrated in property definitions~\ref{eq:p1} and~\ref{eq:p2}. Of course, we provide simple properties for better readability. Still, we note that properties can become as complex as the language allows (remember that the choice of language for specifying properties is an implementation decision, and one may opt to work with highly expressive languages).
\begin{equation}\label{eq:p1}
    %p_1 \quad \eqdef \quad  \frac{\#_{\mathit{offers}}}{\#_{requests}} > 1
    p_1 \quad \eqdef \quad  \big( \#_{\mathit{offers}} \;/\;\#_{requests} \big) \;>\; 1
\end{equation}
\begin{equation}\label{eq:p2}
    %p_2 \quad \eqdef \quad  \frac{\#_{volunteering}}{\#_{requests}} > 1
    p_2 \quad \eqdef \quad  \big( \#_{volunteering} \;/\; \#_{requests} \big) \;>\; 1
\end{equation}
The two different properties for `balanced give and take' illustrate how different semantics may be provided for the same abstract value. In this specific case, one can imagine that $p_2$ was initially defined. However, some users are never chosen after some interactions despite volunteering. As such, the system prevents them from asking for help, ensuring that a balanced give and take is respected. To compensate for this, $p_1$ gets added to the taxonomy so that balanced give and take considers the offers for help instead of being chosen. %This illustrates how the taxonomy may evolve by learning from experience what the best grounding semantics (properties) of a given abstract concept are for a particular context. 
This example illustrates how the taxonomy can evolve by learning from experience to identify the most appropriate grounding semantics (properties) for a given abstract concept within a specific community (or for a specific collective).

The computational approach, or the property node, describing the balanced division of labour is implemented through property $p_3$, which states that tasks are equally distributed amongst volunteers. One way to specifying $p_3$ 
is by measuring the difference between the uniform distribution $U$ and the actual distribution $D$ of tasks assigned to each volunteer, where this difference should be smaller than a predefined threshold $\epsilon$ (property definition~\ref{eq:p3}).

\begin{equation}\label{eq:p3}
    p_3 \quad \eqdef \quad  \mathit{difference}(D,U) < \epsilon
\end{equation}
The difference between two distributions may be calculated using approaches like the Kullback–Leibler divergence~\cite{mackay2003information} or the earth mover's distance~\cite{emd}. 

This taxonomy (Figure~\ref{fig:valueTaxonomy}) does not specify the property node providing the grounding semantics for fair reward. Some abstract concepts may remain abstract for a while. 

We note that while stakeholders discuss and agree on abstract concepts of values, engineers must define the properties that ground the semantics of those values. Alternatively, AI can be used to support the construction and updating of these value semantics (and even taxonomies), such as by learning from past interactions. The construction of these taxonomies is one of the envisaged lines for future work, as discussed in Subsection~\ref{sec:valueIdentification}.

A crucial feature of the value taxonomy is the importance of nodes. %We envision each context having its own taxonomy with associated importance measures (see Subsection~\ref{sec:context}). As such, an adapted uHelp taxonomy for fairness will be made available for the community of single parents and another for the community supporting older people. However, whether importance measures are available for the general taxonomy of uHelp depends on the application. uHelp may or may not define importance measures for its general taxonomy for fairness. For illustration purposes, we assume that it does. 
Importance measures are generally tied to specific contexts, as we elaborate in Subsection~\ref{sec:context}, and may not always be defined for the general taxonomy. However, for the purpose of illustration, we assume they are provided for uHelp's general taxonomy (Figure~\ref{fig:valueTaxonomy}).
The focus in this subsection is on how such importance measures are obtained. 
Getting stakeholders to specify the importance of all nodes is usually challenging. If a partial order is provided, it might only cover some nodes. As such, we give an example where importance is assigned to a subset of these nodes. In this example (Figure~\ref{fig:valueTaxonomy}), we represent importance measures explicitly provided through bold underlined numbers. From those, our propagation mechanism (Algorithm~\ref{alg:propagation}) then calculates possible coherent importance measures for the remaining nodes, presented in regular font and non-underlined. 

Finally, we have provided in this subsection the value taxonomy of one uHelp value, fairness. We envision applications to have several relevant values and corresponding taxonomies, such as privacy and security (see Figure~\ref{fig:valueSystem}). %Finding a set of independent and relevant values has been consistent in our work with medical professionals and firefighters as well. %For example, in the medical domain, doctors at Hospital del Mar, Barcelona have already recognised a number of relevant values (as mentioned earlier).  
%We then expect concurrent processing over these various taxonomies. 

The impact of formalising values through taxonomies that define values in terms of property nodes and their importance is highlighted in Section~\ref{sec:why}, which focuses on connecting values with behaviour. Next, we discuss the issue of how values change with different contexts. 

\subsection{How do values change with context? Context-based value taxonomies}\label{sec:context}

Our stance is that values are context-dependent and evolve over time. This is the stance of value-sensitive design~\cite{vandePoel2018}, as well as the stance of many social psychologists before them~\cite{Rohan2000}.  
van de Poel~\cite{vandePoel2018} argues that new values may emerge, and the importance and definitions of old values may also change with time and context. We are aligned with this view. 
We believe that we all have a general understanding of what a value is, defined through its general taxonomy, and this understanding evolves with our experiences. When new values or new definitions emerge with experience, new nodes (label-based or property-based) and relations must be added to the general taxonomy. For example, if with experience, one learns that not only is `balanced division of labour' important but also `fair reward', then new nodes and relations must be added to reflect this. %Adding new nodes and relations to general taxonomies enables those taxonomies to evolve with experience, allowing what is learned from one experience to be applied in different contexts. 
This dynamic update process ensures that the general taxonomy remains relevant and applicable across different contexts, reflecting ongoing learning and adaptation.
For a specific context, we then suggest moving from a general taxonomy to a context-based one by simply revisiting the importance measures of the general taxonomy (whether they were defined in the general taxonomy or not), making some nodes (and eventually, branches) more prominent than others. 

In our proposal, general taxonomies can only grow by adding new nodes and relations. If there is a need to eliminate existing nodes, then this is currently achieved by setting their importance to zero (if zero is the neutral-based measure of the chosen codomain). 

A definition of a context-based value taxonomy is presented next, which states formally that a context-based taxonomy is an alteration of a general taxonomy where the importance measures are updated for the relevant context:

\begin{definition}[Context-based value taxonomy]\label{def:contextValueTaxonomy}
A context-based taxonomy $\mathcal{V}_{c}=\{N,E,I_{c}\}$ is an alteration of a general taxonomy $\mathcal{V}=\{N,E,I\}$ where the importance of nodes are updated for the given context. 
The importance of nodes in the context-based taxonomy $\mathcal{V}_{c}$ is independent of the importance of those nodes in the general taxonomy $\mathcal{V}$. 
The function calculating the importance of nodes with respect to a context $c$ is defined as $I_{c}: N \to COD_{\bot}$, where $COD_{\bot}$ defines the union of $COD$ with the undefined variable ($\bot$). 
\end{definition}
%This function assesses the importance of nodes with respect to a context $c$, where the context is defined through a set of properties $P_{c} \subseteq P$. The context $c$ is considered to hold ($holds(c)$) ---that is, we can say that we are in that context--- when its properties are satisfied: $(\forall p\in P_{c} \, \cdot \models p ) \implies \models holds(c)$.

We reiterate that if new nodes need to be introduced, they are introduced to the general taxonomy before the context-based one is constructed.

Lastly, we note that a context $c$ may be defined by a set of properties $P_{c} \subseteq P$. The context $c$ is then considered to hold ($holds(c)$) ---that is, we can say that we are in that context--- when its properties are satisfied: $(\forall p\in P_{c} \; (\models p) ) \implies \models holds(c)$. 
Although this paper does not investigate the ramifications of this context definition further, we include it to provide a high-level sense of how a system may approach detecting its current context in general.

\subsubsection{Implementation Choices} 
\paragraph{Constructing Context-Based Taxonomies} 

Different mechanisms for evaluating the importance of nodes in context-based taxonomies may be developed.
One implementation (Algorithm~\ref{alg:context}) employs a bottom-up approach, where only the importance of property nodes for the given context is evaluated (regardless of how they are obtained, whether they are provided manually by stakeholders or learnt from similar past experiences in similar contexts). The idea is that one can better assess the importance of concrete property nodes in specific contexts. 
Lines~\ref{ln:nodes1}--\ref{ln:nodes2} assign new importance measures to the property nodes, assuming those measures have already been provided for this context. %, regardless of how they are obtained. 
The mechanisms for obtaining these measures, specified through function \textsc{GetImportance}, are left for future work, as illustrated in Subsection~\ref{sec:valueIdentification}. Then, to calculate the importance of the remaining nodes in the taxonomy, line~\ref{ln:propagate} propagates importance measures to the rest of the taxonomy, following Algorithm~\ref{alg:propagation} of Subsection~\ref{sec:whatImp}.

%%%% ALGORITHM 2 WAS HERE
\begin{algorithm}[!b]
\caption{Constructing context-based value taxonomies}\label{alg:context}
\begin{algorithmic}[1]
%%%
\Require a general value taxonomy $\mathcal{V}=(N,E,I)$
\Require $N_{\phi} \subset N$ to be the set of property nodes in $N$
\Require a set of properties $P_{c}\in P$ that define the context $c$
\Require \textsc{GetImportance}($n$,$P_c$) to be a function that obtains the importance of node $n$ within context $c$ (the specification of this function is left for future work)
%%%
\Function{ContextTaxonomy}{$\mathcal{V}$,$P_{c}$}
    \State $I_{c}^{0} \leftarrow \emptyset;$
    \For{$n\in N_{\phi}$}\label{ln:nodes1}
        \State $I_{c}(n)$ $\leftarrow$ \textsc{GetImportance}($n$,$P_{c}$);\label{ln:getimportance} 
            \State $I_{c}^{0} \leftarrow I_{c}(n) \cup I_{c}^{0};$
    \EndFor\label{ln:nodes2}
    \State $I_{c} \leftarrow$ \textsc{Propagate}($N$, $E$, $I_{c}^{0}$);\label{ln:propagate}
        %\Comment{Propagate importance values (Algorithm~\ref{alg:propagation}).}
    \State \Return ($N$, $E$, $I_{c}^{0}\cup I_{c}$);
\EndFunction
\end{algorithmic}
\end{algorithm} 

Alternative implementations may be explored. 
Rather than starting with property nodes and following a bottom-up approach, a top-down approach may be implemented to assess abstract concepts regardless of their grounding semantics. Sometimes, a hybrid approach that combines bottom-up and top-down approaches might be most effective. %The domain usually suggests which approach is best. 
The choice of approach often depends on the characteristics of the domain in question.

Finally, we note that inconsistencies may arise between general and context-based taxonomies. This is expected and entirely normal. For example, while reciprocity may be considered very important as an abstract concept for the uHelp app, it may be regarded as less important in a specific context, such as the community of volunteers supporting older people (see Subsection~\ref{sec:howEG}).

\paragraph{Visualising Taxonomies} 
In addition to constructing context-based taxonomies by updating importance measures, and for improved visualisation, we propose an approach (Algorithm~\ref{alg:visualise}) that hides parts of the taxonomy deemed irrelevant for a given context. %This step is optional but may be helpful with growing taxonomies. This helps stakeholders quickly ascertain the relevant nodes for any given context.
This step is optional but can be particularly helpful as taxonomies grow, enabling stakeholders to quickly identify the value concepts most relevant to the context at hand.

%%%% ALGORITHM 3  HERE
\begin{algorithm}[!b]
\caption{Visualising value taxonomies}\label{alg:visualise}
\begin{algorithmic}[1]
%%%
\Require a context-based value taxonomy $\mathcal{V}_{c}=(N,E,I_{c})$
\Require $N_{\phi} \subset N$ to be the set of property nodes in $N$ 
%%%
\Function{VisualiseTaxonomy}{$\mathcal{V}_{c}$}
    \State $relevantNodes \leftarrow \emptyset;$
    \For{$n\in N_{\phi}$}\label{ln:nodes1b}
        \If{$I_{c}(n) \neq 0$}\label{ln:importance} %\Comment{We only consider property nodes with +ive importance.} 
            \State $relevantNodes \leftarrow \{n\} \cup relevantNodes;$
        \EndIf
    \EndFor\label{ln:nodes2b}
%%% Construct the context-based value taxonomy from the selected leaf nodes
    \State $N_{v} \leftarrow relevantNodes;$
    \State $E_{v} \leftarrow \emptyset;$
    \State $I_{v} \leftarrow \{I(n) \; | \; n \in relevantNodes\};$
%    \State $N_{v}^{0} \leftarrow \emptyset;$
    \Do\label{ln:branches1}
        %\Comment{This loop selects the branches leading to the selected nodes.}
        \State $N_{v}' \leftarrow N_{v};$
        \For{$n \in N_{v}$}
            \If{$(p,n) \in E \wedge p \not\in N_{v}$}
                \State $N_{v} \leftarrow {p} \cup N_{v};$
                \State $E_{v} \leftarrow {(p,n)} \cup E_{v};$
                \State $I_{v} \leftarrow I(p) \cup I_{v};$
            \EndIf
        \EndFor
    \doWhile{$N_{v}' \neq N_{v}$}\label{ln:branches2}
%%%
    \State \Return ($N_{v}$, $E_{v}$, $I_{v}$);
\EndFunction
\end{algorithmic}
\end{algorithm}

Different approaches may be followed when deciding which branches are relevant and may be visualised. Algorithm~\ref{alg:visualise}'s strategy uses the importance of property nodes to determine the relevant branches of the taxonomy. The algorithm maintains the branches that lead to relevant property nodes and eliminates those that lead to irrelevant ones.  

Again, different techniques may be followed when deciding which property nodes are relevant. In Algorithm~\ref{alg:visualise}, the irrelevant nodes are those with an importance of zero (lines \ref{ln:nodes1b}--\ref{ln:nodes2b}). 
But based on the application, one can imagine alternative implementations. 
For example, it could be decided that only property nodes with a positive importance are relevant. 
(It would also be possible to set another predefined threshold, other than $0$, to filter relevant importance values.) 
An alternative approach would be to use clustering algorithms (such as the $k$-means~\cite{rai2010survey}) on the importance measures of all property nodes. This helps cluster the nodes into two sets, the important and non-important nodes, without the need for a pre-defined threshold. These are just examples of possible strategies for deciding on relevant property nodes.

Going back to Algorithm~\ref{alg:visualise}, after selecting the relevant property nodes, the algorithm visualises the prominent branches of the taxonomy in lines~\ref{ln:branches1}--\ref{ln:branches2} by keeping only the branches that lead to those relevant nodes. 

%Finally, we note that if property-based leaf nodes did not exist for a prominent branch, then this implies that a relevant value concept is missing its computational semantics. A flag must then be raised for adding a corresponding property-based leaf node to the general taxonomy, because otherwise, a computational approach considering that prominent branch will not be possible. 
Finally, we note that if a prominent branch lacks property-based leaf nodes, this indicates that a relevant value concept is missing its computational grounding. This should trigger a flag to add the necessary property-based leaf node to the general taxonomy; otherwise, that branch cannot be computationally evaluated.

\subsubsection{The Running uHelp Example}\label{sec:howEG}

\begin{figure}[t]
    \centering
    \begin{subfigure}[b]{\linewidth}
         \centering
         \includegraphics[height=11em]{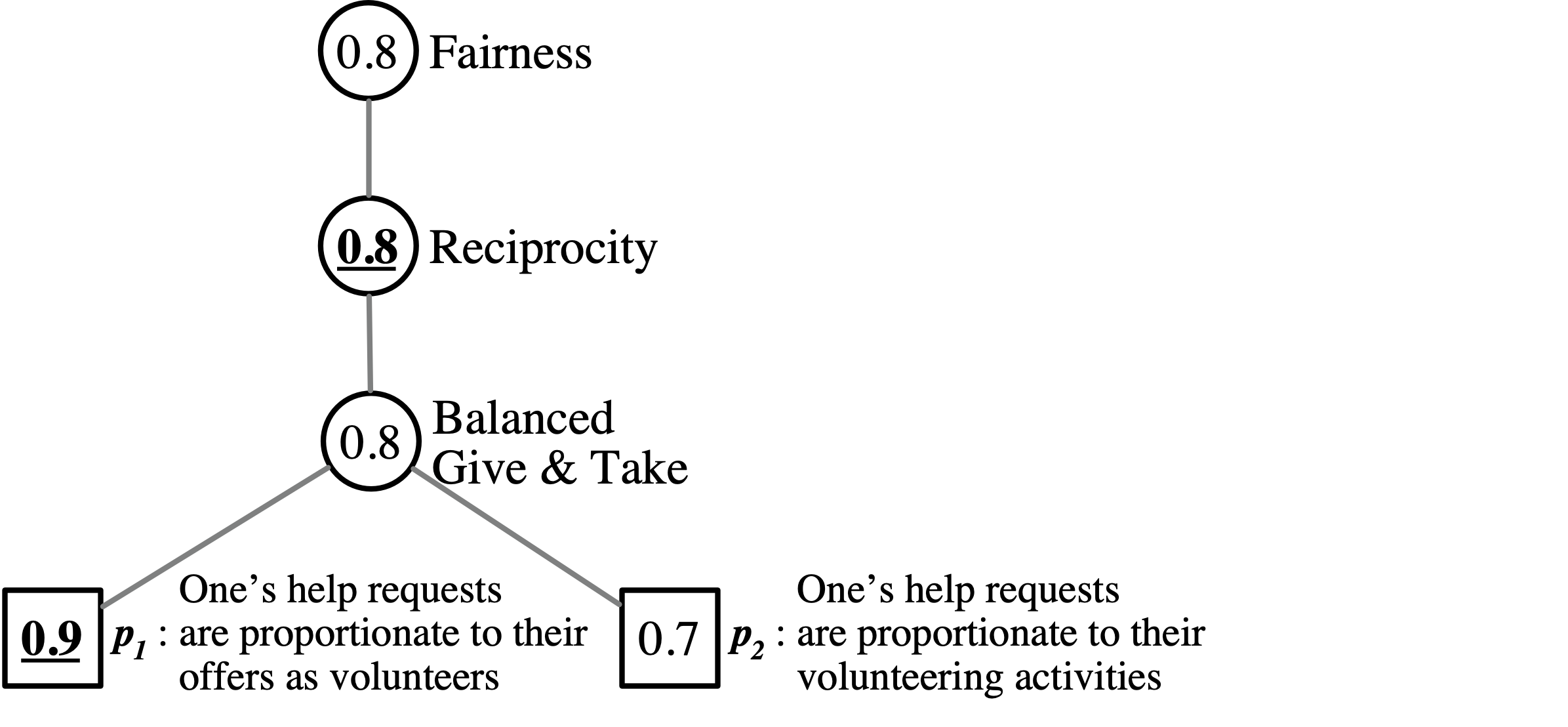}
         \caption{$\mathcal{V}_{c_{s}}$ for the community of single parents, which happens to be the same as the initial general uHelp taxonomy $\mathcal{V}$}
         \label{fig:single-mothers-1}
     \end{subfigure}
     
     %\hfill
     \vspace{1em}
     \begin{subfigure}[b]{\linewidth}
         \centering
         \includegraphics[height=11em]{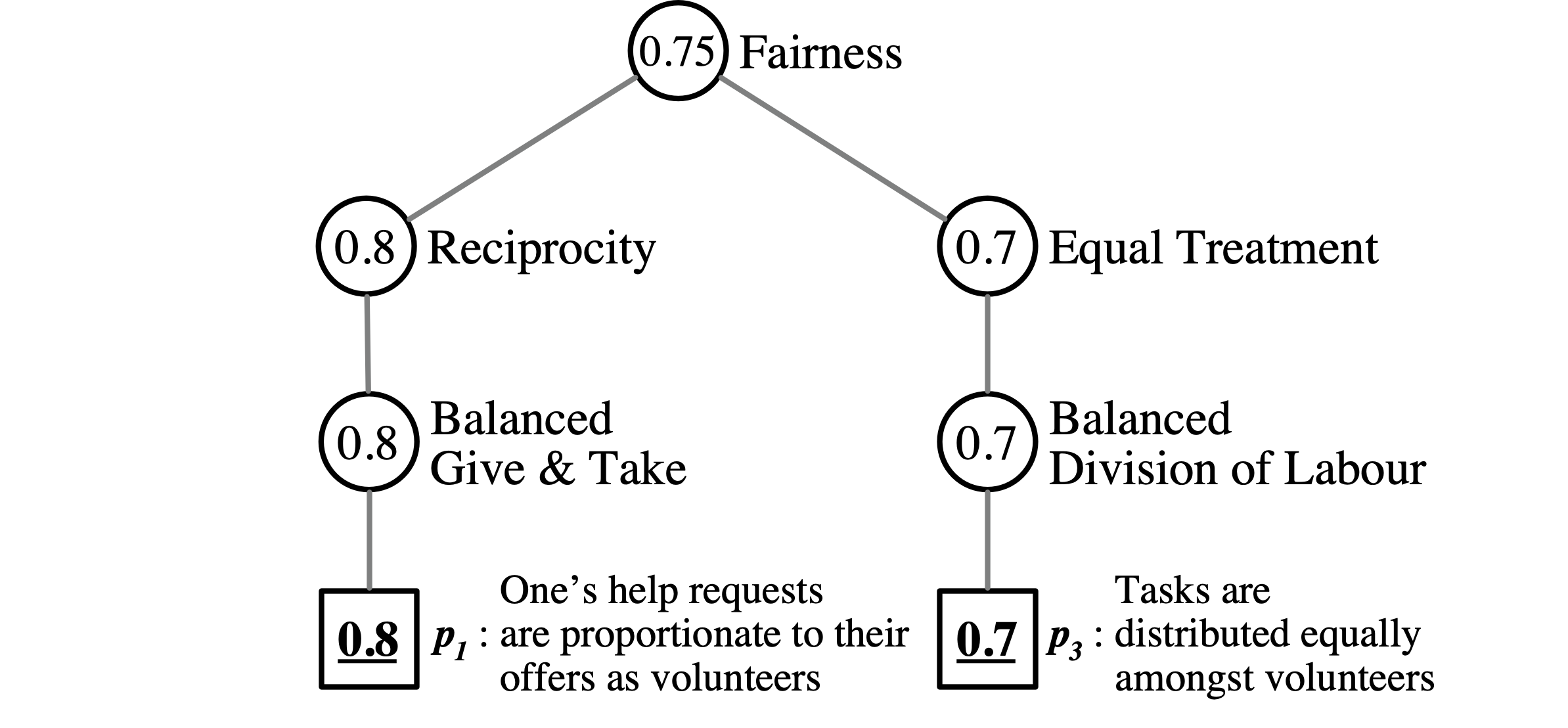}
         \caption{$\mathcal{V}_{c_{s}}'$ for the updated context-based taxonomy for the community of single parents}
         \label{fig:single-mothers-2}
     \end{subfigure}
     
     %\hfill
     \vspace{1em}
     \begin{subfigure}[b]{\linewidth}
         \centering
         \includegraphics[height=11em]{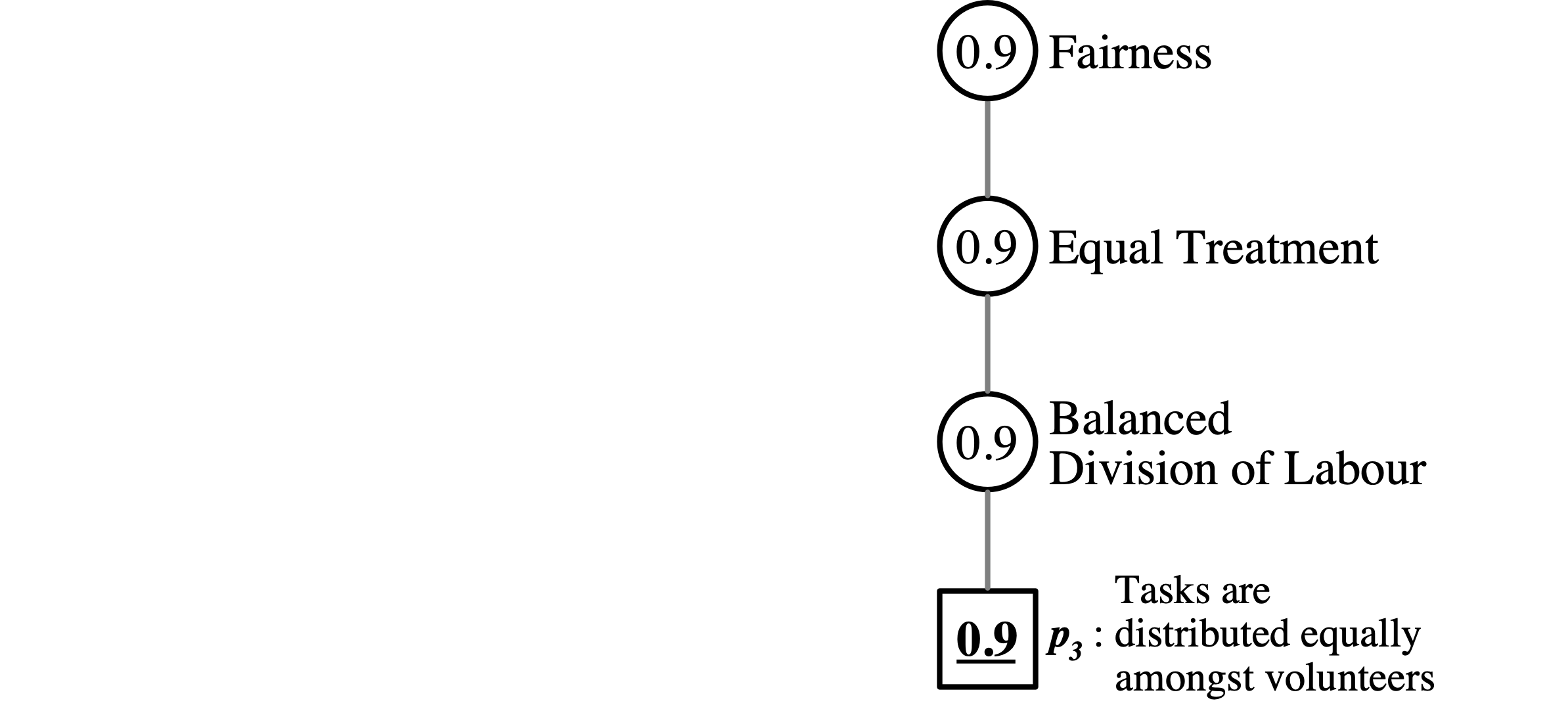}
         \caption{$\mathcal{V}_{c_{e}}$ for a community of volunteers supporting older people}
         \label{fig:elderly}
     \end{subfigure}
    \caption{Different context-based value taxonomies for fairness in uHelp}
    \label{fig:contextValueTaxonomy}
\end{figure}

Figure~\ref{fig:valueTaxonomy} presented uHelp's general taxonomy for fairness. 
This subsection presents examples of context-based taxonomies for different uHelp communities. 
As illustrated in Subsection~\ref{sec:ex-vtax}, the first community to adopt uHelp was the community of single parents (whose context is specified as $c_{s}$). For them, fairness implied a balanced give-and-take. We can imagine that at the beginning, both the general and context-based taxonomy for the $c_{s}$ community were defined through the taxonomy of Figure~\ref{fig:single-mothers-1}. 
But as a new context is considered, which is that of the community supporting older people (whose context is specified as $c_{e}$), new nodes were added to the general taxonomy, resulting in (say) the taxonomy of Figure~\ref{fig:valueTaxonomy}. 

With this new general taxonomy for fairness, the community supporting older people decides that the requirement of having a proportionate number of help requests with respect to the number of help offers ($p_1$) or actual volunteering ($p_2$) is considered irrelevant. In contrast, the requirement that labour should be divided in a balanced way amongst all volunteers is of utmost importance. As such, the importance of the different property nodes of  Figure~\ref{fig:valueTaxonomy} is specified accordingly:
$$\begin{array}{ccccc}
I_{c_{e}}(p_{1}) = 0 & ; &
I_{c_{e}}(p_{2}) = 0 & ; &
I_{c_{e}}(p_{3}) = 0.9 %\phantom{-}0.9
\end{array}$$

The context-based taxonomy of this new community $\mathcal{V}_{c_{e}}$ is constructed by Algorithm~\ref{alg:context}, and is visualised by Algorithm~\ref{alg:visualise}, which eliminates all branches leading to property nodes with zero importance, leaving us with one single branch leading to property node $p_3$. The resulting taxonomy is presented in Figure~\ref{fig:elderly}. The importance of the upper nodes is inherited from the importance of the property node, following Algorithm~\ref{alg:propagation}. 

Now assume that with this new general taxonomy (Figure~\ref{fig:valueTaxonomy}), the community of single parents decides to revisit their context-based taxonomy. They have become interested in the new grounding semantics that have emerged and decided that having requests proportionate to offers ($p_1$) and equally distributed amongst volunteers ($p_3$) are both essential properties. 
This new view is reflected in the importance measures they assign to the property nodes of Figure~\ref{fig:valueTaxonomy}:
$$\begin{array}{ccccc}
I_{c}(p_{1}) = 0.8 & ; &
I_{c}(p_{2}) = 0\phantom{.0} & ; &
I_{c}(p_{3}) = 0.7
\end{array}$$

Algorithm~\ref{alg:context} builds the new taxonomy $\mathcal{V}_{c_{s}}'$. Again, for visual clarity, Algorithm~\ref{alg:visualise} eliminates the branches that lead to property nodes with zero importance. The result is the taxonomy presented in Figure~\ref{fig:single-mothers-2}. As before, the importance of the upper nodes is calculated following Algorithm~\ref{alg:propagation}. 

\subsection{Who holds values? Individuals versus collectives}\label{sec:who}
Our stance is that both individual and collective entities hold values; values do not exist in isolation. In other words, there is no universal taxonomy of values for fairness. From now on, we use the term `entities' to refer to either individuals (human or artificial) or collectives. 
Different entities will hold varying views on values, resulting in distinct taxonomies. We use the notation $\mathcal{V}^{x}$ to represent the value taxonomy held by $x$, where $x$ may represent an individual $i_j$ or a collective $\{i_1,\ldots, i_n\}$. We use the term `collective' to describe a group of interacting individuals (usually referred to as members), which may be a community, an organisation, an institute, a society, or a culture. 
When a collective holds a value taxonomy, this is understood as the value taxonomy describing the values of the collective as a whole, and it usually emerges from the values of the individuals, even though % and not its individuals (or interacting members). 
not all individuals may have their value taxonomy aligned with the collective's.
The issue of how the collective specifies (or agrees on) its taxonomy is left for future work, as illustrated in our proposed roadmap in Subsection~\ref{sec:roadmapVagreement}. For example, preliminary existing work~\cite{LeraLeriLBJLMRS24} uses computational social choice to help compute the collective's view on values by aggregating the views of the individuals.  
To simplify notation, we drop the $x$ from $\mathcal{V}^{x}$ when it is clear who holds the taxonomy.

Humans and artificial agents not only understand their own values or the values of the collectives to which they belong, but they also observe others and form beliefs about the values of these others. 
We use the notation $\mathcal{V}^{x>y}$ to represent what $x$ believes to be the values of $y$, where each of $x$ and $y$ may represent an individual or a collective. 

In summary, we propose how individual, collective, and inferred values (where the latter represents beliefs about the values of others, individual or collective) can be expressed. This provides the basis for reasoning with values on different levels. For example, in some contexts, an agent (human or artificial) might make decisions driven by its own values. In other contexts, it might be driven by collective values, such as medical professionals following their organisation's values. Sometimes, it is essential to consider the values of others with whom one is interacting, leading to a value-enriched theory of mind reasoning. For example, modelling and reasoning about the individual and collective values of countries can be helpful with climate policy negotiations, as this could help clarify priorities, highlight value conflicts, and help identify common ground. We remind the reader that this paper focuses on value representation, whereas reasoning with values is the subject of ongoing and future work (see Section~\ref{sec:roadmap}).

\paragraph{Value Systems} Finally, we note that entities typically hold multiple value taxonomies. When the relations defined in our taxonomy can be established between at least one node from each taxonomy, these taxonomies may join into a unified structure. However, we do not enforce such connections to exist.  
As illustrated by Schwartz's theory of universal values, some values may be inherently opposed, resulting in taxonomies that remain disjoint. We recognise that other types of relationships between value concepts (such as adjacency/ similarity, opposition, and orthogonality) extend beyond those captured in our current model. These additional relations, inspired by Schwartz, can connect nodes across disjoint taxonomies. Exploring these cross-taxonomy relationships is an important direction for future research.  

\begin{figure}[!ht]
\centering
\includegraphics[width=\textwidth]{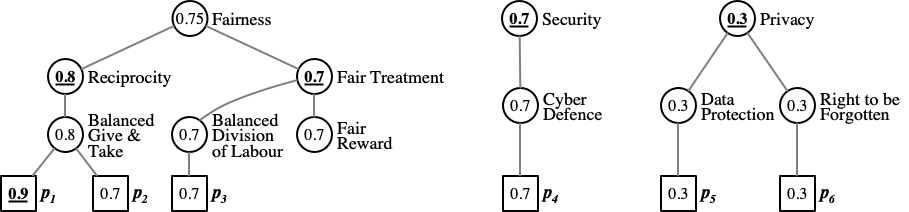}
\caption{A value system is a set of value taxonomies}
\label{fig:valueSystem}
\end{figure}

We refer to the entire set of value taxonomies that an entity holds as the entity's \emph{value system}. Figure~\ref{fig:valueSystem} presents a sample value system with taxonomies defining three central independent values: fairness, security, and privacy.

\subsubsection{Implementation Choices}\label{sec:whoImp}
The issue at stake here is how to implement mechanisms that take into account the value taxonomies of a set of individuals and compute the taxonomy of the collective. This is a complex task, and the domain may dictate how collective values are specified. 
For example, in some cases, a company developing an assistive robot may pre-define the value taxonomy governing the robot's behaviour. 
In another case, such as a hospital, a board of elected medical personnel will convene to collectively agree on the important values of their hospital and how they are lived out in practice. 
Yet, in other cases, such as with the uHelp application, one can imagine the entire community of users coming together to vote on their values. %In other words, the rules dictating whose view should be considered and how, when specifying the value taxonomies of a collective, are domain-dependent and left for future development (see Subsection~\ref{sec:roadmapVagreement}). 
Thus, the rules governing whose views count and how they are combined when specifying a collective's value taxonomy are domain-dependent and remain an open question (see Subsection~\ref{sec:roadmapVagreement})

Depending on the domain and governance structure, different mechanisms may be developed to construct collective values from individual ones. 
For instance, negotiation and argumentation mechanisms can support individuals reach collective agreements on their values. Alternatively, computational social choice may be employed to aggregate individual values into collective ones. We point the interested reader to the work by Lera-Leri et al.~\cite{Lera-LeriBSLR22}, where the aggregation takes into consideration various ethical principles, such as utilitarian (maximum utility) or egalitarian (maximum fairness). 

It is important to note that the relationship between individual and collective values is complex. How collective values emerge, and which values (individual or collective) take precedence, often depends on the domain. For example, in traditional organisations such as hospitals and fire departments, members are generally expected to adhere to organisational values. In contrast, in communities like the uHelp social network, members are expected to follow their own individual values.
%Finally, we note that the relation between individual and collective values is a complex one. How they emerge, and which values (individual or collective) must be prioritised, is usually domain-dependent. For example, in some traditional organisations, such as hospitals and fire departments, members (medical professionals and firefighters) are expected to adhere to the values of their organisations. In other cases, like the uHelp social network, members are expected each to follow his/her own values. 
%
Section~\ref{sec:why} shortly presents %the alignment of an entity with certain values, assuming the relevant values are identified. 
how an entity aligns with certain values, assuming those relevant values have been identified. 

\begin{figure}[!b]
    \centering
    \begin{subfigure}[b]{.45\linewidth}
         \centering
         \includegraphics[width=\linewidth]{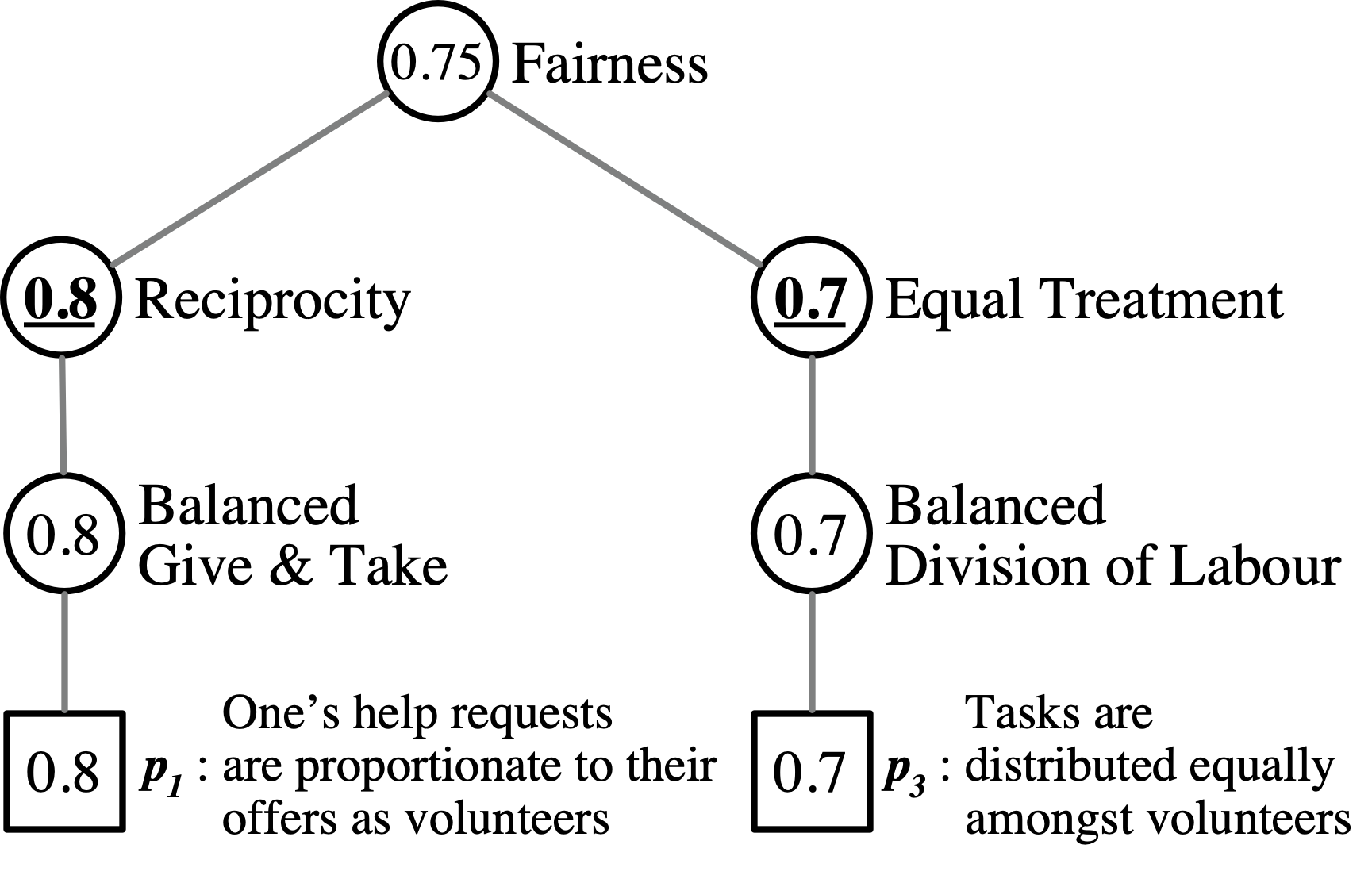}
         \caption{$\mathcal{V}_{c_{s}}^{user^{1}}$}
         \label{fig:user1}
     \end{subfigure}
     \hfill
     \begin{subfigure}[b]{.45\linewidth}
         \centering
         \includegraphics[width=\linewidth]{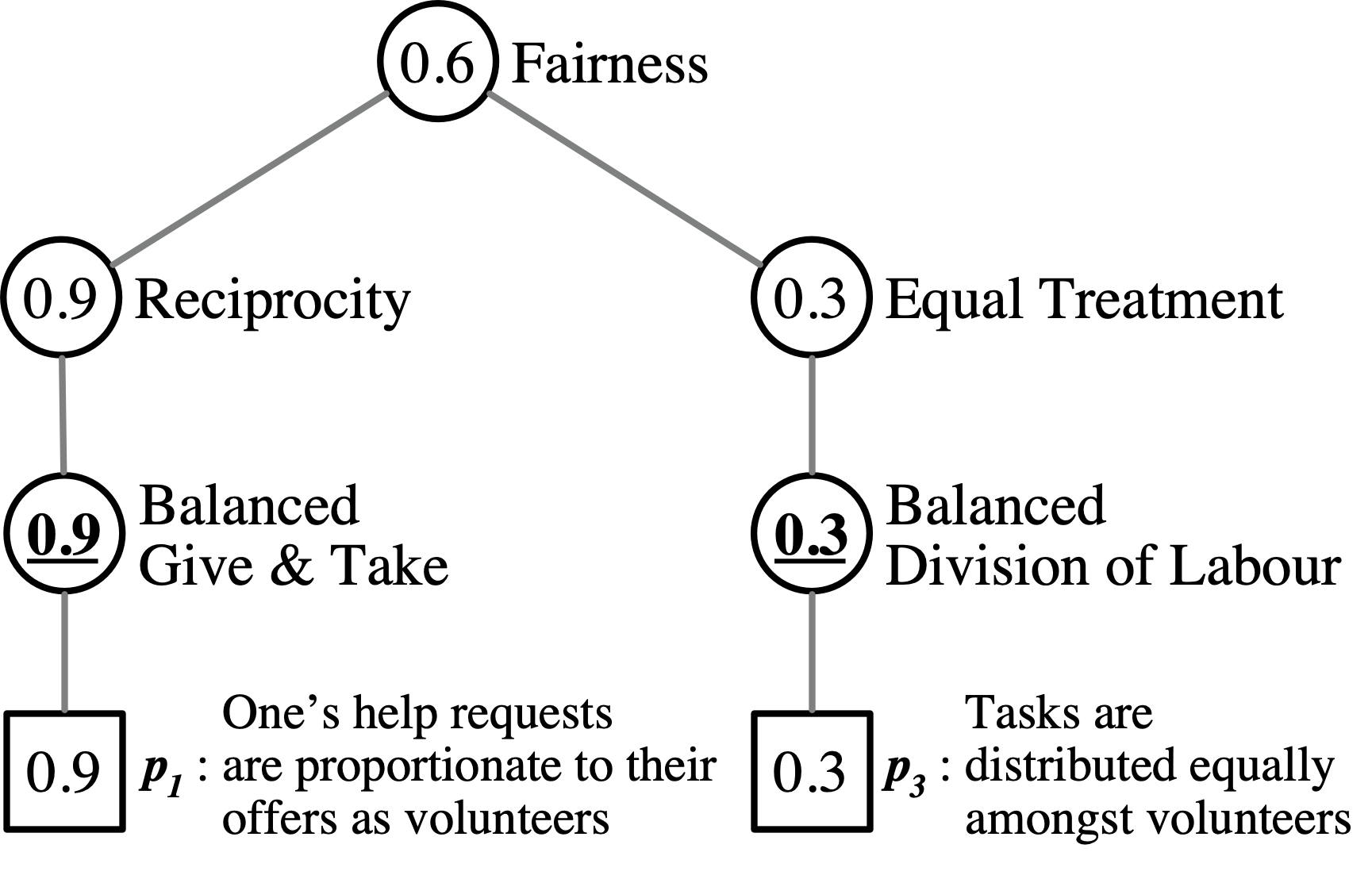}
         \caption{$\mathcal{V}_{c_{s}}^{user^{2}}$}
         \label{fig:user2}
     \end{subfigure}

     \vspace{1em}
     \begin{subfigure}[b]{.45\linewidth}
         \centering
         \includegraphics[width=\linewidth]{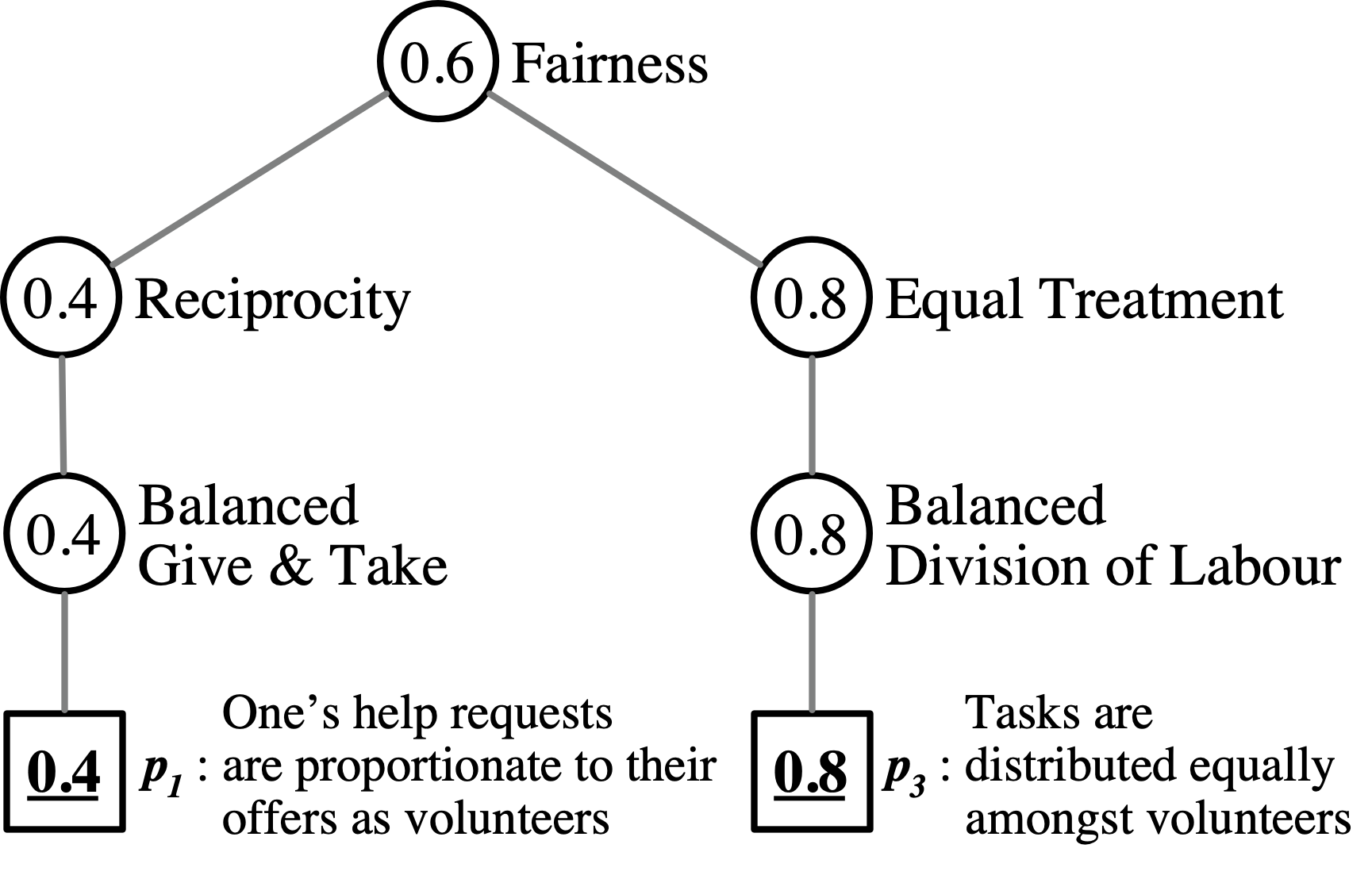}
         \caption{$\mathcal{V}_{c_{e}}^{user^{3}}$}
         \label{fig:user3}
     \end{subfigure}
     \hfill
     \begin{subfigure}[b]{.45\linewidth}
         \centering
         \includegraphics[width=\linewidth]{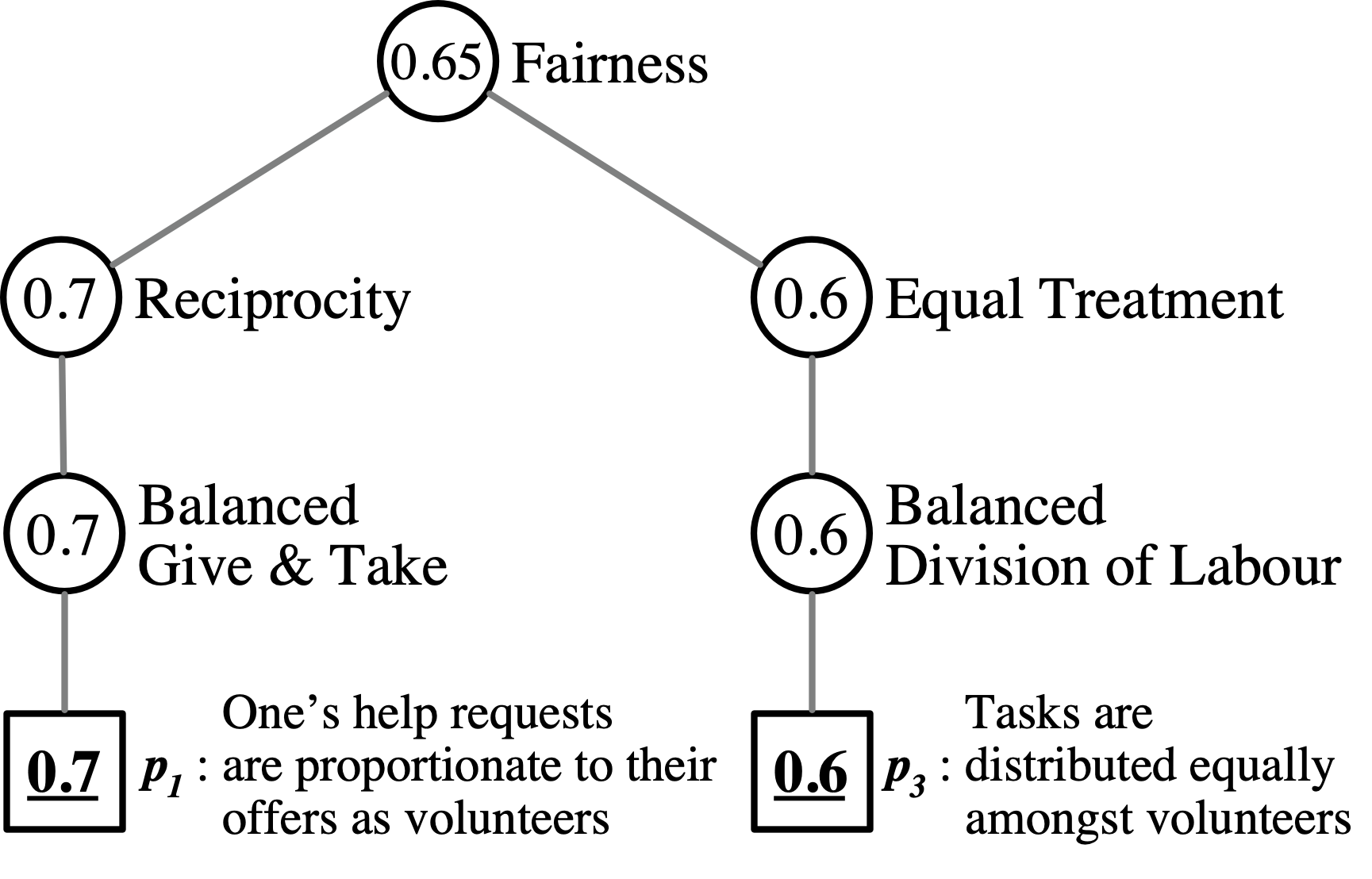}
         \caption{$\mathcal{V}_{c_{e}}^{collective}$}
         \label{fig:group}
     \end{subfigure}
    \caption{Individual and collective value taxonomies for fairness in uHelp's community of single parents}
    \label{fig:individualVcollective}
\end{figure}

\subsubsection{The Running uHelp Example}
This subsection highlights how both individuals and collectives can hold value taxonomies. 
For example, consider three uHelp members from the community of single parents, who hold the taxonomies depicted in Figures \ref{fig:user1}--\ref{fig:user3}. Of course, more members would exist in reality, each with their own taxonomy, but for simplicity, we only show the taxonomies of three individuals. 
Given these taxonomies, the social choice mechanism of Lera-Leri et al.~\cite{Lera-LeriBSLR22} can then be used to compute the importance of each property node for the collective. The propagation mechanism of Algorithm~\ref{alg:propagation} can propagate those aggregated measures to the rest of the collective's taxonomy. The resulting and final taxonomy is shown in Figure~\ref{fig:group}.

\subsection{Why hold values? The value alignment problem}\label{sec:why}
As presented in Section~\ref{sec:whyValues}, it is well established that values are among the primary motivators of behaviour. 
The main objective of the work on values in AI has been to assess and promote value-aligned behaviour. % by assessing how well behaviour aligns with specified values. %the alignment of behaviour with given values. 
Alignment can be evaluated with respect to any selected value concept. When a particular value node is under inspection, alignment is computed against its corresponding sub-taxonomy.

Recall that the property-based nodes of a value taxonomy introduce the foundations for linking abstract value concepts to concrete computational constructs, enabling formal assessment of behavioural alignment with the values in question. 
An entity's value alignment is thus measured by the degree to which its behaviour satisfies the relevant property-based value nodes. %In Definition~\ref{def:contextValueTaxonomy}, we have seen how importance is assigned to different nodes of a context-based taxonomy. Since alignment is typically assessed within specific contexts, such assessments are expected to rely on context-based rather than general taxonomies.
The evaluation of alignment should also account for the following: the more important a property-based node is, the more its satisfaction contributes to the overall alignment of the evaluated behaviour, and vice versa.
%The evaluation of value alignment should consider these essential measures: the more important a property-based node is, the higher its satisfaction contributes to the value alignment of the evaluated behaviour, and vice versa. 
%The alignment of an entity $e$'s behaviour with a context-based value taxonomy $\mathcal{V}_{c}$ is then defined accordingly, which formally states that the alignment of $e$'s behaviour with the value taxonomy $\mathcal{V}_{c}$ is an aggregation of the satisfaction of all property nodes of the taxonomy, taking into consideration the importance of each property node. We note, however, that context-based value taxonomies are expected to describe the values held by some entity. For the sake of simplifying notation, we drop the holder $x$ (and possibly $x$'s view of $y$'s values, if that was the case) and replace $\mathcal{V}^{x}_{c}$ with $\mathcal{V}_{c}$ (or $\mathcal{V}^{x>y}_{c}$ with $\mathcal{V}_{c}$). 
The alignment of an entity $e$'s behaviour with a context-based value taxonomy $\mathcal{V}_{c}$\footnote{Since alignment is typically assessed within specific contexts, such assessments are expected to rely on context-based rather than general taxonomies.} is then defined as the aggregation of the satisfaction degree of all property-based nodes in $\mathcal{V}_{c}$, weighted by their respective importance. 
We note that context-based value taxonomies are intended to represent the values held by a particular entity. To simplify notation, we omit the holder $x$ (or, sometimes, $x$'s view of $y$'s values) and write $\mathcal{V}_{c}$ instead of $\mathcal{V}^{x}_{c}$ (or $\mathcal{V}^{x>y}_{c}$).
We also note that $x$ (or $y$) does not necessarily have to be the same entity $e$ whose behaviour is being assessed. In other words, if $x\neq e$ (or $y\neq e$), then this describes the process of assessing how much $e$ is aligned with the values of $x$ (or $y$).
\begin{equation}\label{eq:alignment}
    \mathcal{A}(e,\mathcal{V}_{c}) = \bigoplus_{p\in N_{\phi,c}}  f(sd(p,e),I_{c}(p))
\end{equation}
where $N_{\phi,c}$ represents the property nodes of the taxonomy $\mathcal{V}_{c}$, $sd(p,e)$ represents the degree of satisfaction of property $p$ with respect to the behaviour of entity $e$, and $I_{c}(p)$ represents the importance of the property-node $p$ within the context-based value taxonomy $\mathcal{V}_{c}$. The function $f$ is used to factor in the importance of property nodes when considering their degree of satisfaction, whereas $\bigoplus$ is used to aggregate the degree of satisfaction of all property nodes in $\mathcal{V}_{c}$. % (with value importance factored in). % (i.e. $N_{\phi,c}$).

With Equation~\ref{eq:alignment}, we provide the basis for calculating value alignment, which supports value-based reasoning and decision making. When low alignment is detected, we believe this should act as a trigger for changing behaviour, for instance, by resulting in a change in the norms that mediate behaviour in multi-agent systems or a change in the agent's decision-making process, so that better alignment can be achieved. For example, if fairness is not being respected in the uHelp app, then the app's norms should be modified to ensure fairness is upheld. This is illustrated shortly in our running example in Section~\ref{sec:alignmentImpactExample}. 
As this paper focuses on the computational model for formalising values, the complex issue of how alignment is used to change behaviour is discussed in our roadmap in Section~\ref{sec:roadmap}. %But as examples of real-life systems, we note that in the hospital setting, we are designing tools that would provide feedback to medical professionals on the alignment of different actions to support their decision making. We are also developing tools that suggest how medical protocols can be modified to improve alignment with one value or another when misalignment is detected~\cite{RodriguezSoto2024VALE,RodriguezSoto2024AWAI}. 
As real-life illustrations, we note that in the hospital setting we are developing tools that provide feedback to medical professionals on the alignment of their actions with institutional values, supporting informed decision making. 
We are also developing tools that suggest how medical protocols can be modified to improve alignment with one value or another when misalignment is detected~\cite{RodriguezSoto2024VALE,RodriguezSoto2024AWAI}. 
%2026, added: 
This allows value alignment to be used not only to assess individual outcomes, but also to compare alternative courses of action.

%2026, added:
\paragraph{Value-aware deliberation} From an agent perspective, the value alignment mechanism described above can be interpreted as part of the deliberation process in a multi-agent system. Given a set of candidate actions, an agent can evaluate the potential outcomes of these actions with respect to the properties grounding the relevant values, and aggregate these evaluations according to the importance assigned to each value in the taxonomy. This enables agents to compare alternative actions in terms of their degree of value alignment, particularly in situations where different values favour different outcomes. While we do not define a complete decision-making procedure, the proposed model provides the representational foundations required to incorporate values into agent reasoning and action selection.

\subsubsection{Implementation Choices} 
One issue is how the satisfaction of property nodes $sd(p,e)$ is calculated. 
In other words, given an entity $e$, how do we assess to what degree the behaviour of $e$ results in the satisfaction of property $p$? This requires knowledge about how $e$ behaves, and different implementation approaches for specifying this knowledge can be followed. 
For example, suppose $e$ is a complex system of communicating entities. In that case, $e$'s model (usually specified via a process calculi) will describe its behaviour through a labelled transition system where the satisfaction of specific properties at different states can be evaluated~\cite{Stirling2001}. 
If $e$ is a normative system, then the norms can help map out the state diagram of the possible interaction outcomes, enabling evaluating the satisfaction of relevant properties~\cite{MontesOS22,aagotnes2007logic,cranefield2011verifying}. 
If $e$ was an agent with a BDI model, then BDI reasoning mechanisms can help assess the degree to which specific properties will be satisfied by $e$'s behaviour~\cite{rao1998decision}. 

In summary, a model of $e$ describing its behaviour is necessary to assess $sd(p,e)$, regardless of which approach is used to model $e$. %As illustrated above, this issue has already been addressed in many fields. 
To ensure our proposal is not limited to one modelling choice, we omit the choice of modelling $e$'s behaviour and assume the degree of satisfaction $sd(p,e)$ to be attainable. 

Returning to the alignment function $\mathcal{A}$, there are other implementation choices to be made, such as deciding on the function $f$ that factors in the importance of a property node and the aggregation operator $\bigoplus$. 
For illustration purposes, we propose in this paper a straightforward implementation that follows a weighted arithmetic mean approach that uses the importance of properties as the weight of their degree of satisfaction (so $f$ is implemented as a multiplication operator) and aggregates over all relevant properties ($\bigoplus$):

\begin{equation}\label{eq:alignmentImp}
    \mathcal{A}(e,\mathcal{V}_{c})= \frac{\displaystyle\sum_{p\in N_{\phi,c}}  I_{c}(p) \cdot sd(e,p)}{\displaystyle\sum_{p\in N_{\phi,c}}  I_{c}(p)}%{|N_{\phi,c}|}
\end{equation}

If we assume the range of value importance $I_{c}$ to be $[-1,1]$, and the degree of satisfaction $sd(e,p)$ to be a percentage within the range $[0,1]$, then the range of $\mathcal{A}$ becomes $[-1,1]$ where negative results describe the degree of misalignment %(or an alignment with detested values) 
and positive results illustrate the degree of alignment with aspired values. 

%For simplicity and clarity, we choose a weighted arithmetic mean for illustration purposes. 
Alternative and more sophisticated approaches to alignment can be explored. For example, Equation~\ref{eq:alignmentImp} can be improved by increasing its robustness to outliers or by designing the alignment score to grow with the number of satisfied property nodes. Additionally, the equation can be refined to reflect a more nuanced conceptual analysis of what constitutes values with higher weights. 
For instance, one might consider not only the importance of a property when assessing satisfaction, but also its structural weight within the value taxonomy, such as the number of distinct paths leading to the property node. 
In other words, the larger the number of paths that lead to a property node, the larger its impact on alignment. %: that is, replacing $I(p)\cdot sd(e,p)$ in Equation~\ref{eq:alignmentImp} with $paths(p) \cdot I(p)\cdot sd(e,p)$, where $paths(p)$ is the number of paths in the value taxonomy that lead to the property node $p$. 

\subsubsection{The Running uHelp Example}\label{sec:alignmentImpactExample} 
Let us return to our uHelp example and examine the context-based value taxonomy $\mathcal{V}_{c_{s}}'$ for a mutual aid community represented by Figure~\ref{fig:single-mothers-2}. 
The concrete definitions of properties $p_1$ and $p_3$ (property definitions \ref{eq:p1} and~\ref{eq:p3}) illustrate what it means, computationally, for the behaviour of some entity to be aligned with the value `fairness' in this context. 
Next, we illustrate how the exact degree of satisfaction of properties 
$p_1$ and $p_3$ can be computed according to these definitions. While the property definition~\ref{eq:p1} of $p_1$ stated that the number of offers must be higher than the number of requests, we now propose an approach to calculate the degree of satisfaction, where the larger the number of offers, the larger the degree of satisfaction.  
Equation~\ref{eq:p1Sat} formally states that the degree of satisfaction of $p_1$ is the actual ratio of requests to offers, normalised to fall into the range $[-1,1]$.
\begin{equation}\label{eq:p1Sat}
    sd(e,p_1)= 
    \begin{cases}
        \displaystyle\frac{(R -1)}{(\max R) - 1}
            &  \text{, if } R > 1 \\[1em]
        R -1
            & \text{, otherwise}
    \end{cases}
\end{equation}
where $R=\# \mathit{offers}/\# requests$ represents the ratio of offers to requests, and $\max R$ is the maximum possible value for $R$. While the range of $R$ is $[0,\infty)$, a maximum value must be selected for our equations. We argue that $\max R$ is domain-specific and should be selected for each context.  

Equation~\ref{eq:p1Sat} states that the degree of satisfaction is computed by mapping the ratio $R$ of offers to requests to the range $[-1,1]$. 
When this ratio is in the range $[1,\max R]$, then this gets normalised to the range $[0,1]$ to describe a positive degree of satisfaction (where $1$ gets mapped to $0$ and $\max R$ gets mapped to $1$). 
And when the ratio is in the range $[0,1]$, then this gets translated to the range $[-1,0]$ to describe a negative degree of satisfaction (where $0$ gets mapped to $-1$ and $1$ gets mapped to $0$).

In summary, $ p_1$'s degree of satisfaction depends on how far the ratio $R$ is from $1$. The larger it is relative to $1$, the higher the satisfaction. The closer it is to $0$, the higher the dissatisfaction.

Next, Equation~\ref{eq:p3Sat} defines the satisfaction of property $p_3$ similarly by formally stating that the degree of satisfaction of $p_3$ is the actual difference between the uniform distribution $U$ and the distribution of tasks over volunteers $D$, normalised to the range [-1,1].
\begin{equation}\label{eq:p3Sat}
    sd(e,p_3)= 
    \begin{cases}
        1 - \displaystyle\frac{\Delta}{\epsilon} 
            & \text{, if } \Delta < \epsilon \\[1em]
        - \displaystyle\frac{ ( \Delta - \epsilon ) }{((\max \Delta) - \epsilon)} 
            & \text{, otherwise}
    \end{cases}
\end{equation}
where $\Delta=\mathit{difference}(D,U)$ represents the difference between the distribution of tasks over volunteers ($D$) and the uniform distribution ($U$), and $\max \Delta$ is the maximum possible value for $\Delta$. 
The range of $\Delta$, whether we use the earth mover's distance or the Kullback–Leibler divergence, is $[0,\infty)$, but a maximum value must be selected for our equations. Again, we argue that $\max \Delta$ is domain-specific and must be chosen for each context. 

Equation~\ref{eq:p3Sat} states that the degree of satisfaction is computed by mapping the difference $\Delta$ to the range $[-1,1]$. When this difference is in the range $[0,\epsilon]$, then this gets inversely normalised to the range $[0,1]$ to describe a positive degree of satisfaction (where $0$ gets mapped to $1$ and $\epsilon$ gets mapped to $0$). And when the ratio is in the range $[\epsilon,\max \Delta]$, then this gets inversely normalised to the range $[-1,0]$ to describe a negative degree of satisfaction (where $\epsilon$ gets mapped to $0$ and $\max \Delta$ gets mapped to $-1$). 

In summary, the degree of satisfaction of $p_3$ depends on how far is the difference $\Delta$ from $\epsilon$. The larger it is with respect to $\epsilon$, the higher the degree of dissatisfaction. The closer it is to $0$, the higher the satisfaction.

Suppose a mutual aid community $e$ provides incentives that motivate people to volunteer and already have norms ensuring tasks are spread equally among volunteers.  
Suppose that these regimented norms result in a  high degree of satisfaction for $p_3$, whereas the incentives result in a mediocre degree of satisfaction for $p_1$:

$$\begin{array}{ccc}
sd(e,p_1) = 0.5 & ; &
sd(e,p_3) = 0.9
\end{array}$$
And %let us 
say the importance of $p_1$ is set to be four times that of $p_2$, as follows: 
$$\begin{array}{ccc}
I_{c}(p_1) = 1 & ; &
I_{c}(p_3) = 0.25
\end{array}$$

Following Equation~\ref{eq:alignmentImp}, it is evident that the alignment of the mutual aid community $e$ with its understanding of the value fairness $\mathcal{V}_{c_{s}}'$ (Figure~\ref{fig:single-mothers-2}) becomes:
%% $$\mathcal{A}(e,\mathcal{V}_{c})= \frac{(I_{c}(p_1) \cdot sd(e,p_1)) + (I_{c}(p_3) \cdot sd(e,p_3))}{2} = \frac{0.5 + 0.45}{2} = 0.475 $$
\begin{align*}
    \mathcal{A}(e,\mathcal{V}_{c}) & = \frac{(1\cdot 0.5) + (0.25 \cdot 0.9)}{2} 
    = 0.385
\end{align*}

This is a positive alignment, however, it is not very high (recall that the range of positive alignment is $[0,1]$). Analysing the alignment of each property, it is evident that the alignment with $p_3$ is already very high. However, $p_1$ is considered much more important, and its alignment is mediocre. As such, this reasoning leads to the conclusion that better incentives are needed to increase the alignment with $p_1$, and hence, increase the alignment with fairness. This example illustrates how the formalisation of values presented in this paper provides the basis needed for future value-based reasoning to help improve value alignment.  

%\subsection{Naming our model}

With this, we conclude the proposed VTM for representing human values. 
To our knowledge, %we have presented the first formal proposal for the explicit computational representation of human values, which provides the foundations for computational reasoning and value-aligned AI engineering.
this is the first formal proposal for the explicit computational representation of human values, which takes into consideration value relations, value importance and computational semantics. This provides the foundations for computational reasoning over values. 
%
%From now on, we will refer to this model as the \emph{Value Taxonomy Model} (VTM). 
%
%The VTM can represent the values of humans, AI systems, and hybrid communities of AI systems and humans.
%
%The \emph{value system} of any of these comprises a set of value taxonomies specified in the VTM, as illustrated in Figure~\ref{fig:valueSystem}. 
%
In the next section, we demonstrate how our VTM proposal is consistent with research from social psychology.

\section{Aligning our Model with Social Psychology Research}\label{sec:SSHalignment}

The purpose of this section is not to reproduce or survey social psychology theories of values, but to analyse how the modelling choices introduced in Section~\ref{sec:valueTax} relate to key concepts established in that literature. This mapping is important because our goal is to provide a computational representation of values that remains grounded in existing understandings of human values while supporting their use in agent-based systems.

The proposed model aligns with several key characteristics of values identified in social psychology research. We assess this alignment by drawing in particular on the well-cited work of Rohan~\cite{Rohan2000}, which provides both a thorough review and a critical synthesis of values-related theory and research, culminating in the identification of key features of values within social psychology.

%Here, we aim to assess the alignment of our proposed VTM making reference to the body of research from social psychology. 
%
%To undertake this, we specifically choose the well-cited work of Rohan~\cite{Rohan2000}, %as it not only provides a thorough review of values-related theory and research but also extensively analyses existing work before identifying and distilling the key features of values within the field of social psychology, making it a comprehensive reference for our comparison. 
as it not only provides a thorough review of values-related theory and research but also offers a critical synthesis that incorporates her own perspective, culminating in the identification and distillation of key features of values within the field of social psychology, making it a comprehensive reference for our comparison. 
Rohan highlights five main aspects of the values construct discussed in the literature, and then proposes a stance on each aspect accordingly. 
Subsections~\ref{sec:SSrohan1}--\ref{sec:SSrohan5} carefully examine how our proposal for value representation is aligned with the discussion on each of those value aspects identified by Rohan. 
In Subsection~\ref{sec:SScontext}, we extend our analysis through the lens of Rohan's work to an additional aspect, the context of values, which we believe is crucial, even though it did not receive a thorough discussion in Rohan's work. 

As we show our consistency with the review of values from Rohan, we also make sure to compare to Schwartz's work~\cite{Schwartz2012AnOO}, since his theory of basic human values is considered to be ``one of the most commonly used and tested transcultural theories in the field of behavioural research''~\cite{GIMENEZ2019e01797}, and also because Schwartz's view on values has become a reference within the artificial intelligence and multi-agent systems communities. 

\subsection{On the use of the word values}\label{sec:SSrohan1}
Rohan identifies a key distinction in the use of value as a noun versus a verb.
\paragraph{Values as a noun}
Rohan, summarising Schwartz and Bilsky, highlights five commonly accepted features of values:
\begin{quote}
“(a) are concepts or beliefs, (b) pertain to desirable end states or behaviors, (c) transcend specific situations, (d) guide selection or evaluation of behavior and events, and (e) are ordered by relative importance”~\cite{Schwartz1987TowardAU}.
\end{quote}
Our value taxonomy aligns with this understanding. In our model, values are abstract constructs (e.g., fairness, equality) represented by label nodes. While we avoid committing to their nature as beliefs (feature a), we operationalise values through desirable end states via property nodes (feature b). Feature (d) is central to our alignment mechanism, which assesses behaviour against values. Feature (e), importance, is captured through weights in our taxonomy (see Equation~\ref{eq:alignment}).
Though values are said to transcend specific situations (c), we allow for their evolution over time and context, following van de Poel~\cite{vandePoel2018} (see Subsections~\ref{sec:context} and~\ref{sec:SScontext}).
%Additionally, Rohan and Schwartz both note a link between values and affect—an area requiring further research. Our example (Subsection~\ref{sec:howEG}) hints at value formation from lived experience, which we flag for deeper exploration in our roadmap (Subsection~\ref{sec:roadmapVdecisions}).
\paragraph{Values as a verb}
For Rohan, values used as a verb refers to evaluating any entity in light of its value system. Feather~\cite{feather1996values} captures this process:
\begin{quote}
“We relate possible actions and outcomes... to our value systems, testing them against our conceptions of what is desirable or undesirable.”
\end{quote}
This aligns precisely with our goal: formalising how actions relate to value priorities (Subsection~\ref{sec:why}, Equations~\ref{eq:alignment} and~\ref{eq:alignmentImp}) by enabling the evaluation of actions and behaviour with respect to values.
This is achieved by reinstating the connection linking abstract values to concrete, computable assessments of behaviour via property-based nodes.

\subsection{On values, value types, value priorities, and value systems}\label{sec:SSrohan2}

Rohan distinguishes between \textit{values}, \textit{value types}, \textit{value priorities}, and \textit{value systems}, all of which are explicitly modelled in our proposed VTM.

\begin{figure}[!t]
    \centering
    \includegraphics[width=\linewidth]{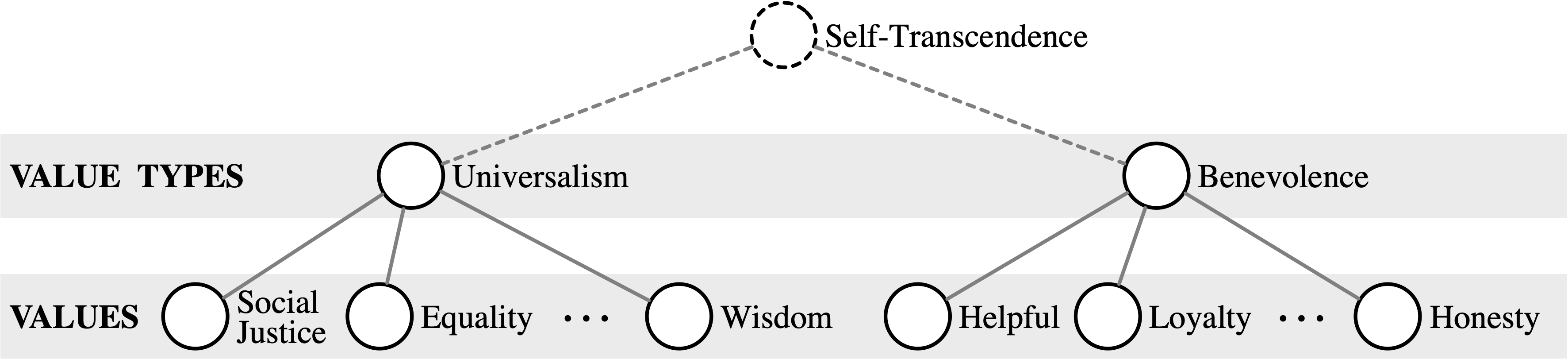}
    \caption{Values and value types}
    \label{fig:valueTypes}
\end{figure}

\paragraph{Values and Value Types} 
Rohan and Schwartz group individual values into broader value types. Examples of two of Schwartz's value types are presented in Figure~\ref{fig:valueTypes}, with a selection of values that fall under each type. While we support value grouping, we argue that restricting to two levels (values and value types) is limiting. Even Schwartz argues that different value types can be further organised into a set of higher-order groups. For example, for Schwartz, universalism and benevolence fall under the self-transcendence group, which our taxonomy can happily represent (see the dashed node and edges in Figure~\ref{fig:valueTypes}). 

Our taxonomy allows multiple levels of abstraction, consistent with van de Poel's work on value-sensitive design~\cite{vandePoel2018}. We generalise the concept and refer to all nodes as \emph{values}, distinguishing only between abstract \emph{label nodes} and grounding \emph{property nodes}. 

\paragraph{Value Priorities (Importance)} 
Whereas Rohan restricts value priorities to value types, we assign importance to all nodes in the taxonomy, regardless of their level of abstraction. This supports fine-grained prioritisation and reflects how values operate in practice. The assignment of importance must be coherent. For example, a high-priority parent cannot have only low-priority children. This mirrors Schwartz's averaging approach, where the importance of a value type is calculated as the average importance of its constituent values~\cite[p. 11]{Schwartz2012AnOO}, typically derived from survey responses that rank individual values.

\paragraph{Value Systems} 
Rohan and Schwartz describe value systems as a number of relevant value types and their relative importance. Structure within a value system is also key. Schwartz provides further information on this structure by introducing adjacency and oppositions between different value types. Our model captures this definition of value systems by allowing multiple disjoint value taxonomies to coexist, each with their own importance, as illustrated in Figure~\ref{fig:valueSystem}. These can evolve with context or domain, as supported by various context-specific value models in the literature~\cite{ChengFleischmann2010}. Future work should focus on the adjacency and opposition relations between our value taxonomies and ensure that the importance of values across taxonomies is coherent with these new relations.

\subsection{On value priorities and ``best possible living''}\label{sec:SSrohan3}

The third aspect identified by Rohan concerns what determines value priorities. While Schwartz links value importance to three universal human needs (biological, social, and group welfare), Rohan argues for a broader, more integrative criterion: the extent to which values enable ``best possible living,'' drawing from Aristotelian ideas of human flourishing (eudaimonia).

In her view, value systems help order which desires or requirements matter most for achieving this ideal. For example, if someone highly values power, then owning a new car may be seen as essential to their well-being through social superiority.

Our VTM proposal reflects this perspective by connecting abstract values (e.g., power) with concrete property nodes (e.g., owning a car), where the importance of values determines which desires or outcomes are most relevant to achieving the best possible living. The model ensures coherence between abstract and concrete levels: high importance at one level implies high importance at the other.

Aligned with Rohan, our approach defines value priorities through stakeholder-determined requirements and desires, without restricting them to Schwartz's three categories.

\subsection{From personal values to collective values}\label{sec:SSrohan4_summary}

Rohan's fourth aspect concerns the distinction between personal, social, and cultural value systems, all of which are supported in our model. Personal value systems reflect an individual's priorities, aligning with our definition of value taxonomies held by specific agents (modelled in VTM as $\mathcal{V}^{i}$). Social value systems represent one's view of others' value priorities, where the other may be an individual or a collective ($\mathcal{V}^{i>j}$ or $\mathcal{V}^{i>y}$). For the sake of completeness, we extend Rohan's definition of social value systems to include a \emph{group}'s view of others' value priorities ($\mathcal{V}^{x>j}$ or $\mathcal{V}^{x>y}$). Finally, cultural value systems refer to the values endorsed by a collective ($\mathcal{V}^{x}$). 

Rohan does not delve into the details of how the structure of collective (or cultural) value systems differs from others, but hints that they may share the same structure. From a computational perspective, the formal definition of value taxonomies for the individual, the view of others, or the collective should all be the same (i.e., the same model for value taxonomies applies to all). 
The difference is essentially in how value taxonomies are created. 
Rohan highlights the conceptual challenge of determining collective value priorities, whether through averaging, leadership input, or consensus. %We echo these concerns and suggest addressing them through mechanisms such as computational social choice or argumentation frameworks. 
We acknowledge these concerns and note that there are established solutions within AI to address them, such as computational social choice and argumentation frameworks (see Sections \ref{sec:whoImp} and~\ref{sec:roadmapVagreement}). %Norm-based governance, potentially via norm synthesis approaches, can also help manage collective value systems.

As for the mechanisms by which an entity forms a view of the values of another, this remains an open research question, as noted in our roadmap (see Subsection~\ref{sec:roadmapVdecisions}). 
We note here the preliminary work on the topic of learning the values of others from their behaviour, which in that specific work is defined through their responses to questionnaires~\cite{RodriguezSoto2024VALE}. 

%In summary, our model incorporates and extends the models of both Rohan and Schwartz in a consistent and coherent manner. %, though we allow for evolving and context-sensitive value taxonomies rather than assuming a fixed universal structure.

\subsection{On worldviews, ideologies, and other value-relevant concepts}\label{sec:SSrohan5}

Rohan introduces \emph{worldviews} and \emph{ideologies} as value-relevant constructs. Worldviews refer to the beliefs individuals hold about how the world is or should be, shaped by their value priorities. At the same time, ideologies are value-laden rhetorical tools used to justify decisions and actions. She states that these constructs are context-dependent and can be manipulated to align behaviours or narratives with desired values.

Concerning worldviews, Rohan highlights that people's value priorities shape their worldviews and perceptions; for example, influencing how they perceive authority or competition. These value priorities can change over time through experiences and social interaction. While our work does not yet explore these dynamics, understanding how values, beliefs, and experiences influence each other is a key direction in our proposed research roadmap.

%Our VTM proposal does not yet model how values influence perceptions or how they evolve through experience. We recognise the importance of these dynamics and include them in our future research roadmap, which calls for interdisciplinary collaboration with the social sciences.

%Rohan warns that ideologies, due to their rhetorical flexibility, can obscure genuine value priorities. We argue that AI can help clarify value-based reasoning through transparent and rule-based mechanisms, supporting better value-aligned decisions. 
Turning her attention to ideologies, Rohan states that value systems are used to support and justify complex decision making, which often requires conscious, deliberative, and justifiable reasoning. She argues that the rhetoric associated with ideologies can be manipulated to link almost any behavioural choice or state of affairs to a constructed set of value priorities. As a result, ideologies become ``remarkably slippery social constructions that take on different meanings over time and across political cultures''~\cite[p.~34]{4239}. 
Our proposal, along with others in the AI and value alignment community, offers a way to reason about values automatically and transparently. We believe this provides a promising avenue to mitigate the ambiguities and manipulations associated with ideologies, as identified by Rohan.

Schwartz does not directly address worldviews or ideologies but discusses other related constructs such as \emph{attitudes}, \emph{beliefs}, \emph{norms}, and \emph{traits}. These influence or justify behaviour and interact with value priorities in complex ways. We acknowledge the complexity of motivations and justifications of behaviour (Section~\ref{sec:whyValues}) and highlight the need for future work to explore these relationships in decision-making processes, as outlined in our roadmap (see Subsection~\ref{sec:roadmapVdecisions}).

\subsection{On the context dependency of values}\label{sec:SScontext}

Although Rohan and Schwartz acknowledge the context-dependency of values, neither provides an in-depth treatment. Rohan~\cite{Rohan2000} suggests that personal value priorities can shift with changing circumstances or social interactions. Furthermore, she states that values may be understood in terms of Bartlett's schemata~\cite[p.~201]{bartlett1995remembering}, as ``active organisations of past experience''. As for Schwartz, while his work focuses on universal, fixed values, he acknowledges that value importance can vary by person and context~\cite{Schwartz2012AnOO}. 
Our proposed model is aligned with Rohan and Schwartz's views in that value importance can change from one context to another (Section~\ref{sec:context}). Furthermore, we enable structural changes to capture evolving values by learning from experiences: nodes and edges can be added or modified as needed. This aligns with the view that values are shaped by past experiences, as in Bartlett's schemata. 
Holding different (or modified) value systems for different contexts is in fact further evidenced by the numerous domain-specific value systems proposed by different social scientists~\cite{ChengFleischmann2010}.

%This is evidenced by the numerous domain-specific value systems proposed by different social scientists~\cite{ChengFleischmann2010}, as well as our practical findings, such as the evolving uHelp taxonomy (see Subsection~\ref{sec:howEG} and Figure~\ref{fig:contextValueTaxonomy}).

%Our approach reflects the five types of value change identified in value-sensitive design~\cite{vandePoel2018}:
When discussing the change of values with context, it is important to note that our proposed VTM is aligned with the work on value change taxonomy in value-sensitive design~\cite{vandePoel2018}, which discusses the different ways in which values may change over time: (1) new values may emerge, (2) the relevance of a value may change for a given context, (3) the importance of a value may change, and (4) there may be changes in how values are conceptualised, as well as (5) changes in how values are specified, and translated into norms and design requirements. All these different types of change are reflected in our model through the possible addition of new abstract value concepts, new property nodes, or changes in the importance of existing value nodes. 

%A key innovation of our proposed VTM is its ability to visualise dynamically and prune taxonomies based on the current context, enabling a focus on context-relevant values while maintaining a richer, evolving general structure informed by prior experiences.
%%%% NARDINE!!! 

%Holding different (or modified) value systems for different contexts is in fact evidenced by the numerous domain-specific value systems proposed by different social scientists~\cite{ChengFleischmann2010}. %, as well as our practical findings, such as the evolving uHelp taxonomy (see Subsection~\ref{sec:howEG} and Figure~\ref{fig:contextValueTaxonomy}).
%The novelty of our proposal, however, is in visualising and hiding irrelevant branches of the taxonomy based on the current context. This enables taxonomies to evolve and grow with experience over time into larger and increasingly complex taxonomies that encode learnings through experiences in related contexts,  which then are temporarily pruned to match the current context. This allows focusing on relevant values based on context. 

Having shown how our proposed VTM is grounded in research from social psychology, we next outline a roadmap for future research on values in AI.

\section{A Roadmap for Value-Aligned AI Research}
\label{sec:roadmap}

This section presents a roadmap of what we believe are the key research challenges for developing AI systems that can represent and reason with human values. %The ultimate goal is enabling the identification and modelling of human values, reasoning about them, and explaining one's own behaviour and those of others in terms of those values~\cite{Osman24,VALE24,VALE23}. 
%
%(i)	identify and model human values,
% 
%(ii) reason about those values to determine behaviours for itself and of others, and 
%
%(iii) explain the reasoning processes which led to behaviour choices in terms of those values. 
%
%{\bf Nardine - please check the above key first line! And also the paragraph below!}
%
%
%There are different types of systems we envisage which all include an AI agent modelling values according to our proposed VTM, these include:
%We envisage different types of systems that would incorporate an AI agent that models values using our proposed VTM. 
We envisage different types of systems that would benefit from such capabilities and incorporate value reasoning using our proposed VTM. 
These include:
1) a single AI agent interacting with a human agent (such as a robot working with a patient, where the AI agent must understand and respond to the patient's individual values); 
2) a single AI agent interacting with a group of human agents (such as an AI advisor supporting medical decision making, where the AI identifies and reasons about potentially conflicting values of the patient, their family members, and the hospital, with the aim of raising awareness of these conflicts and supporting informed decisions by the human stakeholders); and %such as a hospital's medical professionals who are seeking advice from the AI on the alignment of their individual and collective action choices with respect to different sets of values); and 
3) a system of interacting human and AI agents (such as a smart urban mobility network, where autonomous vehicles and human users coordinate in real time, and the AI agents must represent and reason about both collective values and the individual values held by the different autonomous vehicles, each reflecting their owners' preferences). %such as a factory where robots and humans work together, where the AI may explicitly represent the values of the organisation so that continued negotiation and shared activity can take place effectively).

%Based on  
%(i)	existing literature on values in AI; 
%(ii) our investigation into research on human values from %social psychology, 
%(iii) working with different stakeholders to identify and formally specify their relevant values and developing appropriate value alignment mechanisms, and 
%(iv) through the process of developing our proposed VTM, we identify four key challenges in our roadmap. 

%These four challenges are as follows: 
We identify four key challenges in this field: 
\begin{enumerate*}[1) ]
    \item the value identification and representation challenge, 
    \item the value aggregation and agreement challenge, 
    \item the value-aware decision-making challenge, and 
    \item the value-aligned multi-agent system challenge.
\end{enumerate*}

%We address each of these challenges in the following four subsections, that are structured as follows:
%
%\begin{enumerate*}[1) ]
%    \item describes the research challenge, 
%    \item presents the key work in this area, and 
%    \item set out the research roadmap by detailing the research goals within each challenge and some associated proposals for research strategies. 
%\end{enumerate*}
%The VTM would benefit these research goals and strategies and the potential integration of that research, which we reiterate in the fifth and final subsection. 

We address each of these challenges in the following four subsections, where each is structured to include:
\begin{enumerate*}[1) ]
\item a description of the research challenge,
\item a summary of key work in the area, and
\item a roadmap outlining the research goals associated with each challenge, along with proposed strategies to pursue them.
\end{enumerate*}
The final, fifth subsection concludes by highlighting VTM's overall contribution to developing AI that can represent and reason with human values. 

\subsection{The value identification and representation challenge}
\label{sec:valueIdentification}

%For any reasoning over values to be possible, the relevant values in which the AI system will operate should first be identified and then computationally represented to facilitate automatic reasoning.  
%For any reasoning over values to occur, the AI system must first be capable of identifying the relevant values in its operational context and then representing them computationally to enable automatic reasoning. 
%
%This is typically broken down into two stages. 
%
%The first step is \emph{value identification}, which involves determining the relevant values for an individual or collective within the specific context where the AI will operate, %. %It is the challenge of establishing the relevant values for an individual or collective in the operating context within which the AI is being developed. 
%
%This includes understanding what words are used to describe relevant higher-level abstract value concepts, 
%such as ``fairness'' or ``transparency''. 
%
%The second step is \emph{value representation}, which involves providing formal computational models of those values and their underlying structure, including identifying their inter-relationships and relative importance and representing their grounding semantics.
For any reasoning over values to occur, the AI system must first be capable of identifying and representing the relevant values within its operational context. This begins with \emph{value identification}, which involves determining which values matter to an individual or collective in the specific domain of application, such as ``fairness'' or ``transparency''. Next is \emph{value representation}, which entails constructing formal computational models of those values and their underlying structure, including their interrelationships, relative importance, and grounding semantics.

Broadly, three different approaches may be followed for value identification and representation: 1) offline approaches, where relevant stakeholders manually identify and specify their values; 2) online approaches, where AI mechanisms like machine learning are used to identify and possibly specify values; and 3) mixed approaches, where AI mechanisms and users/stakeholders work together in tandem to identify and represent values. 

In all cases, whether values are manually provided or the AI has learned those values from human behaviour, we stress that these values should always represent the values of human stakeholders rather than artificial agents. 

\paragraph{Key research in this area} 

Current AI research on this topic has focused primarily on value identification, eliciting and learning relevant values from (typically written) records of human/agent interactions. 
Natural language processing techniques are used to estimate underlying human values from text in a (semi-)automatic manner. 
For instance, Liu et al.~\cite{Liu2019PersonalityOV} analyse values based on words used in e-commerce reviews, and Lin et al.~\cite{Lin2018AcquiringBK} estimate relevant values in tweets by combining textual features and contextual knowledge from Wikipedia. 
Brugnoli et al.~\cite{valawai23} employ a neural network model to categorise tweets based on the Moral Foundations Theory~\cite{GRAHAM201355}.  
In more recent work by Obi et al.~\cite{NEURIPS2024}, the authors aim to understand the alignment of LLMs with societal values and norms by examining the human values that are operationalised through the datasets used to fine-tune those LLMs. 
However, all these techniques are employed only once a predefined high-level value list has been identified in advance, such as the well-known Schwartz value system~\cite{Schwartz2012AnOO} or the Moral Foundations Theory~\cite{GRAHAM201355}. 
Using any predefined fixed list is a limitation, not only because it assumes the list is appropriate for the context, but also because it prevents values from changing over time, a view shared by the value-sensitive design community~\cite{vandePoel2018}. 

Amongst the approaches not starting with a predefined value list are the works of Wilson et al.~\cite{89b26068890444aa88b4a15afe36625c}, who present a crowd-powered algorithm to generate a hierarchy of general values, and Axies~\cite{s10458-022-09550-0}, which uses human and automatic techniques for identifying context-specific values using natural language processing. 

Another significant line of research involves learning the computational semantics of values, typically as reward functions or preference models. For example, (Cooperative) Inverse Reinforcement Learning (CIRL) aims to infer a human's reward function (a proxy for their values) from observed behaviour, with CIRL specifically modelling the collaborative nature of this learning process~\cite{10.5555/3157382.3157535}. 
Similarly, techniques for deep reinforcement learning from human preferences allow AI agents to learn complex behaviours aligned with human values by training on comparisons between behaviour trajectories provided by humans~\cite{10.5555/3294996.3295184}. 
Rodriguez-Soto et al.~\cite{RodriguezSoto2024VALE} addresses the specific problem of function identification, discussing how appropriate mathematical functions can be selected or designed to effectively learn and represent the underlying value semantics. 
%

%{\bf Nardine ... be careful of sentences like ''In~\cite{RodriguezSoto2024VALE}'' I think we agreed we wouldn't do this ... I may be wrong! }

%The preliminary work by Dewey~\cite{Dewey11} proposes that instead of building AI that maximises rewards, we should design AI that can learn what humans truly value by updating its goals (rewards) based on evidence from the world, making it possible for AI to pursue human values even if those values aren't fully known in advance. 
%
%In Leshinskaya and Chakroff~\cite{valueSemantics23}, the ``value-as-semantics'' framework proposes that human values can be represented as continuous attribute dimensions within a large language model's semantic space, allowing AI systems to quantitatively assess the desirability of any concept expressible in natural language while distinguishing between different types of values (like moral and hedonic). 

\paragraph{Key research goals with some suggested strategies}
The concrete representation of human values in most AI research typically remains abstract. 
Values are usually articulated through textual headings or labels (such as `fairness') without further exploring the concrete relations, semantics, or importance of each value listed, and there is no mechanism for deliberating and reasoning about these values. Initial works have begun exploring the computational semantics of values.  
%
%For example, in~\cite{s10458-022-09550-0}, values are defined through a name, a set of keywords, and a description in plain text of what it means to hold that value. 
%
The value representation we have presented uses property-based nodes to formally define the computational semantics of values, %(as property nodes essentially define how a value may be interpreted and assessed) 
while also explicitly capturing their interrelationships and relative importance. 

%The key research goals/challenges we have identified alongside potential associated strategies are presented next.
Next, we present the key research goals and challenges we have identified, alongside potential strategies, which are closely related to the limitations outlined above. 
The four proposed research goals focus on identifying value relations, value semantics, and value importance, as well as addressing issues such as value conflict resolution and dynamic value evolution.

\begin{enumerate}
\item \emph{Extending existing research on value identification and elicitation %(e.g.~\cite{s10458-022-09550-0})
so that relations between those values %initially identified by human/AI processes 
can be established.} 
We argue that the relations between values are essential for deliberation. Once these relations are identified, value taxonomies can be constructed to model them. 
This line of research also includes exploring how these structures might themselves be dynamic or context-dependent; and how nodes and edges evolve over time. This dynamism, crucial for real-world deployment, has not been addressed so far. 

\item \emph{Developing mechanisms for constructing property nodes for values and linking those property nodes to the abstract label nodes.} 
This is crucial for achieving any computational approach to building AI systems that can explicitly reason about values, as it is property nodes that make such reasoning possible. 
However, obtaining those property nodes is a challenging task. 
Designers and engineers must specify the formulae representing these property nodes, or an AI mechanism must be designed to learn the formulae that best describe the abstract value concept in question. Building on existing learning mechanisms~\cite{RodriguezSoto2024VALE,10.5555/3294996.3295184,10.5555/3157382.3157535}, future research can investigate how such methods can be extended to derive interpretable, structured representations of property nodes from data and interaction traces.
Research here should also investigate how to represent and manage the uncertainty or ambiguity inherent in the data and interaction traces.

\item \emph{Developing mechanisms for determining value importance, as it is the importance of values that ultimately influences behaviour.} 
As illustrated earlier, humans or AI may provide such measures of importance. For example, an AI can learn which values are more important than others from past interactions.
Humans and/or AI can provide a partial order over a subset of the value nodes of a taxonomy. Transformation mechanisms, such as that presented by Serramia et al.~\cite{10.5555/3237383.3237891}, can transform a partial order into concrete measures for value importance assigned to each node appearing in that partial order. Propagation mechanisms (such as the proposed Algorithm~\ref{alg:propagation}) can then be developed to compute the importance measure of the remaining nodes in the taxonomy, ensuring coherence of importance across the entire taxonomy.
Such transformation and propagation mechanisms will be helpful in practice because obtaining the importance measure of every taxonomy node is not always feasible (as per the discussion below). Research is also needed to explore how the importance of values might change in different contexts and over time, leading to dynamic prioritisation of values.

\item \emph{Designing methods for conflict detection and resolution \emph{within} and \emph{across} value taxonomies.} 
As entities typically hold sets of disjoint value taxonomies that collectively form their value system, conflicts may arise not only within a single taxonomy but also across separate taxonomies. For instance, privacy and security may reside in distinct taxonomies but still exhibit oppositional or trade-off relationships. 
Our proposed VTM supports resolving conflicts relating to the relative importance of values within a unified taxonomy. Extending it to handle cross-taxonomy relationships requires incorporating additional relational types such as adjacency (similarity), opposition, and orthogonality. These relations, inspired by work such as Schwartz's theory of universal values, enable reasoning over potentially conflicting but structurally separate value concepts. 
%
%Future work should develop reasoning mechanisms that can interpret and act on these richer inter-taxonomy relationships, especially when taxonomies remain disjoint but interact within an entity's broader value system. This would enable AI systems to resolve value conflicts in a way that better reflects the real-world complexity of human value structures.
\end{enumerate}

%These four research goals focus on identifying value relations, value semantics, and value importance, as well as addressing issues such as value conflict resolution and dynamic value evolution. This research line addresses the issue of value representation and building adaptable, context-aware value taxonomies, which provide the basis for deliberating and reasoning about values. 

%However, one thing we learn pretty quickly from social psychology research, is just how complex and nuanced human values are. As a result, it is rarely straightforward for humans to explicitly specify their value taxonomies. 
%
While many research ethicists working in the field of value-sensitive design (e.g., \cite {vandenHoven2015}) have been explicitly eliciting relevant value concepts from stakeholders, we believe that asking stakeholders to undertake such a process (as in the case of the uHelp app) may be too demanding in practice. Furthermore, identifying abstract value concepts, explicitly specifying value relations, importance, and formal semantics, as well as how these might change or conflict, is an even more significant challenge.
In our experience, only some stakeholders are ready to specify some value aspects, such as value concepts, relations, or importance. %This was, to some extent, the case of the medical doctors (see Subsections \ref{sec:whatImp} and~\ref{sec:ex-vtax}). 
% However, such an exercise is not straightforward for all stakeholders. 
% (Please note that when we use the word stakeholders, we include all users in that meaning.) 
This is why we propose to develop AI that can help identify values aspects from observing past interactions,  
and present them to stakeholders for review and refinement. This human-AI collaboration enables the development of robust, adaptive value taxonomies without requiring full manual specification.%We expect stakeholders to have a broad understanding of what an AI system learns of their value systems and the explicit way it models them (i.e. the constructed value taxonomies), including how the AI perceives value priorities and potential conflicts.
%
%We then expect these stakeholders to approve or disapprove of various aspects of the learned value systems and guide the AI in learning and representing values. %, potentially refining conflict resolution strategies or parameters for value evolution.
%
%What is difficult, in general, is to have any expectation that all stakeholders will have the time, willingness, or capability to specify the value importance of each node formally, the exact relationships between nodes, and the precise semantics of different abstract value concepts. %and the rules governing their adaptation or interplay in conflicting situations.
%
%Integrating AI with human stakeholders offers enormous potential benefits for helping individuals and organisations specify and maintain dynamic and robust value systems. %, but ut takes effort!

\subsection{The value aggregation and agreement challenge}\label{sec:roadmapVagreement} 
While the value identification and representation challenge focuses on identifying the relevant values of a single entity (e.g. individual, organisation, company) and how to represent them computationally, the value aggregation and agreement challenge focuses on the mechanisms required for defining the values of a collective (a group of individual entities). 
The objective is to move from a set of individual value systems (each defined by its own value taxonomies) to a single collective value system. 
This involves identifying and formalising the values that best represent the collective, based on the values held by its individual members.
%That is, to identify and formalise the values that best represent a collective by analysing the values of the individuals in that collective. 
%
This is a challenging task, as there is usually no clear consensus, resulting in inconsistencies between an individual's value system and the identified value system of the collective to which they belong.  

The research questions include: How do we transition from a set of individual value systems to one that represents the collective? How do we maintain some consistency between them (or not)? What does consistency look like? How do we deal with inconsistencies, and what do they entail? 

\paragraph{Key research in this area} 
Gabriel~\cite{10.1007/s11023-020-09539-2} argues that we live in a pluralistic world, where different entities hold different value systems. To ensure behaviour in a multi-agent system (MAS) is aligned with human values, decisions must be made about what collective value system the MAS should follow.  
%
%To arrive at the MAS value system, potential conflicting value systems of individuals or subgroups of individuals need to be addressed. 
However, this requires addressing the often conflicting value systems of individual agents or subgroups.  
Gabriel~\cite{10.1007/s11023-020-09539-2} defines this problem as identifying the value system that receives ``reflective endorsement despite widespread variation in people's moral beliefs''. 
Pigmans et al.~\cite{Pigmans2017,Pigmans2019} highlight the challenges of addressing conflicting individual interests in water policy-making, demonstrating that structured deliberation among stakeholders, explicitly centred on values, can support conflict resolution. %and report on how explicit deliberation processes focussed on the value systems of different stakeholders can help address such conflicts. 
Other work in this field uses computational social choice to aggregate individual value systems and yield a consensus value system~\cite{Lera-LeriBSLR22}. This approach considers a range of ethical perspectives, from utilitarian (maximising utility) to egalitarian (maximising fairness). 

However, these aggregation efforts face deep theoretical challenges. Foundational results in social choice and judgment aggregation show that constructing a coherent collective value system from individual inputs is often impossible without significant trade-offs. Arrow's Impossibility Theorem~\cite{arrow1963social}, %famously shows that no aggregation rule can convert individual preference orderings into a collective preference that satisfies a set of seemingly minimal fairness conditions (non-dictatorship, universality, independence of irrelevant alternatives, and Pareto efficiency). 
for example, demonstrates that no aggregation rule can convert individual preference orderings into a collective preference while satisfying basic fairness conditions (non-dictatorship, universality, independence of irrelevant alternatives, and Pareto efficiency). 
Similarly, Sen's ``Impossibility of a Paretian Liberal''~\cite{sen1970collective} illustrates a conflict between individual rights and collective rationality, highlighting that giving individuals minimal liberty in personal decisions can prevent the collective from satisfying Pareto efficiency. 
List and Pettit~\cite{list2002aggregating} further extend this line of reasoning to judgment aggregation, showing that even if individuals hold logically consistent sets of beliefs (e.g., over values or value-related propositions), aggregating these into a collective set can result in logical inconsistency. 
These impossibility results underscore the inherent difficulty of constructing collective value systems.  They suggest that trade-offs, structured limitations on the aggregation procedure itself (known as procedural constraints), or relaxed rationality assumptions are often necessary. 

\paragraph{Key research goals with some suggested strategies} 
Whilst research on value aggregation and agreements is emerging, challenges remain (as illustrated above). 
As with the challenge of value identification and representation, the focus will be on novel value aspects, such as value relations, value semantics, and value importance, which are necessary for enabling automated deliberation and reasoning over values. 

The identified key research goals, along with their associated strategies, are presented next.

%{\bf Nardine - would it help readability if we italicised the 1st sentence of each of these research goals so it looks like a clear heading?}

\begin{enumerate}
    \item \emph{Developing value aggregation mechanisms grounded in computational social choice.} These mechanisms must consider the value relations, semantics, and importance of the collective.   
    In other words, complex aggregation mechanisms are needed, advancing the work of Lera-Leri et al.~\cite{Lera-LeriBSLR22}, not only to aggregate the value importance of individual value concepts but to aggregate entire value taxonomies and value systems, including structural features like inter-value relations and property-based semantics.
    Research is needed to explore various aggregation rules, such as distance-based approaches (e.g., \cite{BrandtConitzerEtAlCOMSOC2016} for an overview including the Kemeny-Young distance-based rule) or methods tailored for aggregating structured information like those in judgment aggregation \cite{List2012JudgmentAggregation, GrossiPigozzi2014JAIntro}. The focus will be on how these rules handle the rich information in value taxonomies. 
    However, foundational results like Arrow's Impossibility Theorem~\cite{arrow1963social}, Sen's Liberal Paradox~\cite{sen1970collective}, and List and Pettit's judgment aggregation impossibility~\cite{list2002aggregating} highlight deep challenges in this space: no aggregation mechanism can satisfy all desirable properties. Future work should explicitly address these limitations by identifying acceptable trade-offs or relaxing constraints (e.g., independence or unanimity).
    
    \item \emph{Developing value agreement mechanisms based on agreement technologies.} %~\cite{AT2013}.} 
    As an alternative to aggregation, agreement mechanisms~\cite{AT2013} such as argumentation~\cite{rahwan2014argumentation} and negotiation~\cite{lopes2008negotiation} may help stakeholders establish a shared value system that is acceptable to all involved, even if it is not ideal.  
    This includes debating the inclusion of specific value concepts, their relations, their semantic interpretations (property nodes), and their relative importance. 
    The objective is to collectively agree on the value taxonomy that \emph{best} represents the group as a whole.  
    This approach may be especially promising when aggregation fails to satisfy fairness or coherence criteria, or when stakeholders seek procedural legitimacy through deliberation. 

    \item \emph{Developing mechanisms to manage inconsistencies between individual and collective value systems and to support their dynamic co-evolution.} 
    Perfect alignment between individual value systems and a derived collective system is rare. This research goal focuses on formalising such divergences, understanding their impact, and modelling how both individual and collective value systems evolve over time. 
    This includes at least three distinct but related avenues of research.  
    First, developing formal tools to represent and quantify discrepancies between value systems, including degrees of conflict or misalignment.  
    Second, investigating how explicit modelling of such discrepancies can stimulate deliberative refinement of collective values or prompt individual reflection and revision. 
    Third, modelling other dynamic drivers of value evolution, such as changes through experience, shifts in context, or learning over time, even in the absence of explicit disagreement or deliberation. This could include how agents may adjust value priorities through interaction, adapt pragmatically to collective views without fully endorsing them, or even withdraw from collectives when misalignment exceeds their tolerance. 
    A nuanced understanding of acceptance thresholds, tolerances for misalignment, and longitudinal shifts in value commitments (such as discarding obsolete values or integrating new ones) is critical to supporting real-world applications of value-driven AI. 

    In summary, tightly integrating conflict representation with mechanisms for value adaptation will enable more resilient and responsive value systems at both the individual and collective levels.

%\item Studying how uncertainty, vagueness, and ambiguity affect the modelling and agreement over values. 
%
%Since values like ``justice'' or ``autonomy'' often lack crisp definitions and may shift meaning across domains, future work is required to explore approaches that allow for partial or fuzzy semantics, probabilistic commitments, and bounded rationality in deliberation.
\end{enumerate}

\subsection{The value-aware decision-making challenge}\label{sec:roadmapVdecisions} 

Identifying value systems of individuals and collectives (Subsections \ref{sec:valueIdentification} and~\ref{sec:roadmapVagreement}) provides the basis for reasoning over values. 
Armed with the knowledge of their own value system, that of the collective to which they belong, and those of fellow individuals in the collective, an agent can reason about how to behave accordingly. 
Based on this understanding, agents can decide which actions to perform or not, including whether to join or leave specific groups or collectives. 
The computational challenge lies in developing advanced value-aware decision-making mechanisms that can take into account potentially conflicting value systems when guiding behaviour and choice, including value systems of the agent, the collective, and other individuals.

\paragraph{Key research in this area} 

Recent review work by Morgan and Wyner on AI agent reasoning with values highlights the diversity of approaches for integrating values into agent decision processes, including argumentation, deliberative mechanisms, trust-based reasoning, and value-aware planning~\cite{morganwyner2025}.

%Recent review work by Morgan and Wyner on AI agent reasoning with values highlights the diversity of approaches for integrating values into agent decision processes, including trust-based reasoning, value-aware planning, argumentation, and deliberative mechanisms~\cite{morganwyner2025}.

In value-driven decision making, persuasion has been one approach to motivating an agent to act in a specific way. 
In the work of Bench-Capon and Atkinson~\cite{Bench-Capon2009}, an argumentation framework is presented, illustrating how persuasion can rely on the strength of arguments, which in turn depends on a sophisticated understanding of social values. 

In the work of di Tosto and Dignum~\cite{TostoD12}, an agent model is described where agent actions are driven by their needs and values, where values are used to prioritise those needs. 
In agent-based simulation, Vanhée and Dignum~\cite{vanh.ae2018} propose a model in which agent decisions are guided by both rationality and culturally grounded values. The authors then analyse how individual decision making affects collective behaviour, measured by the aggregated success of agents in completing their tasks. Aguilera et al.~\cite{Aguilera2025}, on the other hand, propose a model that specifies values as reward functions in Markov Decision Processes (MDPs) to motivate action behaviour, and apply this work to promote better human development, such as addressing homelessness in cities.  

In the work of Chhogyal et al.~\cite{ChhogyalNGD19}, trust has been explored as a mechanism for influencing decision making, where the past reliability of an agent's actions is used to decide whether that agent can be trusted or not. 
The authors argue that when past experiences cannot be used to assess the reliability of others, the sharing of values between the trustor and trustee can help, and an approach is developed to evaluate trust based on the degree to which shared values can be established. 
In the work of Cranefield et al.~\cite{ijcai2017p26}, reasoning about values is employed to assist agents in making choices about which plans to adopt. Whereas in~\cite{RodriguezSoto2024AWAI}, values are used to help medical personnel make more value-aware decisions. 

Recent work by Karanik et al.~\cite{10.1145/3605098.3636057,10.1007/978-3-031-58202-8_11} has further highlighted the importance of explicitly modelling the structure of value systems, showing that relationships between values can significantly influence value-aligned decision making and behavioural outcomes.

In the work of Winikoff et al.~\cite{WINIKOFF2021103554}, the concept of ``valuing'' is introduced to the BDI model by allowing for preferences over actions and their outcomes. The evaluation of this work finds that ``valuing'' was the most preferred aspect of explaining courses of action. 
Beyond decision support and explanation, Dennis et al.~\cite{DENNIS20161} use formal verification to prove that autonomous systems satisfy specific ethical rules—operationalised representations of values—thereby providing high assurance in safety- and ethics-critical domains. %Winfield et al.~\cite{} propose architectural solutions that embed ethical constraints and social norms directly into a robot's control logic, treating them not as external evaluators but as integral components of the agent's reasoning process. 

More recently, learning-based methods have gained prominence in value-aligned decision making. Reward functions in reinforcement learning (RL) can be crafted to reflect human values, or agents can learn them directly from human feedback~\cite{10.5555/3294996.3295184}. Multi-objective reinforcement learning (MORL), as demonstrated by Rodriguez-Soto et al.~\cite{Rodriguez-SotoS22}, provides mechanisms for navigating and balancing multiple potentially conflicting values, allowing agents to trade off fairness, utility, and other normative considerations in their policy choices. The more recent work of Rodriguez-Soto et al.~\cite{Rodriguez-SotoS25} presents a multi-objective optimisation approach to help medical professionals assess medical actions against bio-ethical principles, such as beneficence, non-maleficence, autonomy, and justice.

\paragraph{Key research goals with some suggested strategies}  
One of the central objectives of the work on values in AI is to ensure that agent behaviour aligns with human values, known as the value alignment problem. Addressing this requires mechanisms for reasoning about values, particularly in decision making, deliberation, and explainability. %Introducing property nodes into value taxonomies provides the key construct for computational reasoning over values.
This underscores the need for the property nodes we introduce in our taxonomy, which provide the computational semantics necessary for representing and reasoning about values. 

%{\bf Nardine I was tempted to delete this last sentence as it feels out of place to me. Why are we suddenly speaking about our model? }

Next, we present the vital research goals and associated potential research strategies to address the value alignment of AI agents.    
\begin{enumerate}
    \item \emph{Developing mechanisms for reasoning about actions from the perspective of given value systems.}  
    A range of approaches could be pursued in this regard. One is to extend practical reasoning in cognitive agent models, such as BDI (Belief-Desire-Intention) frameworks, by explicitly incorporating structured value representations, including value taxonomies. As noted in Subsection~\ref{sec:SSrohan5}, the relationship between values and beliefs requires further interdisciplinary investigation.

    Another promising avenue is the development of value-enhanced theory of mind models, where agents observe each other's actions, build a model of each other's likely value systems and intentions accordingly, and reason about those actions and their potential underlying intentions, as in~\cite{MontesLORS23,valawai23b}. This would allow agents to make more informed, context-sensitive decisions and improve coordination, particularly in human-AI interaction settings.  
    Understanding the value systems of others can also inform and influence agent strategies, for example, to better persuade and influence them to modify their behaviour, which provides a bridge to the next research goal.  
%
    %But first, we note that while many existing works use values primarily as flat labels, this research line highlights the need for value models that encode value relations, semantics, and importance, as it is these aspects that enable computationally grounded reasoning over values.

        %{\bf Nardine IT feels a bit overkill to talk about our own work here. Should we not just do that in Section 6.5?}
    %
    \item \emph{Developing value-driven deliberation mechanisms that influence behaviour through persuasion, argumentation, or negotiation.} %This would extend existing work %, such as that of~\cite{Bench-Capon2009}, 
    %by incorporating more sophisticated value-based reasoning that considers value relations, semantics, and importance.
    This line of work enriches existing mechanisms by incorporating sophisticated value-based reasoning, which includes reasoning about value relations, semantics, and importance.

    Notably, one could aim not only to influence an agent's actions (deliberating \emph{with} values), but also to persuade others to revise their own value systems (deliberating \emph{about} values).  %
    Convincing others to change their value taxonomies can be an indirect approach to influencing their behaviour. 
    %
    %This could be achieved by developing negotiation/argumentation mechanisms that individuals can use to influence each other's value systems, either by introducing new values or changing the importance of existing values. 
    %
    Swaying individuals to modify their value systems can help find novel solutions for mutual agreements, as %. Traditional negotiation and argumentation mechanisms try to find acceptable solutions for all parties involved. By 
    modifying individuals' value systems %, we can enhance the traditional mechanisms as new value systems 
    can open the door to new solutions that would otherwise be off the table. 

    Furthermore, this aligns with the topic of \emph{value evolution} introduced in Subsection~\ref{sec:roadmapVagreement}. Deliberation mechanisms can help promote value evolution. Though this would require argumentation or negotiation frameworks that support proposals for adding, re-weighting, or reinterpreting values, thereby enabling dynamic reconfiguration of agents' internal value systems. 
    \item \emph{%Develop value-based explainability mechanisms to support human understanding of why certain actions are taken, both their own actions and those of others, in value-laden contexts.
    Develop value-based explainability mechanisms to help humans understand the rationale behind actions (both their own and others') in value-laden contexts.}  
    This includes explaining which values are promoted or demoted by a given action, as well as how value alignment is assessed. Property nodes in the value taxonomy can provide the semantic scaffolding needed for this reasoning. 

    Beyond justifying actions, explainability mechanisms can help users understand how different values are prioritised in decision making, making it easier to analyse, critique, or refine agent behaviour.
    \item \emph{Conduct interdisciplinary investigations into the interplay between values, emotions, and other motivational constructs.} 
    While our focus is on values as core motivators, Schwartz~\cite{Schwartz2012AnOO} highlights the role of emotions, attitudes, traits, and social norms. For example, Schwartz notes that ``[w]hen values are activated, they become infused with feeling.'' This suggests that emotions may not function as an independent motivator, but rather modulates the activation and perceived importance of values, amplifying their influence on decision making.   
    %
    %A promising next step would be to investigate how affective states (such as feeling satisfied about completing a difficult task), whether experienced internally or expressed externally, interact with the salience and importance of specific values, particularly as represented in structured value taxonomies. Such insights could inform models where value-driven behaviour dynamically shifts based on emotional and contextual cues. 
    A promising next step would be to investigate how affective states (such as feeling satisfied after completing a difficult task) interact with an individual’s value system to reinforce or shift value importance over time. This will require collaboration across disciplines including psychology, affective computing, cognitive science, and neuroscience. 
    %
    %This research direction requires collaboration across disciplines including psychology, affective computing, cognitive science, and neuroscience, to develop models that capture how values and emotions co-regulate behaviour and how this co-regulation might be represented computationally in AI systems. 
    %While we stress the importance of considering the role of values in guiding behaviour, Schwartz talks about the role of emotions and how behaviour can be explained in terms of one's attitudes, beliefs, traits and social norms. 
    %  
    %Emotions have received growing attention in the AI literature, and as such, relating affect (the display or experience of emotion) to values and/or behaviour could be one of the next steps in AI research on values. 
    %
    %Schwartz~\cite{Schwartz2012AnOO}, for example, states that ``[w]hen values are activated, they become infused with feeling.'' If the affect is linked to values and is not an independent motivator in guiding behaviour, then linking the proposed value taxonomies with that affect would constitute the first step in this novel research line. % Affect used as a noun here

    Beyond affect (i.e. emotions), future research is needed to explore how other motivators, such as attitudes, traits, and beliefs, interact with value-based reasoning. This line of enquiry will again require a critical, multidisciplinary analysis from the social sciences and humanities, as well as potentially cognitive science and neuroscience, to understand % the interplay of values, emotions, and cognition.  
    the interplay of these psychological and cognitive factors in human and AI reasoning. 
\end{enumerate}
    
\subsection{The value-aligned multi-agent system challenge}\label{sec:roadmapVdesign}

While the third challenge focuses on the value alignment of an individual's reasoning and decision-making process, this fourth challenge concerns designing and developing MASs whose overall behaviour is driven by human values. %promotes a set of agreed human values. 
The objective is to focus not on the individual agents (human or software) but on the behaviour of the resulting interactions of those agents as a whole. % and whether its mediation results in better-aligned behaviour. 

The research questions here are: How do we design a system that promotes value alignment? How can such a system evolve and adapt to maintain optimal alignment?

\paragraph{Key research in this area} 
The design of technologies aligned with our human values is a well-established field known as value-sensitive design (VSD)~\cite{10.1561/1100000015}. 
The VSD approach typically involves conducting conceptual investigations, often through participatory design workshops with stakeholder groups. 
This enables designers to understand the experience of stakeholders as they use a given technology and any aspired values they may have for future engagement with the technology. 
Empirical investigations inform technology designers of these values, and technical investigations evaluate the adherence of system behaviour to the desired values and analyse how people use the technology. Sometimes, new considerations emerge due to how people use the technology, and the participatory design process is repeated as needed. 
Whilst VSD relies on offline participatory design and offline evaluations, the proposed AI research complements this approach by providing an online reasoning mechanism that computationally assesses the degree to which these systems align with different human values. 

Since norms have traditionally been used in MAS to mediate behaviour, proposed mechanisms that assess a MAS's alignment have been reduced to assessing the value alignment of the MAS's norms. If a set of norms brings about outcomes more aligned with a given value system, the set of norms and its corresponding MAS are said to be more aligned with that value system. The research in this field has focused chiefly on choosing an optimal set of norms that optimise the value alignment of the MAS~\cite{MontesS21,SerramiaLR20}. 
In the work of Serramia et al.~\cite{SerramiaLR20}, norm synthesis is automated using prior knowledge about the relationships between specific norms and the values they promote. 
The work of Montes and Sierra~\cite{MontesS21} proposes a value-promoting norm synthesis approach that, in essence, optimises the value alignment mechanism proposed by Sierra et al.~\cite{abs-2110-09240}. 
In that work~\cite{abs-2110-09240}, value preferences are understood as preferences over world states and the value alignment of a set of given norms is based on the degree to which those norms move us towards preferred states. 
Rodriguez-Soto et al.~\cite{RodriguezSoto2024AWAI,RodriguezSoto2024VALE,Rodriguez-SotoS25}, is designing mechanisms based on multi-objective optimisation for assessing the alignment of medical protocols with bio-ethical principles like beneficence, non-maleficence, autonomy and justice. 
Another interesting application area is the use of simulations to analyse norms (policies) from the perspective of values that relate to fighting inequality and discrimination~\cite{alba24,CurtoMSOC22}. In this line of work, inequality is formally defined using quantitative measures such as the Gini index~\cite{gini}, and simulations are employed to evaluate the impact of discriminatory norms on levels of inequality.  

%{\bf Do we say enough in this final line to merit it being a part of our text?}

\paragraph{Key research goals with some suggested strategies} 
Subsection~\ref{sec:roadmapVdecisions} discussed the research goals for reasoning about values on a micro level: the agent level and its individual decision-making processes. 
This subsection presents the research goals for reasoning about values on a macro level: the systemic behaviour emerging from multi-agent interaction and collective self-governance. 
This involves not only top-down governance through norms, but also designing the fundamental rules of interaction and fostering bottom-up emergence of value-aligned behaviour. 
Some goals overlap with the previous subsection ---namely, reasoning over value alignment, explainability, and deliberation mechanisms--- but the emphasis shifts from individual choice to collective outcomes and institutional dynamics. 
The identified key research goals, along with their associated strategies, for the macro level are presented next. 

\begin{enumerate}
    \item \emph{Developing mechanisms for assessing the value alignment of norms.} This includes both traditional approaches to norms in multi-agent systems: regimented norms, where norms are enforced structurally, and the more general enforced norms, where compliance is encouraged through incentives and sanctions, similar to incentive-based approaches from mechanism design. 
    
    Despite some success with the preliminary research on the value alignment of norms, some mechanisms reason about values without fully capturing their underlying semantics~\cite{SerramiaLR20}. 
    Providing value semantics (specified through property nodes in our proposed taxonomy) enhances the ability to reason about value alignment, and enables clearer and more grounded explanations (discussed shortly). 
    
    However, a major limitation of current approaches is their dependence on manually defined mappings between norms and values. This challenge could be mitigated by leveraging automatic value identification and representation techniques, as discussed in Subsection~\ref{sec:valueIdentification}. 

    A further challenge is to ensure these mechanisms remain robust as value systems evolve, whether through cultural shifts, stakeholder input, or domain-specific developments. 
    
    %\item \emph{Developing value-based explanation mechanisms for normative choices.} %Developing explanation mechanisms that support human users in understanding why one set of norms is preferred to another with respect to a given value system. These mechanisms should help human users and stakeholders understand why certain norms (or sets of norms) are more aligned with a given value system, especially when trade-offs are involved.  Explanations should build on the alignment assessments that reflect both value semantics and importance hierarchies, allowing for richer reasoning and explanations. Such explanation tools could not only support the design of MASs, but also aid policy-making, participatory governance, and the iterative refinement of socio-technical systems. Examples include medical triage protocols, climate adaptation policies, or regulatory schemes for AI safety.

    \item \emph{Developing mechanisms for value-aware self-governance in MAS.} 
    The goal is to enable MASs to autonomously evaluate and adapt their own institutional norms in response to evolving contexts and changing value priorities. 
    This includes two complementary research directions:

    First, there is the issue of norm discovery and adaptation. Using norm synthesis, formal analysis of norms, simulation, and optimisation techniques, systems can explore normative spaces and assess alignment. Importantly, these mechanisms should explicitly account for changing value importance and the emergence of new value conflicts.
    Explanatory mechanisms should be integrated into this process to ensure transparency in why certain norms are adopted or revised.
    
    Second, there is the issue of value-driven normative deliberation. Collective norm choice among agents requires negotiation frameworks informed by value alignment metrics and explanation tools. 
    The system must support agents in reaching an agreement not only on coordination goals, but also on the underlying preferred values and any trade-offs agreed upon during negotiation. 
    This becomes particularly important in dynamic or pluralistic environments where agents hold differing value systems, and consensus must be built through deliberative processes. 
    
    %Third, self-governance must also encompass general value-aware deliberation mechanisms that drive collective behaviour beyond explicit norm deliberation.
    %%
    %These include negotiation, argumentation, and consensus-building processes that address value conflicts and collective decision making even in the absence of formal norms.
    %
    %Such mechanisms should leverage value alignment assessments and explanations to guide agents towards mutually acceptable agreements or shared value-based rationales for action.

    Finally, self-governance must extend beyond operational norms to include meta-level governance structures: that is, rules governing how norms are created, revised, or enforced. Designing such meta-institutions to reflect values such as inclusivity, fairness, and accountability is a significant challenge. 

    \item \emph{Developing value-driven deliberation mechanisms that influence collective behaviour.} 
    This goal focuses on multi-agent deliberation where agents negotiate or argue not only to coordinate, but to shape collective outcomes in value-aligned ways, sometimes even in the absence of explicitly codified norms. Agents may engage in joint decision making where pluralism or value conflicts exist. 
%Argumentation frameworks in which values shape both the construction and evaluation of arguments;
%Negotiation protocols that incorporate value-based preferences, constraints, and justifications;
%Interaction strategies for mixed-motive settings (e.g., trade, conflict resolution) where aligning on action requires understanding and navigating differing values.
    A prominent example is in climate policy negotiations, where decisions about mitigation strategies or financial support for climate adaptation depend on how individual nations' values (e.g., economic growth, sovereignty, historical responsibility) interact with global collective values (e.g., intergenerational justice, ecological sustainability, fairness). Mechanisms that help agents understand, negotiate, and reason about these intertwined value layers are crucial for effective coordination. 
    This includes developing value-aware social reasoning mechanisms, where agents maintain and update models of the values, value priorities, and value interpretations of other agents in order to anticipate disagreements, tailor arguments or explanations, adapt interactions, and improve coordination. This can be understood as a value-oriented form of Theory of Mind, where agents reason not only about the beliefs or intentions of others, but also about the values that may motivate their behaviour and decisions. 
    %Mechanisms developed here also include developing argumentation frameworks and negotiation protocols that integrate collective value considerations and influence collective behaviour.  
    Mechanisms developed here also include argumentation frameworks and negotiation protocols that integrate collective value considerations and influence collective behaviour.
    %These mechanisms are essential in open, dynamic systems where agents rely on reasoning about both individual and collective values to coordinate behaviour. 
    While this is closely related to the deliberation goal of Section~\ref{sec:roadmapVdecisions}, the emphasis here shifts from influencing and persuading individual agent behaviour to shaping shared decisions and agreements that govern collective behaviour. 

    \item \emph{Developing value-based explanation mechanisms.}  Explanation tools play a cross-cutting role in both normative and non-normative approaches.  
    They help users, designers, and agents understand how norms, negotiation outcomes, or deliberative decisions relate to underlying value systems. 
    Explanations should build on alignment assessments that incorporate both value semantics and importance hierarchies, enabling richer reasoning and explanations by capturing trade-offs and adapting to evolving value priorities. 
    Such tools could not only support the design of MASs, but also aid policy-making, participatory governance, and the iterative refinement of socio-technical systems. 
    %These explanations should reflect the semantics and structure of the value systems, including trade-offs, importance hierarchies, and evolving priorities. 
    Examples include medical triage protocols, climate adaptation policies, or regulatory schemes for AI safety.
\end{enumerate}

\subsection{A note on the VTM's contribution to future work}
In presenting the research challenges, related research goals, and associated research strategies outlined here, it becomes clear that one main recurring drawback of current research is its need for greater consideration of concrete representations of values. 
For example, in some cases, value concepts are identified but not their importance, although the importance of values ultimately influences behaviour. 
In other cases, value importance is considered, but not value semantics, which critically limits reasoning about values. Value semantics are needed for enhanced value reasoning and explainability. 
These pitfalls are a natural consequence of the absence of a universally accepted, formal model for representing values. Our proposed value taxonomies address this by introducing value relations, semantics, and importance into a structurally coherent and conceptually intuitive foundational model for value representation. %, thereby addressing the challenges identified in this section.

\section{Strengths, Limitations, and Implications for Value-Aware MAS}%Reflections on the Strengths and Challenges of the VTM} 
%: What does the model buy us? And what are the general challenges of modelling human qualities - such as values - for computational reasoning in AI?}
% NARDINE - should we talk about the pitfalls of BDI modelling here as well!! Just a thought ... 
\label{sec:strengthsLimit} 

%After presenting our formal model of human values (the proposed VTM), %and demonstrating its critical relationship with research outside of computer science, exploring how it can be used to build a roadmap for value-aligned AI systems, we now reflect on whether our work has fulfilled the principles set out in Section~\ref{sec:intro}. 
%and outlining how it can inform a roadmap for developing value-driven AI systems, including its integration with research beyond computer science, we now reflect on whether our work fulfils the principles outlined in Section~\ref{sec:intro}.

Having presented the VTM and outlined how it can support future research on value-aware AI, we now discuss the strengths and limitations of the model with respect to its role as a foundational representation for value-aware multi-agent systems. We organise this discussion around the guiding principles introduced in Section~\ref{sec:intro}, while also making explicit the current boundaries of the proposal.

%%%%%%%%
\paragraph{Using Formal Methods} 
According to Rohan~\cite{Rohan2000}, research into value theory has suffered from ``definitional confusion'' due to the word ``values'' being abused and overused by both non-psychologists and psychologists alike. 
Our proposed VTM provides an intuitive, precise and concrete way of defining values and various value-related concepts (such as value priorities, value semantics and value systems) (Principle 1). Furthermore, our proposed VTM is designed to serve as a foundation for the data structures and algorithmic design necessary for value-based reasoning (Principle 2).
Recall that the interest in working with values, as Schwartz~\cite{zis-Schwartz2007Value} puts it, is because values are ``abstract motivations that guide, justify, and explain attitudes, norms, opinions, and actions.''. 
%
%For us, the objective of modelling values computationally means that we can systematically reason and explain AI or human behaviour from the perspective of values. 
For us, the objective of modelling values computationally is to provide the representational basis for systematically analysing and explaining AI or human behaviour from the perspective of values.
Last, our proposed model also subsumes and relates to established concepts in AI research (Principle 4). The roadmap illustrates how our model can help extend and build upon existing research in the field. 

%%%%%%%%
\paragraph{Remaining Agnostic to Implementation Choices} 
When establishing any foundational model, choices regarding representation and language must be made. 
Each choice carries opportunities and limitations. It is essential to be mindful of any implicit assumptions or representation limitations a choice may introduce. 
This is why one of our guiding principles (Principle 3) is to remain agnostic on specific implementation choices, allowing us to focus on theoretical notions that provide clarity over potential implementation details.  
While we deliberately leave the implementation choice open for system and research development, we do not hesitate to propose some initial implementation choices that, along with our running example, better illustrate the significance and implications of our model. 

%%%%%%%%
\paragraph{Being Grounded in Social Psychology} 
When we assume the task of modelling something as nuanced and complex as human values, we begin with insights from the social sciences. 
In particular, we ground our proposed VTM in the wealth of research from social psychology (Principle 5), aligning with the views of Schwartz and other leading researchers~\cite{Rohan2000}. 
%
%By doing so, we have set out to design an academically significant and conceptually intuitive model for both non-computer scientists and AI researchers. 
%
%This conceptual intuitiveness supports the promise of becoming an agreed-upon conceptual model for future interdisciplinary research and the practical development of value-based AI systems. It is with this in mind that we developed the proposed research roadmap. 

By doing so, we have set out to develop a conceptually intuitive model accessible to both non-computer scientists and AI researchers. This intuitive foundation underpins the development of an interdisciplinary framework for value-based AI, as proposed in the research roadmap. 

\paragraph{Remaining Agnostic to  Specific Theories of Values} 
%
%We also argue that a computational model of human values should be agnostic to a specific theory of values (Principle 3), aligning with proposals from the social psychology literature in general. 
%
%This is precisely what our proposed VTM achieves. It allows the specification of contextual dynamic value systems that accommodate different values, such as moral, economic and epistemic values. 
We also argue that a computational model of human values should remain agnostic to any specific theory of values (Principle 3), while aligning with proposals from the social psychology literature in general. 
Our proposed VTM supports this by enabling the specification of context-dependent, dynamic value systems that can accommodate different values, such as moral, economic and epistemic values. The dynamic nature of values aligns with some works in value-sensitive design (e.g.~\cite{vandePoel2018}).

%%%%%%%%
\paragraph{Demonstrating Applicability to Real-Life Scenarios} 
An essential challenge in modelling human values is ensuring the expressiveness and practicality of the resulting model so it applies to real-life scenarios (Principle 6) with demonstrable practical benefit. 
Our proposed VTM was inspired by our collaboration with real stakeholders in different settings (e.g., social networking application, hospital working practices for medical professionals, and training of firefighters in emergency scenarios).  
Our practical work has supported the development of the formal model, and the formal model has given us the foundational and structural grounding to design our value-aware AI systems. 
In this paper, we have provided a detailed running example of an existing system (uHelp) that illustrates the applicability and potential of our model. %demonstrates the applicability, impact, and potential of our model. 

From a multi-agent systems perspective, the VTM provides a representation layer that can support several forms of value-aware reasoning: agents can evaluate behaviours with respect to value-grounding properties, collectives can be represented through shared or context-specific value taxonomies, and norms or institutional mechanisms can be assessed according to the values they promote. The model therefore does not replace existing MAS constructs such as beliefs, goals, norms, or utilities, but provides an additional layer for representing and evaluating the values that motivate or justify them.

The practical relevance of this representation layer is further illustrated in collaborations in healthcare and emergency response domains. In collaboration with Hospital del Mar in Barcelona, we investigate the formalisation and operationalisation of values relevant to clinical decision-making, specifically the four bioethical principles of beneficence, non-maleficence, autonomy, and justice~\cite{RodriguezSoto2024VALE}. Through user studies, we examine how medical professionals interpret these values within institutional contexts and use the resulting computational semantics to support tools for evaluating the alignment of clinical actions and protocols with institutional values~\cite{Rodriguez-SotoS25,RodriguezSoto2024AWAI}. Such tools are particularly relevant in situations involving value conflicts, such as tensions between respecting patient autonomy and ensuring well-being. In parallel, our collaboration with the Catalan Fire Brigade has led to the development of tools that support the training of firefighter students by helping instil the institutional values of their fire department and training them to make decisions guided by these organisational values rather than solely by personal value preferences~\cite{Osman2026,RodriguezSoto2025VALE}.

\paragraph{Staying Mindful of the Limitations of Formal Modelling} 

Modelling human values using formal methods is inherently challenging, not only because it requires deciding which aspects of a complex world to represent, but also because values themselves are often vaguely understood or inconsistently defined by people. Unlike physical or technical systems, where relevant properties are typically well agreed upon, human values come from diverse, subjective, and sometimes conflicting perspectives. They are also often only loosely or intuitively understood, even by the individuals who hold them. 

Our proposed Value Taxonomy Model (VTM) provides a structured framework for formally representing values. It introduces elements such as value importance, semantic property nodes, and their relationships. These elements can be specified manually by humans or learned from observations and interactions. However, asking humans to specify these elements presents serious challenges. People may struggle to explain what a value like ``fairness" means in practical terms, they may disagree on its interpretation, or find it difficult to link abstract values to specific behaviours or properties. Furthermore, prioritising values or articulating how they relate to each other often requires a level of reflection and deliberation that people may not have previously undertaken. 
Learning these components from observations and interactions is also not straightforward, since it involves interpreting implicit value signals and reconciling inconsistencies across different perspectives. Even when data is available, systems must infer semantics from behaviours or preferences that are often context-dependent, incomplete, or ambiguous.

Thus, whether specified or learned, capturing the computational semantics of human values remains a complex task. The difficulty lies not only in the technical representation but also in the inherent vagueness, variability, and context-dependence of human values. Achieving even a partial, operational understanding of values is difficult, but it is essential for building AI systems that can reason meaningfully about value-aligned behaviour. %challenging; yet, it is essential for enabling AI systems to reason meaningfully about value-aligned behaviour. 

These limitations also clarify the intended scope of the present paper: the VTM is not a complete architecture for value-aware agency, but a representational foundation upon which such architectures can be built.
\section{Concluding Remarks}\label{sec:conc}

This paper contributes to the growing research on values in AI by proposing the Value Taxonomy Model, a formal computational framework for representing human values as a foundation for value-aware reasoning in multi-agent systems. Grounded in social psychology and connected to existing AI research, the model provides explicit structures for representing value relations, value importance, and computational semantics. To our knowledge, this is the first proposal that brings these dimensions together in a unified formal representation of human values.

Through the VTM, we introduced a representation of values based on taxonomies composed of abstract value labels and grounding property nodes, enabling the evaluation of behaviour with respect to values. We further demonstrated how the model can represent contextual variation, individual and collective value systems, and mechanisms for assessing value alignment. In doing so, the VTM extends existing approaches to values in AI by providing a computational representation designed to support future work on value-aware decision-making, explanation, coordination, and governance in multi-agent systems. 

We illustrated the proposed model using the uHelp application, a social network where users request and offer help with everyday tasks. Through this example, we demonstrated how values such as fairness can be represented computationally, how their alignment can be assessed, and how value priorities may vary across time and context. The example also illustrated how assessments of value alignment can inform adaptations in norms or decision-making processes, ultimately shaping agent behaviour within a multi-agent system.

While the present work focuses on the representational foundations required for computational models of values, many important challenges remain open. These include learning values from behaviour, integrating values with norms and other motivational constructs, supporting deliberation in the presence of conflicting values, and enabling value-aware coordination and governance in multi-agent systems. We outlined a research roadmap for these challenges in Section~\ref{sec:roadmap} and argued that explicit computational representations of values are necessary to support progress in these areas.

%We have illustrated the proposed model using the uHelp application, a social network where users request and offer help with everyday tasks. The example demonstrates how the value of fairness can be specified and how it can evolve over time and in different contexts. 
%We also discussed how assessing alignment with such values can trigger adaptations in system norms or decision-making processes, ultimately shaping behaviour.

The ultimate aim of this research is to support the design of real-world applications in which human and artificial agents can represent, evaluate, and deliberate about values explicitly. By providing a formal representation layer for values, the VTM can support future work on value-aware decision-making, norm design, explanation, and governance in multi-agent human-AI systems. In this way, the model contributes not only to value alignment in automated systems, but also to broader efforts to make individual, collective, and organisational decision-making more value-aware.

%The ultimate aim of this research is not only to provide a foundation for future theoretical work but also to enable the design of real-world applications that support value-driven decision-making for multi-agent human-AI systems. 
%Equipping AI with the ability to reason about human values not only fosters alignment in automated decisions but also promotes value awareness at the individual, collective, and organisational levels of humans, thereby strengthening their capacity for making value-aware decisions.
%
%This is illustrated further in our ongoing work with Hospital del Mar, Barcelona where we focus on formalising and operationalising the values relevant to clinical decision making: the four bio-ethical principles of beneficence non-maleficence, autonomy and justice~\cite{RodriguezSoto2024VALE}. We run a user study to learn how doctors in a given hospital understand these values. We then build on these learned computational semantics of values to develop tools that help medical staff assess how well specific actions align with institutional values, and explore how clinical protocols might be modified to improve alignment~\cite{Rodriguez-SotoS25,RodriguezSoto2024AWAI}. This is especially useful in cases of values conflicts, such as tensions between respecting a patient's autonomy and ensuring beneficence. 
This is further exemplified in our ongoing collaboration with Hospital del Mar in Barcelona, where we focus on formalising and operationalising values relevant to clinical decision-making: specifically, the four bioethical principles of beneficence, non-maleficence, autonomy, and justice~\cite{RodriguezSoto2024VALE}. 
We conduct user studies to learn how doctors within a given hospital interpret these values. We then utilise the learned computational semantics to develop tools that assist medical staff in evaluating the alignment of specific actions with institutional values and in exploring how clinical protocols can be revised to enhance such alignment~\cite{Rodriguez-SotoS25,RodriguezSoto2024AWAI}.
This is particularly useful in cases involving value conflicts, such as tensions between respecting a patient's autonomy and ensuring their well-being (the principle of beneficence).

\backmatter

%\bmhead{Acknowledgements}
%This work was supported by the EU-funded VALAWAI (\#~101070930) project and the Spanish-funded EVASAI (\#~PID2024-158227NB-C31) project. 

\section*{Declarations}
\begin{itemize}
\item Funding: This work was supported by the EU-funded VALAWAI (\#~101070930) project and the Spanish-funded EVASAI (\#~PID2024-158227NB-C31) project. 
\item Competing interests: The authors have no competing interests to declare that are relevant to the content of this article.
\item Ethics approval and consent to participate: Not applicable. 
\item Consent for publication: Not applicable.
\item Data availability: Not applicable.
\item Materials availability: Not applicable.
\item Code availability: The code implementing the algorithms presented in this study is publicly available at the links provided in the manuscript. 
\item Author contribution: All authors contributed to the study conception and design. All authors contributed to manuscript writing and approved the final manuscript.
\end{itemize}
\noindent

\bibliography{bibliography}
\end{document}